\documentclass[twoside,11pt]{article}
\usepackage{jmlr2e}
\usepackage{amsmath}
\usepackage{tikz}
\usepackage{subfigure,epic}
\usepackage{multirow}
\usepackage{array}
\usepackage{booktabs}
\usepackage{algorithm}
\usepackage{algpseudocode}

\usetikzlibrary{arrows,shapes,chains}
\newtheorem{problem}{Problem}
\newtheorem{assumption}{Assumption}

\makeatletter
\newenvironment{breakablealgorithm}
  {% \begin{breakablealgorithm}
   \begin{center}
     \refstepcounter{algorithm}% New algorithm
     \hrule height.8pt depth0pt \kern2pt% \@fs@pre for \@fs@ruled
     \renewcommand{\caption}[2][\relax]{% Make a new \caption
       {\raggedright\textbf{\ALG@name~\thealgorithm} ##2\par}%
       \ifx\relax##1\relax % #1 is \relax
         \addcontentsline{loa}{algorithm}{\protect\numberline{\thealgorithm}##2}%
       \else % #1 is not \relax
         \addcontentsline{loa}{algorithm}{\protect\numberline{\thealgorithm}##1}%
       \fi
       \kern2pt\hrule\kern2pt
     }
  }{% \end{breakablealgorithm}
     \kern2pt\hrule\relax% \@fs@post for \@fs@ruled
   \end{center}
  }
\makeatother

\usepackage{lastpage}
\ShortHeadings{AdaDKRR for  Data Silos}{Wang et al.}
\firstpageno{1}

\begin{document}

\title{Adaptive Distributed Kernel Ridge Regression: A Feasible Distributed Learning Scheme for Data Silos}

\author{\name Di Wang \email wang.di@xjtu.edu.cn
       % \addr Center of Intelligent Decision-Making
       %  and Machine Learning\\
       %  School of Management\\
       % Xi'an Jiaotong University\\
       % Xi'an, China
       \AND
       \name Xiaotong Liu \email ariesoomoon@gmail.com
       \AND
       \name Shao-Bo Lin\thanks{Corresponding author} \email sblin1983@gmail.com \\
       \addr Center for Intelligent Decision-Making
        and Machine Learning\\
        School of Management\\
       Xi'an Jiaotong University\\
       Xi'an, China
       \AND
       \name Ding-Xuan Zhou \email  mazhou@cityu.edu.hk \\
       \addr  School of Mathematics and Statistics,\\
       University of Sydney, Sydney, Australia
}

\editor{}

\maketitle

\begin{abstract}
Data silos, mainly caused by privacy and interoperability, significantly constrain collaborations among different organizations with similar data for the same purpose. Distributed learning based on divide-and-conquer provides a promising way to settle the data silos, but it suffers from several challenges, including autonomy, privacy guarantees, and the necessity of collaborations. This paper focuses on developing an adaptive distributed kernel ridge regression (AdaDKRR) by taking autonomy in parameter selection, privacy in communicating non-sensitive information, and the necessity of collaborations in performance improvement into account. We provide both solid theoretical verification and comprehensive experiments for AdaDKRR to demonstrate its feasibility and effectiveness. Theoretically, we prove that under some mild conditions, AdaDKRR performs similarly to running the optimal learning algorithms on the whole data, verifying the necessity of collaborations and showing that no other distributed learning scheme can essentially beat AdaDKRR under the same conditions. Numerically, we test AdaDKRR on both toy simulations and two real-world applications to show that AdaDKRR is superior to other existing distributed learning schemes. All these results show that AdaDKRR is a feasible scheme to defend against data silos, which are highly desired in numerous application regions such as intelligent decision-making, pricing forecasting, and performance prediction for products.
\end{abstract}


\section{Introduction}
Big data has made a profound impact on people's decision-making, consumption patterns, and ways of life  \citep{davenport2012big,tambe2014big}, with many individuals now making decisions based on analyzing data rather than consulting experts; shopping online based on historical sales data rather than going to physical stores; gaining insights into consumer behaviors and preferences based on the consumption data rather than language communications. With the help of big data, organizations can identify patterns, trends, and correlations that may not be apparent in data of small size, which leads to more accurate predictions, better understanding of behaviors, and improved operational efficiencies.

However, data privacy and security \citep{jain2016big,li2017anonymizing} have garnered widespread attention, inevitably resulting in the so-called data silos, meaning that large-scale data distributed across numerous organizations cannot be centrally accessed, that is, organizations can only use their own local data but cannot obtain relevant data from elsewhere. For example, a large amount of medical data are stored in fragmented forms in different medical institutions but cannot be effectively aggregated; massive amounts of operational data are distributed among various companies but cannot be centrally accessed; and numerous consumer behavior data are collected by different platforms but cannot become public resources due to privacy factors. Data silos is a significant challenge \citep{fan2014challenges} for the use of big data, requiring ingenious multi-party collaboration methods to increase the willingness of data holders to cooperate and improve their efficiency of data analysis without leaking their sensitive information. Designing and developing feasible methods to avoid data silos is a recent focus of machine learning, which not only determines the role that machine learning plays in the era of big data but also guides the future direction of machine learning development.

\begin{figure*}[t]
    \centering
    \subfigcapskip=-2pt
	\subfigure[Flow of training]{\includegraphics[width=6.7cm,height=4.3cm]{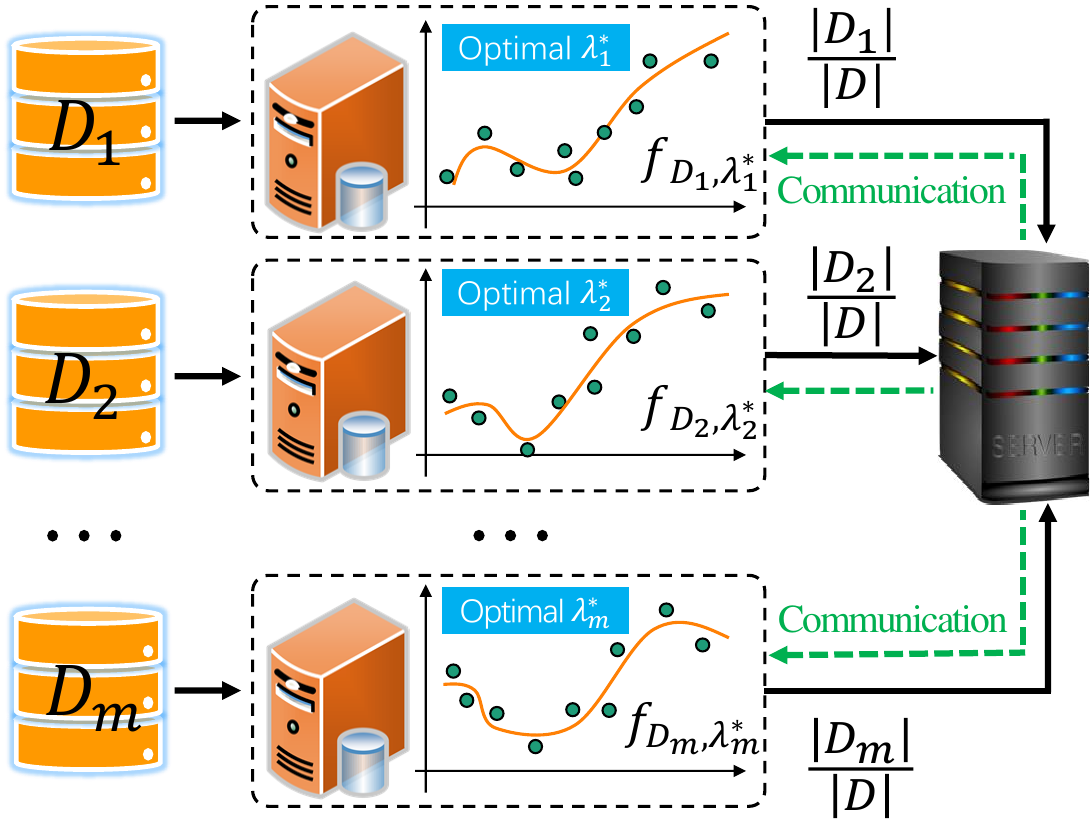}}\hspace{0.4in}
    \subfigure[Flow of testing]{\includegraphics[width=6.7cm,height=4.3cm]{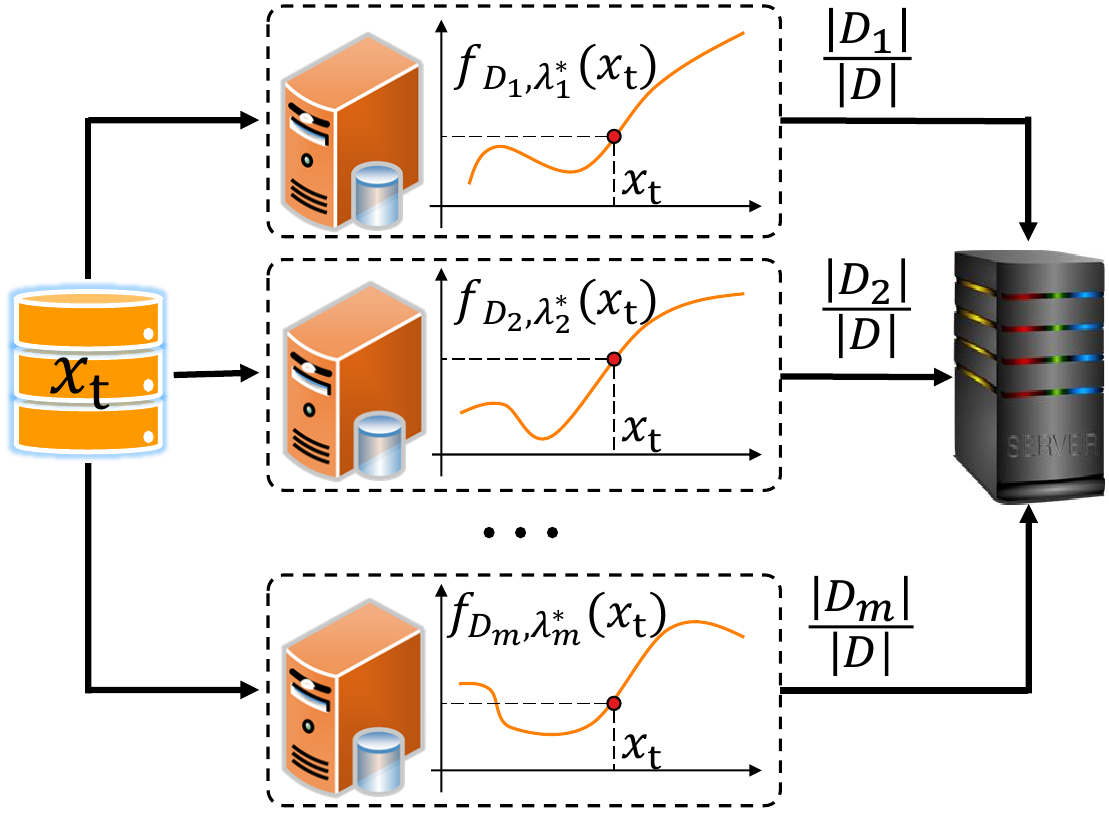}}
	\caption{Training and testing flows of DKRR}
\label{Fig:Flow1}\vspace{-0.2in}
\end{figure*}

Distributed learning \citep{balcan2012distributed,lea2013distributed} is a promising approach for addressing the data silos, as it enables multiple parties to collaborate and learn from each other's data without having to share the data themselves. As shown in Figure \ref{Fig:Flow1}, there are generally three ingredients in a distributed learning scheme. The first one is local processing, in which each local machine (party) runs a specific learning algorithm with its own algorithmic parameters and data to yield a local estimator. The second one is communication, where several useful but non-sensitive pieces of information are communicated with each other to improve the quality of local estimators. To protect data privacy, neither data nor information that could lead to data disclosure is permitted to be communicated. The last one is synthesization, in which all local estimators are communicated to the global machine to synthesize a global estimator. In this way, multiple parties can collaborate on solving problems that require access to the whole data from different sources while also addressing privacy concerns, as sensitive data are kept in their original locations and only non-sensitive information is shared among the parties involved.

Due to the success in circumventing the data silos, numerous distributed learning schemes with solid theoretical verification have been developed, including distributed linear regression \citep{zhang2013communication}, distributed online learning \citep{dekel2012optimal}, distributed conditional maximum entropy learning \citep{mcdonald2009efficient}, distributed kernel ridge regression \citep{zhang2015divide}, distributed local average regression \citep{chang2017divide}, distributed kernel-based gradient descent \citep{lin2018distributed}, distributed spectral algorithms \citep{mucke2018parallelizing}, distributed multi-penalty regularization algorithms \citep{guo2019distributed}, and distributed coefficient regularization algorithms \citep{shi2019distributed}. In particular, these algorithms have been proven to achieve optimal rates of generalization error bounds for their batch counterparts, as long as the algorithm parameters are properly selected and the number of local machines is not too large. However, how to choose appropriate algorithm parameters without sharing the data to achieve the theoretically optimal generalization performance of these distributed learning schemes is still open, because all the existing provable parameter selection strategies, such as the logarithmic mechanism for cross-validation \citep{liu2022enabling}, generalized cross-validation \citep{xu2019distributed}, and the discrepancy principle \citep{celisse2021analyzing}, need to access the whole data. This naturally raises the following problem:
 \begin{problem}\label{problem:parameter}
How to develop a feasible parameter selection strategy without communicating the individual data of local machines with each other to equip distributed learning to realize its theoretically optimal generalization performance and successfully circumvent the data silos?
% The problem is, however, that without sharing the data, it remains open on how to select appropriate algorithmic parameters in practice to realize the theoretically optimal generalization performances of these distributed learning schemes, since all the existing provable parameter selection strategies, like the logarithmic mechanism for cross-validation \citep{liu2022enabling}, generalized cross-validation \citep{xu2019distributed}, and the discrepancy principle \citep{celisse2021analyzing}, require access to the whole data.
 \end{problem}
In this paper, taking distributed kernel ridge regression as an example, we develop an adaptive parameter selection strategy based on communicating non-sensitive information to solve the above problem.
Our basic idea is to find a fixed basis, and each local machine computes an approximation of its derived rule (the relationship between the input and output) based on the basis and transfers the coefficients of the basis to the global machine. The global machine then synthesizes all the collected coefficients through a specific synthesis scheme and communicates the synthesized coefficients back to each local machine.
% Our basic idea is to find a set of fixed basis, with which each local machine can transfer its derived local rule (the relationship between the input and output) to the global machine by communicating the coefficients of the basis. Then, the global machine synthesizes all the collected coefficients through a specific synthesis scheme and communicates the synthesized coefficients back to each local machine.
In this way, each local machine obtains a good approximation of the global rule and uses this rule for cross-validation to determine its algorithm parameters. The road map of our approach is shown in Figure \ref{Fig:Roadmap}. Using the developed parameter selection strategy, we propose a novel adaptive distributed kernel ridge regression (AdaDKRR) to address the data silos.
% We conduct both theoretical analysis and numerical experiments to demonstrate the effectiveness and efficiency of AdaDKRR.
Our main contributions can be concluded as follows:
\begin{figure*}[t]
    \centering	 \subfigure{\includegraphics[width=14cm,height=1.6cm]{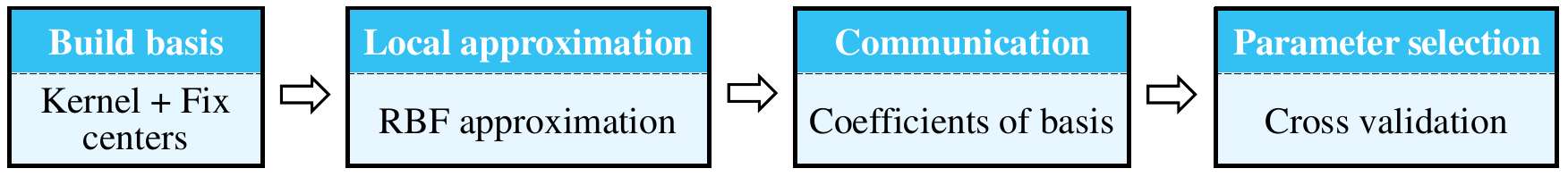}}
	\caption{Road map of the proposed parameter selection strategy} \label{Fig:Roadmap}
 \vspace{-0.1in}
\end{figure*}

$\bullet$ {\it Methodology novelty:} Since data stored in different local machines cannot be communicated, developing an adaptive parameter selection strategy based on local data to equip distributed learning is not easy. The main novelty of our approach is a nonparametric-to-parametric model transition method, which determines the algorithm parameters of distributed nonparametric learning schemes by communicating the coefficients of fixed basis functions without leaking any sensitive information about the local data. With such a novel design, we develop a provable and effective parameter selection strategy for distributed learning based on cross-validation
% and, consequently, AdaDKRR
to solve the data silos. As far as we know, this is the first attempt at designing provable parameter selection strategies for distributed learning to address the data silos.

$\bullet$ {\it Theoretical assessments:} Previous theoretical research \citep{zhang2015divide,lin2017distributed,mucke2018parallelizing,shi2019distributed} on distributed learning was carried out with three crucial assumptions: 1) the sizes of data in local machines are almost the same; 2) the parameters selected by different local machines are almost the same; 3) the number of local machines is not  {so large}. In this paper, we present a detailed theoretical analysis by considering the role of the synthesization strategy and removing the assumption of the same data size. Furthermore, we borrow the idea of low-discrepancy sequences \citep{dick2010digital} and the classical radial basis function approximation \citep{wendland2005approximate,rudi2015less} to prove the feasibility of the proposed parameter selection strategy and remove the above-mentioned same parameter assumption.  Finally, we provide an optimal generalization rate for AdaDKRR in the framework of statistical learning theory \citep{cucker2007learning,steinwart2008support}, which shows that if the number of local machines is not  {so large}, the performance of AdaDKRR is similar to running KRR on the whole data. This provides a solid theoretical verification for the feasibility of AdaDKRR to address the data silos.

$\bullet$ {\it Experimental verification:} We conduct both toy simulations and real-world data experiments to illustrate the excellent performance of AdaDKRR and verify our theoretical assertions. The numerical results show that AdaDKRR is robust to the number of basis functions, which makes the selection of the basis functions easy, thus obtaining satisfactory results. In addition, AdaDKRR shows stable and effective learning performances in parameter selection for distributed learning, regardless of whether the numbers of samples allocated to local machines are the same or not. We also apply AdaDKRR to two real-world data sets, including ones designed to help determine car prices and GPU acceleration models, to test its usability in practice.
% resulting in an easy selection of the basis for satisfactory results.

The rest of this paper is organized as follows. In the next section, we introduce the challenges, motivations, and some related work of parameter selection in distributed learning. In Section \ref{Sec:AdaDKRR}, we propose AdaDKRR and introduce some related properties. In Section \ref{Sec:theory}, we provide theoretical evidence of the effectiveness of the proposed adaptive parameter selection strategy and present an optimal generalization error bound for AdaDKRR. In Section \ref{sec:experiment}, we numerically analyze the learning performance of AdaDKRR in toy simulations and two real-world applications. Finally, we draw a simple conclusion. The proofs of all theoretical results and some other relevant information about AdaDKRR are postponed to the Appendix.

 \section{Challenges, Our approaches, and Related Work}\label{Sec.Related-work}
% In this section, after introducing DKRR, we highlight the challenges of parameter selection in DKRR and then present the motivations of our work. Finally, we compare our work with some related work.

Let $({\mathcal H}_K, \|\cdot\|_K)$ be a reproducing kernel Hilbert space (RKHS) induced by a
Mercer kernel $K$ \citep{cucker2007learning} on a compact input space ${\mathcal X}$. Suppose there is a data set $D_j=\{(x_{i,j},y_{i,j})\}_{i=1}^{|D_j|}\subset\mathcal X\times\mathcal Y$ stored in the $j$-th local machine with $1\leq j\leq m$ and $\mathcal Y\subseteq\mathbb R$ as the output space. Without loss of generality, we assume that there are no common samples of local machines, i.e. $D_j\cap D_{j'}=\varnothing$ for $j\neq j'$.
Distributed kernel ridge regression (DKRR)  with  regularization parameters $\vec{\lambda}:=\{\lambda_1,\dots,\lambda_m\}$  is defined by \citep{zhang2015divide,lin2017distributed}
\begin{equation}\label{DKRR}
\setlength{\abovedisplayskip}{0pt}
\setlength{\belowdisplayskip}{3pt}
         \overline{f}_{D,\vec{\lambda}}=\sum_{j=1}^m\frac{|D_j|}{|D|}f_{D_j,\lambda_j},
\end{equation}
where $\lambda_j>0$ is a regularization parameter for $j=1,\dots,m$, $D=\cup_{j=1}^mD_j$, $|D|$ denotes
the cardinality of the data set $D$, and the local estimator $f_{D_j,\lambda_j}$ is defined by
\begin{equation}\label{KRR-local}
\setlength{\abovedisplayskip}{2pt}
\setlength{\belowdisplayskip}{3pt}
    f_{D_j,\lambda_j} =\arg\min_{f\in \mathcal{H}_{K}}
    \left\{\frac{1}{|D_j|}\sum_{(x, y)\in D_j}(f(x)-y)^2+\lambda_j\|f\|^2_{K}\right\}.
\end{equation}
Therefore, in DKRR defined by \eqref{DKRR}, each local machine runs KRR \eqref{KRR-local} on its own data $D_j$ with a specific regularization parameter $\lambda_j$ to generate a local estimator, and the global machine synthesizes the global estimator $\overline{f}_{D,\vec{\lambda}}$ by using a weighted average based on data sizes. If $\lambda_j$ is given, then it does not need additional communications in DKRR to handle the data silos. However, how to choose $\lambda_j$ to optimize DKRR is important and difficult, as data are distributively stored across different local machines and cannot be shared.

\subsection{Challenges and road-map for parameter selection in distributed learning}
% Before presenting the detailed implementation of the adaptive parameter selection scheme, we first demonstrate the challenges of parameter selection of distributed learning.
 % The main challenge to do this is the contradiction between the over-fitting requirement of local estimators and the minimum-grasping property of existing parameter selection strategies explained as follows.

 \begin{figure*}[t]
    \centering	
    \subfigcapskip=-2pt
    \setlength{\abovecaptionskip}{-2pt}
    \subfigure[DKRR-machine 1] {\includegraphics[width=5cm,height=3.8cm]{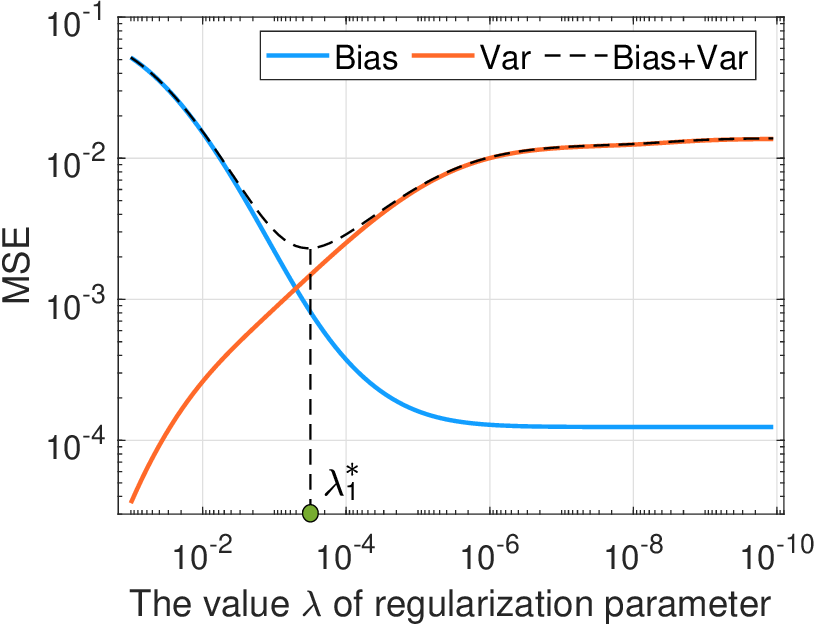}}\hspace{-0.05in}
    \subfigure[DKRR]{\includegraphics[width=5cm,height=3.8cm]{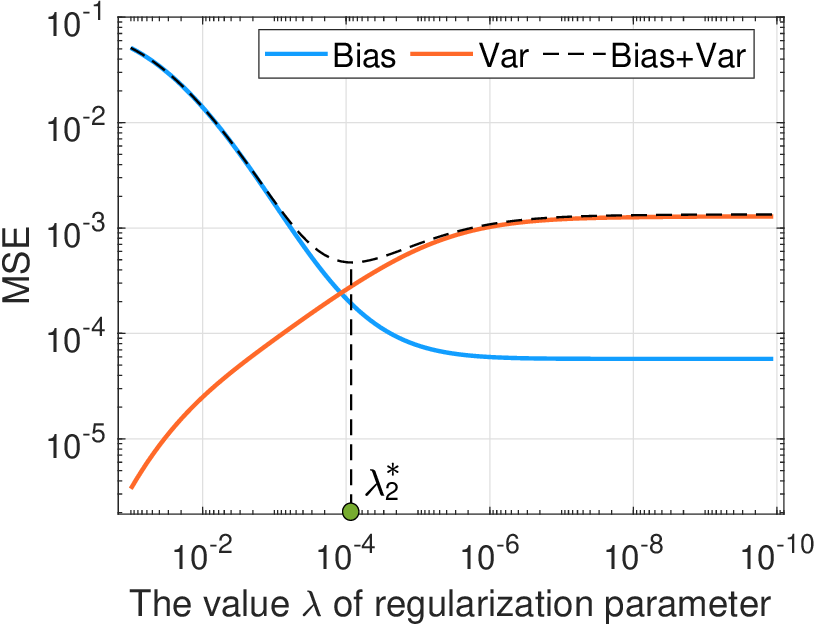}}\hspace{-0.05in}
    \subfigure[KRR]{\includegraphics[width=5cm,height=3.8cm]{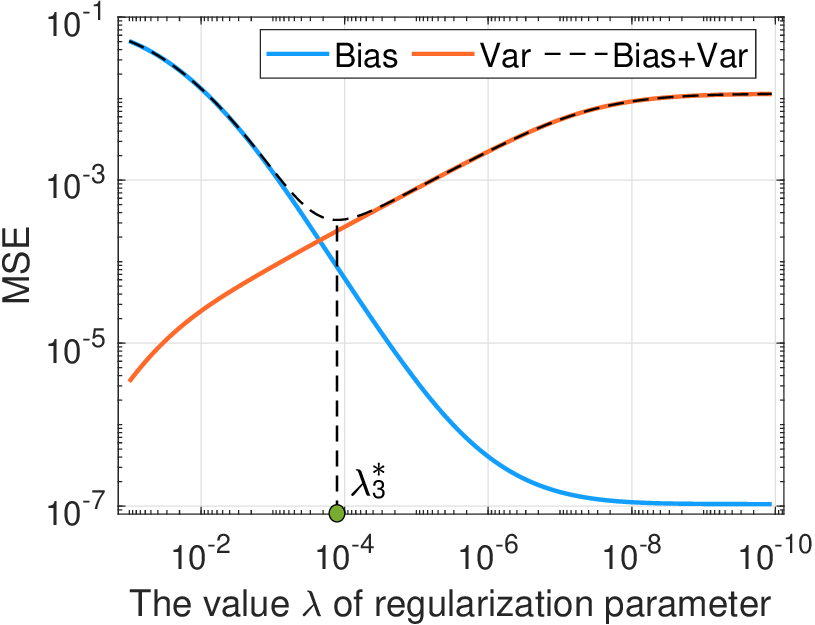}}
	\caption{Relationship between bias (variance) and regularization parameter values. Training data $\{(x_i,y_i)\}_{i=1}^{5000}$ are generated by drawing $\{x_i\}_{i=1}^{5000}$ i.i.d. according to the uniform distribution on $[0,1]^3$ and $y_i=g_1(x_i)+\varepsilon$, where $g_1(x)$ is defined by \eqref{g1} and $\varepsilon\sim\mathcal N(0,0.1^2)$; testing data $\{(x_i',y_i')\}_{i=1}^{1000}$ are generated similarly to the training data but with a promise that $y_i'=g_1(x_i')$. The training samples are uniformly distributed to $m$ local machines, and the number $m$ is set to 10. ``DKRR-machine1" represents running KRR on a local machine with a data subset of size $5000/m$.}\label{Fig:RelationBiasVarLambda}
 \vspace{-0.2in}
\end{figure*}

\begin{figure*}[t]
    \centering	
    \subfigcapskip=-2pt
    \subfigure[Bias] {\includegraphics[width=7cm,height=4.5cm]{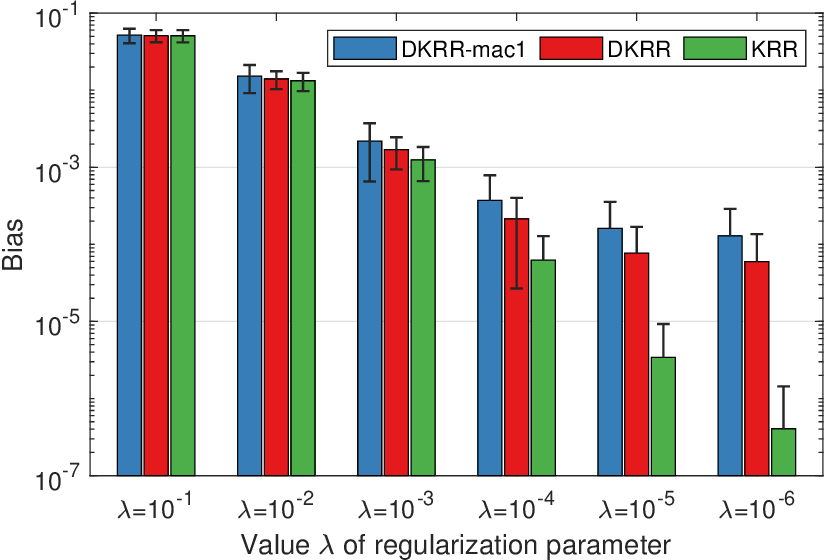}}
    \subfigure[Variance]{\includegraphics[width=7cm,height=4.45cm]{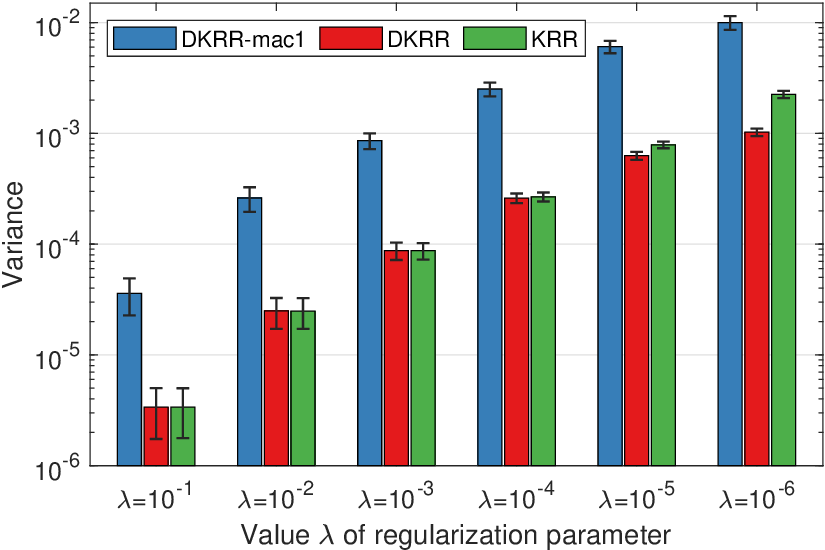}}
	\caption{Comparisons of bias and variance with different regularization parameter values. The data and simulation settings are the same as in Figure \ref{Fig:RelationBiasVarLambda}. }\label{Fig:ComparisonMSEtestbarDim10}
 \vspace{-0.3in}
\end{figure*}

Recalling that the introduction of the regularization term in \eqref{KRR-local} is to avoid the well-known over-fitting phenomenon \citep{cucker2007learning} that the derived estimator fits the training data well but fails to predict other queries, the optimal regularization parameter is frequently selected when the bias is close to the variance. However, as shown in Figure \ref{Fig:RelationBiasVarLambda},  if we choose the theoretically optimal regularization parameter based on its own data in each local machine, it is usually larger than the optimal parameter of the global estimator, i.e., $\lambda_1^*>\lambda_2^*\sim\lambda_3^*$, resulting in the derived global estimator under-fitting. This is not surprising, as the weighted average in the definition of \eqref{DKRR} helps to reduce the variance but has little influence on the bias, just as Figure \ref{Fig:ComparisonMSEtestbarDim10} purports to show.
% This is not surprising since the weighted average in the definition of \eqref{DKRR} helps to reduce the variance but does not affect the bias much, just as Figure \ref{Fig:ComparisonMSEtestbarDim10} purports to show.

Therefore, a smaller regularization parameter than the theoretically optimal one is required for each local machine based on its own data, leading to over-fitting for each local estimator. The weighted average in \eqref{DKRR} then succeeds in reducing the variance of DKRR and avoids over-fitting. The problem is, however, that each local machine only accesses its own data, making it difficult to determine the extent of over-fitting needed to optimize the performance of distributed learning. This refers to the over-fitting problem of parameter selection in distributed learning, and it is also the main challenge of our study.

% With communicating data, almost all existing parameter selection strategies, including the cross-validation \citep{gyorfi2002distribution,caponnetto2010cross}, balancing principle \citep{de2010adaptive,lu2020balancing}, discrepancy principle \citep{raskutti2014early,celisse2021analyzing}, are capable of searching the optimal rather than over-fitted regularization parameters for local estimators.

Generally speaking, there are two ways to  {settle} the over-fitting problem of parameter selection in distributed learning. One is to modify the existing parameter selection strategies, such as cross-validation \citep{gyorfi2002distribution,caponnetto2010cross}, the balancing principle \citep{de2010adaptive,lu2020balancing}, the discrepancy principle \citep{raskutti2014early,celisse2021analyzing}, and the Lepskii principle \citep{blanchard2019lepskii}, to force the local estimator in each local machine to over-fit their own data.
% A typical example is  the logarithmic mechanism \citep{liu2022enabling} that uses $\hat{\lambda}_j^{\log_{|D_j|}|D|}$   to shrink the regularization parameter $\hat{\lambda}_j$  that is selected to be optimal via $D_j$ only.
A typical example is the logarithmic mechanism \citep{liu2022enabling}, which uses $\lambda_j^{\log_{|D_j|}|D|}$ to reduce the regularization parameter $\lambda_j$ selected by $D_j$ alone as the optimal one. Recalling that it is unknown what the extent of over-fitting should be, it is difficult for this approach to get appropriate regularization parameters to achieve the theoretically optimal learning performance established in \citep{zhang2015divide,lin2017distributed} of DKRR. The other is to modify the target functions for parameter selection in each local machine so that the existing strategies can directly find the optimal regularization parameter for distributed learning. We adopt the latter in this paper since it is feasible for this purpose by designing delicate communication strategies. In particular, it is possible to find a good approximation of the global estimator $\overline{f}_{D,\vec{\lambda}}$ by communicating non-private information.

Our approach is motivated by four interesting observations. First, it can be seen in Figure \ref{Fig:RelationBiasVarLambda} that the optimal regularization parameter for the local estimator $f_{D_j}$ is not the optimal one for the global estimator. If we can find an approximation of the global estimator $\overline{f}_{D,\vec{\lambda}}$ and use this approximation instead of the local estimator $f_{D_j,\lambda_j}$ as the target of parameter selection in the $j$-th local machine, then it is not difficult to determine a nearly optimal regularization parameter for the global estimator $\overline{f}_{D,\vec{\lambda}}$ through the existing parameter selection strategies. Second, due to privacy, it is impossible to communicate the local estimator $f_{D_j,\lambda_j}$ directly since such a communication requires not only the coefficients of linear combinations of shifts of kernels but also the centers of the kernel that should be inputs of data $D_j$.
% Second, due to privacy, data cannot be communicated with each other, making it impossible to communicate the local estimator $f_{D_j,\lambda_j}$ directly since such communication requires not only the coefficients of linear combinations of shifts of kernels but also the centers of the kernel that should be inputs of data in $D_j$.
However, these local estimators can be well approximated by linear combinations of some fixed basis functions, which is a classical research topic in approximation theory \citep{narcowich2004scattered,wendland2005approximate,narcowich2006sobolev}. Third, the well-developed sampling approaches including Monte Carlo sampling and Quasi-Monte Carlo sampling  \citep{dick2010digital,leobacher2014introduction} introduced several low-discrepancy sequences, such as Sobol sequences, Niederreiter sequences, and Halton sequences, to improve the efficiency of the above approximation. Based on this, each local machine can generate the same centers of the kernel to establish a set of fixed basis functions, thus realizing the communication of functions by transmitting coefficients. Finally, though data cannot be communicated, some other non-private information, such as predicted values of queries, gradients, and coefficients of some basis functions, is communicable in distributed learning \citep{li2014communication,lee2017communication,jordan2018communication}. According to the above four important observations, we design the road map for parameter selection in distributed learning, as shown in Figure \ref{Fig:Roadmap}.

As stated above, there are five crucial ingredients in our approach: basis functions generation, local approximation, communications, global approximation, and local parameter selection. For the first issue, we focus on searching low-discrepancy sequences \citep{dick2010digital,leobacher2014introduction} to form the centers of the kernel and then obtain a linear space spanned by these basis functions. For the second issue, we use the radial basis function approximation approach \citep{narcowich2004scattered,narcowich2006sobolev,rudi2015less} with the noise-free data $\left\{\left(x_{i,j},f_{D_j,\lambda_j}(x_{i,j})\right)\right\}$ to provide a local approximation of the local estimator. For the third issue, each local machine transmits the coefficients of its local approximation to the global machine without leaking any sensitive information about its own data. For the fourth issue, the global machine synthesizes these coefficients by weighted average like \eqref{DKRR} and transmits the synthesized coefficients back to all local machines. For the last issue, each local machine executes a specific parameter selection strategy to determine the regularization parameter of the global approximation. Noting that besides the coefficients of some fixed basis functions, the sensitive information of the data in local machines is not communicated, which implies that the proposed approach provides a feasible scheme to settle the data silos.
% Since the local approximation is derived based on $\{x_{i,j},f_{D_j,\lambda_j}(x_{i,j})$, it may approximate the local estimator well, provided there are enough basis functions. Then, the global approximation is good estimator of the global estimator \eqref{DKRR}. Under this circumstance, selecting parameters for the global approximation is near to selecting parameters for the global estimator \eqref{DKRR}.

\subsection{Related work}
Since data silos caused by a combination of data privacy and interoperability impede the effective integration and management of data, it is highly desirable to develop feasible machine learning schemes to settle them and sufficiently explore the value of big data.
% Federated learning \citep{li2020federated}, given a pre-training model that all data holders know, aiming at collaborative training through multiple rounds of communications of non-sensitive information of the data holders to aggregate a golden model, is a popular approach to handling the data silos.
Federated learning \citep{li2020federated} is a popular approach to handling the data silos. It starts with a pre-training model that all data holders know and aims at collaborative training through multiple rounds of communications of non-sensitive information from the data holders to aggregate a golden model.
Although it has been numerically verified that federated learning is excellent in some specific application areas \citep{tuor2021asynchronous,li2022federated}, the exploration of pre-training models and multiple rounds of communications leads to essential weaknesses of the current defense against privacy attacks, such as data poisoning, model poisoning, and inference attacks \citep{li2020federated,lyu2020threats}.
More importantly, the lack of solid theoretical verifications restricts the use of federated learning in high-risk areas such as natural disaster prediction, financial market prediction, medical diagnosis prediction, and crime prediction.
% Federated learning \citep{li2020federated}, given a pre-training model known by all data holders, aiming to collaboratively training via multi-turn communications of non-sensitive information of the data holders to aggregate a golden model, is a popular approach to settle the data silos.
% Though it has been numerically verified that federated learning performs excellently in some specific application ranges \citep{tuor2021asynchronous,li2022federated}, the exploration of pre-training model and multi-turn communications lead to essential weakness of current defense for privacy attacks such as data poisoning, model poisoning, and inference attacks \citep{li2020federated,lyu2020threats}.

Theoretically, nonparametric distributed learning based on a divide-and-conquer strategy \citep{zhang2015divide,zhou2020differentially} is a more promising approach for addressing the data silos. As shown in Figure \ref{Fig:Flow1}, it does not need a pre-training model or multiple rounds of communications. Furthermore, solid theoretical verification has been established for numerous distributed learning schemes, including DKRR \citep{zhang2015divide,lin2017distributed}, distributed gradient descents \citep{lin2018distributed}, and distributed spectral algorithms \citep{guo2017learning,mucke2018parallelizing}, in the sense that such a distributed learning scheme performs almost the same as running the corresponding algorithms on the whole data under some conditions.
These interesting results seem to show that distributed learning can successfully address the data silos while realizing the benefits of big data without communicating sensitive information about the data. However,  all these exciting theoretical results are based on the assumption of proper selection of the algorithm (hyper-)parameters for distributed learning, which is challenging in reality if the data cannot be shared.
% These interesting results seem to show that distributed learning, without communicating sensitive information about data, can realize the advantages of big data by successfully addressing the data silos.
% However, all these exciting theoretical results are based on the assumption that the algorithm (hyper-)parameters of distributed learning are properly selected, which is difficult in practice if the data cannot be communicated.
This is the main reason why nonparametric distributed learning has not been practically used for settling the data silos, though its design flow is very suitable for this purpose.

As an open question in numerous papers \citep{zhang2015divide,lin2017distributed,mucke2018parallelizing,zhao2019debiasing}, the parameter selection of distributed learning has already been noticed by \citep{xu2019distributed} and \citep{liu2022enabling}.
In particular, \cite{xu2019distributed} proposed a distributed generalized cross-validation (DGCV) for DKRR and provided some solid theoretical analysis. It should be noted that the proposed DGCV essentially requires the communication of data, making it suffer from the data silos. \cite{liu2022enabling} proposed a logarithmic mechanism to force the over-fitting of local estimators without communicating sensitive information about local data and theoretically analyzed the efficacy of the logarithmic mechanism. However, their theoretical results are based on the assumption that the optimal parameter is algebraic with respect to the data size, which is difficult to verify in practice.

% Compared with all these related works, the main novelty of our study is that we propose an adaptive parameter selection strategy to equip non-parametric distributed learning schemes to settle the data silos. It should be highlighted that our proposed approach only needs two rounds of communications of non-sensitive information. We provide the optimality guarantee in theory and the feasibility evidence in applications.
Compared with all these related works, our main novelty is to propose an adaptive parameter selection strategy to equip non-parametric distributed learning schemes and thus settle the data silos. It should be highlighted that our proposed approach only needs two rounds of communications of non-sensitive information. We provide the optimality guarantee in theory and the feasibility evidence in applications.

 \section{Adaptive Distributed Kernel Ridge Regression}\label{Sec:AdaDKRR}

In this section, we propose an adaptive parameter selection strategy for distributed kernel ridge regression, which is named AdaDKRR, to address the data silos. As discussed in the

\begin{breakablealgorithm}
	\caption{\quad  AdaDKRR with hold-out}
	\label{Algorithm:AdaDKRR:1}
  \begin{algorithmic}[1]
 \Statex  \hspace{-0.3in} \textbf{Input:} Training data subset $D_j=\{(x_{ij},y_{ij})\}_{i=1}^{|D_j|}$ with $x_{ij}\in\mathcal X$ and $|y_{ij}|\leq M$ stored in the $j$-th local machine for $j=1,\cdots,m$, a candidate set of the regularization parameter $\Lambda=\{\lambda_\ell\}_{\ell=1}^L$, and a query point $x$. Divide $D_j=\{(x_{ij},y_{ij})\}_{i=1}^{|D_j|}$ into training and validation sets, and denote them as $D_{j}^{tr}$ and $D_{j}^{val}$, respectively.

 \State Local machines: given $\lambda_\ell\in\Lambda$ and $j$,  run KRR with data $D_{j}^{tr}$ to obtain a local estimator
\begin{equation}\label{KRR-local-training}
\setlength{\abovedisplayskip}{0pt}
\setlength{\belowdisplayskip}{3pt}
    f_{D_j^{tr},\lambda_\ell} =\arg\min_{f\in \mathcal{H}_{K}}
    \left\{\frac{1}{|D_j^{tr}|}\sum_{(x, y)\in D_j^{tr}}(f(x)-y)^2+\lambda_\ell\|f\|^2_{K}\right\}.
\end{equation}
 \Comment{Local Processing}
 \State  Local machines: generate the same set of centers $\Xi_n=\{\xi_k\}_{k=1}^n$ and define a set of basis functions
 % $B_{n,K}=\{K_{\xi_k}\}_{k=1}^{n}$
 $B_{n,K}:=\left\{\sum_{k=1}^na_kK_{\xi_k}:a_k\in\mathbb R\right\}$
 with $K_{\xi}(x)=K(\xi,x)$. \Comment{Basis Generation}
\State Local machines:  for some $s\in\mathbb N$, generate a set of points $\{x_{i,j}^*\}_{i=1}^{s} \subseteq \mathcal X$ and
 define an approximation of $f_{D_j^{tr},\lambda_{\ell}}$ by running KRR on data $\left\{\left(x_{i,j}^*,f_{D_j^{tr},\lambda_{\ell}}(x^*_{i,j})\right)\right\}_{i=1}^{s}$, that is,
\begin{equation}\label{local-approximation}
\setlength{\abovedisplayskip}{0pt}
\setlength{\belowdisplayskip}{3pt}
     {f}^{local}_{D_j^{tr},\lambda_{\ell},n,\mu,s}=\arg\min_{f\in B_{n,K}}\frac1{s}\sum_{i=1}^{s}\left(f(x^*_{i,j})
     -f_{D_j^{tr},\lambda_{\ell}}(x^*_{i,j})\right)^2+\mu\|f\|_K^2
\end{equation}
for some $\mu>0$, and denote
$ {f}^{local}_{D_j^{tr},\lambda_{\ell},n,\mu,s}=\sum_{k=1}^na^{local}_{j,k,\ell}K_{\xi_k}.
$
\Comment{Local Approximation}
 \State Local machines: transmit the cofficient matrix $\left(a^{local}_{j,k,\ell}\right)_{k=1,\ell=1}^{n,L}$ to the global machine.
 \Statex \Comment{Communication(I)}
\State Global machine: synthesize the coefficients by
${a}^{global}_{k,\ell}=\sum_{j=1}^m\frac{|D^{tr}_j|}{|D^{tr}|}a^{local}_{j,k,\ell}$ and communicate  $\left(a_{k,\ell}^{global}\right)_{k=1,\ell=1}^{n,L}$ to each local machine. \Comment{Synthesization and Communication(II)}
\State Local machines: obtain a global approximation ${f}^{global}_{D^{tr}, {\lambda}_\ell,n,\mu,s}$ as
\begin{equation}\label{global approximation}
\setlength{\abovedisplayskip}{0pt}
\setlength{\belowdisplayskip}{3pt}
     {f}^{global}_{D^{tr}, {\lambda}_\ell,n,\mu,s}:=\sum_{k=1}^n {a}^{global}_{k,\ell}K_{\xi_k} = \sum_{j=1}^m\frac{|D_j^{tr}|}{|D^{tr}|}\sum_{k=1}^na^{local}_{j,k,\ell}K_{\xi_k}=\sum_{j=1}^m\frac{|D_j^{tr}|}{|D^{tr}|}{f}^{local}_{D_j^{tr},\lambda_{\ell},n,\mu,s}
\end{equation}
and define
\begin{equation}\label{def.optimal-parameter}
\setlength{\abovedisplayskip}{0pt}
\setlength{\belowdisplayskip}{3pt}
    \lambda_j^*=\arg\min_{\lambda_{ \ell}\in\Lambda}\frac1{|D_j^{val}|}\sum_{(x,y)\in D_j^{val}} \left(\pi_M{f}^{global}_{D^{tr}, {\lambda}_\ell,n,\mu,s}(x)-y\right)^2
\end{equation}
with
$\pi_Mf(x)=\mbox{sign}(f(x))\min\{|f(x)|,M\}.
$\Comment{Local Validation}
\State Local machines: calculate $\pi_M{f}^{global}_{D^{tr}, \lambda^*_j,n,\mu,s}(x)$ and communicate it to the global machine.
\Statex \Comment{Communication(III)}
\State Global machine: synthesize the AdaDKRR estimator as  \Comment{Global Estimator}
\begin{equation}\label{AdaDKRR}
\setlength{\abovedisplayskip}{0pt}
\setlength{\belowdisplayskip}{3pt}
          \overline{f}^{Ada}_{D,\vec{\lambda}^*}(x):=\overline{f}^{Ada}_{D,\vec{\lambda}^*,n,\mu,s}(x)=\sum_{j=1}^m\frac{|D_j|}{|D|}\pi_M{f}^{global}_{D^{tr}, \lambda^*_j,n,\mu,s}(x).
\end{equation}
% $$
%        f^{\mbox{\scriptsize{Ada}}}_{D,\vec{\lambda}^\ast}(D'):=\sum_{j=1}^m\frac{|D_j|}{|D|}\pi_M f_{D_j,\lambda_j^\ast}(D') =\sum_{j=1}^m\frac{|D_j|}{|D|}\pi_M \left(\mathbb{K}_{D',D_j}\vec{\alpha}_{D_j}^\ast\right).
% $$
% $$
%       \overline{f}^{\mbox{\scriptsize{Ada}}}_{D,\vec{\lambda}^\ast}(D'):=\sum_{j=1}^m\frac{|D_j|}{|D|}\pi_M f_{D_j,\lambda_j^\ast}(D').
% $$
\Statex \hspace{-0.3in} \textbf{Output:} The global estimator  $\overline{f}^{Ada}_{D,\vec{\lambda}^*}(x)$.
%    \EndProcedure
  \end{algorithmic}
\end{breakablealgorithm}

\noindent previous section, our approach includes five important ingredients: basis generation, local approximation, communications, global approximation, and parameter selection. To ease the description, we use the ``hold-out'' approach \citep{caponnetto2010cross} in each local machine to adaptively select the parameter, though our approach can be easily designed for other strategies. The detailed implementation of AdaDKRR is shown in Algorithm \ref{Algorithm:AdaDKRR:1}.

Compared with the classical DKRR \citep{zhang2015divide,lin2017distributed}, AdaDKRR presented in Algorithm \ref{Algorithm:AdaDKRR:1} requires five additional steps (Steps 2--6) that include basis generation, local approximation, global approximation, and two rounds of communications with $\mathcal O(mnL)$ communication complexity. Algorithm \ref{Algorithm:AdaDKRR:1} actually presents a feasible framework for selecting parameters of distributed learning, as the basis functions, local approximation, and global approximation are not unique. It should be highlighted that Algorithm \ref{Algorithm:AdaDKRR:1} uses the ``hold-out'' method in selecting the parameters, while our approach is also available for cross-validation \citep{gyorfi2002distribution}, which requires a random division of the training data $D_j$. We refer the readers to Algorithm \ref{Algorithm:AdaDKRRCV} in the Appendix for the detailed training and testing flows of the cross-validation version of AdaDKRR.

In the basis generation step (Step 2), we generate the same set of basis functions in all local machines so that the local estimators defined in \eqref{KRR-local-training} can be well approximated by linear combinations of these basis functions. Noting that the local estimators are smooth and in $\mathcal H_K$, numerous basis functions, such as polynomials, splines, and kernels, can approximate them well from the viewpoint of approximation theory \citep{wendland2005approximate}. Since we have already obtained a kernel $K$, we use the kernel to build up the basis functions, and then the problem boils down to selecting a suitable set of centers $\Xi_n:=\{\xi_k\}_{k=1}^n$ so that $\mbox{span}\{K_{\xi_k}\}$ can well approximate functions in $\mathcal H_K$. There are roughly two approaches to determining $\Xi_n$. One is to generate a set of fixed low sequences, such as Sobol sequences and Halton sequences, with the same size \citep{dick2010digital}. It can be found in \citep{dick2010digital} that the complexity of generating $n$ Sobol sequences (or Halton sequences) is $\mathcal O(n\log n)$. Furthermore, it can be found in \citep{dick2010digital,dick2011higher,Feng2021radial} that there are $c,\beta>0$ such that
\begin{equation}\label{ass-points}
   \sup_{\|f\|_K\leq 1,\|g\|_K\leq 1}\left|\int f(x)g(x)dP_u(x)-\frac1n\sum_{k=1}^nf(\xi_k)g(\xi_k)\right|\leq cn^{-\beta},
\end{equation}
where $P_u$ denotes a uniform distribution. The other method is to generate $n$ points (in a random manner according to a uniform distribution) in the global machine, and then the global machine transmits this set of points to all local machines. In this paper, we focus on the first method to reduce the cost of communications, though the second one is also feasible.

In the local approximation step (Step 3), we aim to finding a good approximation of the local estimator $f_{D_j^{tr},\lambda_\ell}$. The key is to select a suitable set of points $\{x_{i,j}^*\}_{j=1}^s$ and a suitable parameter $\mu$ so that the solution to \eqref{local-approximation} can well approximate $f_{D_j^{tr},\lambda_\ell}$. Since there are already two point sets, $\{x_{i,j}\}_{i=1}^{|D_j^{tr}|}$ and $\{\xi_k\}_{k=1}^n$,
we can select one of them as $\{x_{i,j}^*\}_{j=1}^s$.
% we can select $\{x_{i,j}^*\}_{j=1}^s$ as either the set of input samples $\{x_{i,j}\}_{i=1}^{|D_j^{tr}|}$ or the set of centers $\Xi_n=\{\xi_k\}_{k=1}^n$.  In this paper, we use the former, but the latter is also sound due to the nice theory of \citep{wendland2005approximate} which illustrates that the solution to \eqref{local-approximation} is a good approximation of the local estimator.
% we can select either the set of input samples $\{x_{i,j}\}_{i=1}^{|D_j^{tr}|}$ or the set of centers $\Xi_n=\{\xi_k\}_{k=1}^n$ as $\{x_{i,j}^*\}_{j=1}^s$.
In this paper, we use the former, but choosing the latter is also reasonable because the solution to \eqref{local-approximation} is a good approximation of the local estimator \citep{wendland2005approximate}. Noting $s=|D_j^{tr}|$, we write
 $f^{local}_{D_j^{tr},\lambda_\ell,n,\mu,s}$ as
  $f^{local}_{D_j^{tr},\lambda_\ell,n,\mu}$.
Recalling the idea of Nystr\"{om} regularization \citep{rudi2015less,sun2021nystr} and regarding \eqref{local-approximation} as a Nystr\"{om} regularization scheme with the noise-free data $\left\{\left(x_{i,j},f_{D_j^{tr},\lambda_\ell}(x_{i,j})\right)\right\}_{i=1}^{|D_j^{tr}|}$,
we obtain from \eqref{local-approximation} that
$
    {f}^{local}_{D_j^{tr},\lambda_\ell,n,\mu}(\cdot)=\sum_{k=1}^{n} \alpha^{local}_{j,k,\ell} K_{\xi_k}(\cdot),
$
where
% \begin{equation}
% \vec{\alpha}_{j,\ell}^{local}:=\left(\alpha_{j,1,\ell}^{local},\dots,\alpha_{j,n,\ell}^{local}\right)^T = \left(\mathbb K_{n|D_j|}^T\mathbb K_{n|D_j|}+\lambda_\ell n\mathbb K_{|D_j||D_j|}\right)^\dagger \mathbb K_{n|D_j|}^Tf_{D_j,\lambda_\ell}^T,
% \end{equation}
\begin{equation}\label{Analytic-solution1}
\vec{a}_{j,\ell}^{local} := \left(a_{j,1,\ell}^{local}, \cdots, a_{j,n,\ell}^{local}\right)^T \hspace{-0.03in}= \hspace{-0.03in}\left(\mathbb{K}_{\left|D_{j}^{tr}\right|,n}^T \mathbb{K}_{\left|D_{j}^{tr}\right|,n} + \mu \left|D_{j}^{tr}\right| \mathbb{K}_{n,n} \right)^\dag \mathbb{K}_{\left|D_{j}^{tr}\right|,n}^T  {\vec{f}}_{D_j^{tr},\lambda_\ell},
\end{equation}
$A^\dagger$ and $ A^T$ denote the Moore-Penrose pseudo-inverse and transpose of a matrix $A$, respectively, $\left(\mathbb K_{|D_j^{tr}|,n}\right)_{i,k}=K(x_{i,j},\xi_k)$, $(\mathbb K_{n,n})_{k,k'}=K(\xi_k,\xi_{k'})$, and $\vec{f}_{D_j^{tr},\lambda_\ell}=\left(f_{D_j^{tr},\lambda_\ell}(x_{1,j}),\dots,\right.$ $\left.f_{D_j^{tr},\lambda_\ell}(x_{\left|D_{j}^{tr}\right|,j})\right)^T$.
Therefore, it requires {$\mathcal O\left(|D_j^{tr}|n^2+n^3\right)$} floating computations to derive the local approximation. Since $\left\{\left(x_{i,j},f_{D_j^{tr},\lambda_{\ell}}(x_{i,j})\right)\right\}_{i=1}^{|D_j^{tr}|}$ is noise-free, the parameter $\mu$ in \eqref{local-approximation} is introduced to
overcome the ill-conditionness of the linear least problems and thus can be set to be
small (e.g., $\mu=10^{-4}$).

\begin{figure*}[t]
    \centering	 \subfigure{\includegraphics[width=15.2cm,height=12.8cm]{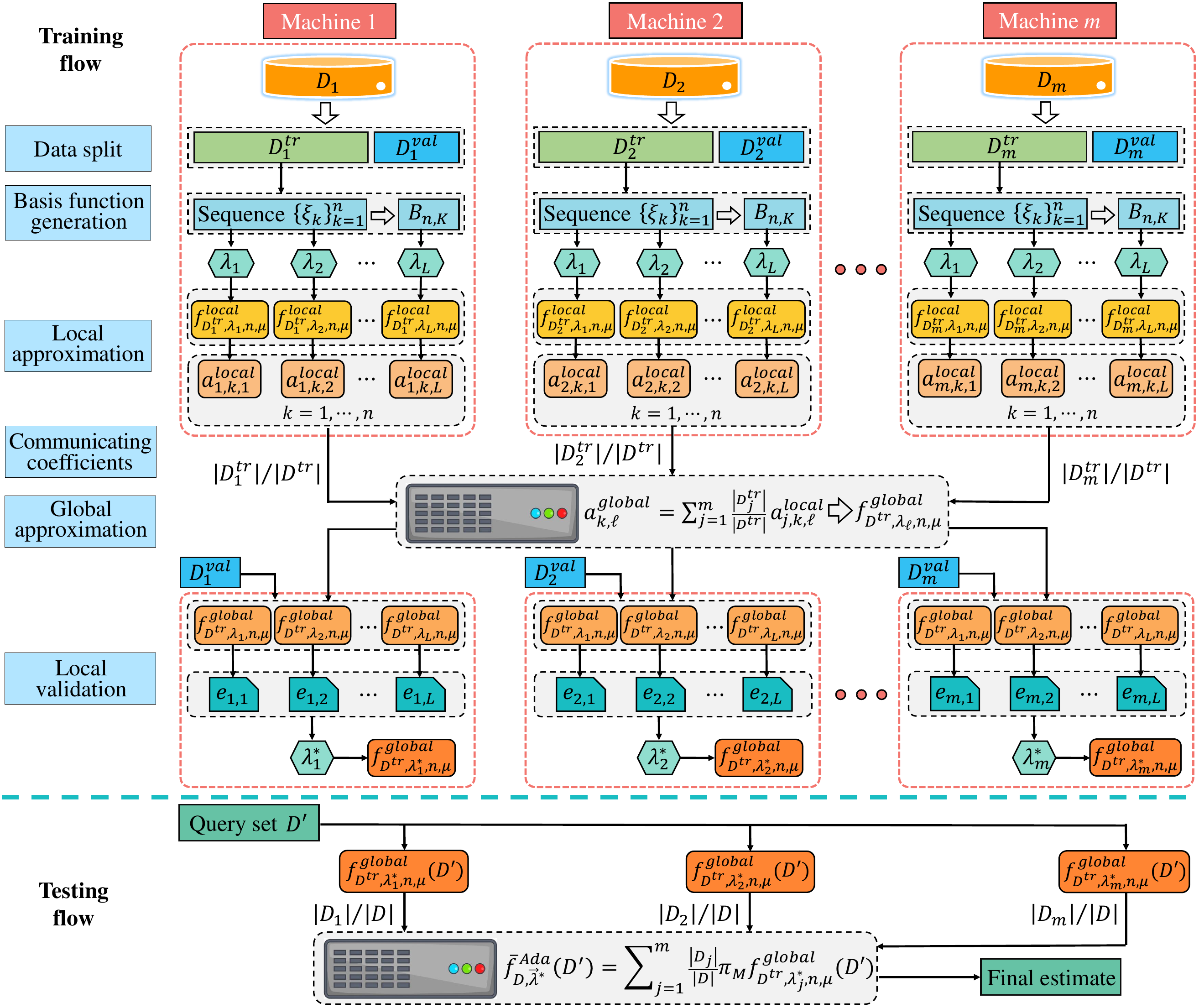}}
	\caption{Training and testing flows of the proposed method} \label{Fig:TrainTestFlows}
 \vspace{-0.2in}
\end{figure*}

In the global approximation step (Step 6), the global approximation is obtained through a weighted average. The optimal parameters of local machines are then searched for the global approximation via the validation set. If $f^{local}_{D_j^{tr},\lambda_\ell,n,\mu}$ is a good approximation of $f_{D_j^{tr},\lambda_\ell}$, then $f^{global}_{D^{tr},\lambda_\ell,n,\mu}$ is a good approximation of the global estimator $\overline{f}_{D^{tr},\lambda_\ell}$
% $f_{D^{tr},\lambda_\ell}$
defined by \eqref{DKRR}. Therefore, the optimal parameters selected for the global approximation are close to those of the global estimator. It should be noted that introducing the truncation operator $\pi_M$ in parameter selection is to ease the theoretical analysis and does not require additional computation. It requires  $\mathcal O\left(|D_j^{val}|nL\right)$ floating computations in this step. The flows of AdaDKRR adopted in this paper can be found in Figure \ref{Fig:TrainTestFlows}.

% As described in Algorithm \ref{Algorithm:AdaDKRR:1}, there are totally nine steps including three times of communications in AdaDKRR. The first step concerns the basis generating, which requires to generate a set of centers of the kernel. It should be
% mentioned that this step can be done via two methods.  In the second step, we conduct a data division of the training set.

%  Step 3 and Step 4 focus  on developing a  local approximation of the local estimator so that communicating functions can be reduced to communicating the coefficients of basis. It requires $\mathcal O(|D_j^{tr}|^3+|D_j^{tr}|n^2)$ float computing in these two step.
%  Furthermore, $\{x_{i,j}^*\}_{i=1}^{|D_j^{tr}|}$ can be selected directly as $\{x_{i,j}\}_{i=1}^{|D_j^{tr}|}$ or   low discrepancy sequences \cite{dick2010digital}.
%  Step 8 aims to transmit the estimator to the local machine. In the implementation of the distributed learning, a query point $x$ should be given at first and  this step indeed transmits a real value $\pi_M{f}^{global}_{D^{tr}, \lambda^*_j,\mu}(x)$ to the global machine and thus needs $\mathcal O(m)$ complexity in communications.   The last step synthesizes the final global estimator via weighted averaging and  makes a prediction of the query
% point. In this way, it requires complexities  $\mathcal O(|D_j|^3+|D_j|n^2L)$ in computations and $\mathcal O(nLm)$ in communications.

\section{Theoretical Verifications}\label{Sec:theory}
In this section, we study the generalization performance of AdaDKRR defined by \eqref{AdaDKRR} in a framework of statistical learning theory \citep{cucker2007learning,steinwart2008support}, where samples in $D_j$ for $j=1,2,\dots,m$ are assumed to be independently and identically drawn according to an unknown joint distribution $\rho:=\rho(x,y)=\rho_X(x)\rho(y|x)$ with the marginal distribution $\rho_X$  and the conditional distribution $\rho(\cdot|x)$. The regression function $f_\rho(x)=E[y|X=x]$ minimizes the generalization error $ \mathcal E(f):=\int_{\mathcal Z}(f(x)-y)^2d\rho$ for $f\in L_{\rho_X}^2$, where $L^2_{\rho_{_X}}$ denotes the Hilbert space of $\rho_X$-square integrable functions on $\mathcal X$, with the norm denoted by $\|\cdot\|_\rho$. Therefore,
the purpose of learning is to obtain an estimator $f_D$ based on $D_j$ for $j=1,2,\dots,m$  to approximate the regression function $f_\rho$ without leaking the privacy information of $D_j$. In this way, the performance of the global estimator $\overline{f}^{Ada}_{D,\vec{\lambda}^*}$ is quantitatively measured by the excess generalization error
% The purpose of learning is to obtain an estimator $f_D$ based on $D_j$ for $j=1,2,\dots,m$ without leaking the privacy information of $D_j$ to approximate the regression function $f_\rho(x)=E[y|X=x]$ that minimizes the generalization error
% $ \mathcal E(f):=\int_{\mathcal Z}(f(x)-y)^2d\rho$ for $f\in L_{\rho_X}^2$, where
% $L^2_{\rho_{_X}}$ denotes the Hilbert space of $\rho_X$-square integrable
% functions on $\mathcal X$, with the norm denoted by $\|\cdot\|_\rho$.
% In this way, the purpose is quantitatively measured by the generalization error
\begin{equation}\label{equality-generalization}
  \mathcal E(\overline{f}^{Ada}_{D,\vec{\lambda}^*})-\mathcal E(f_\rho)=\left\| \overline{f}^{Ada}_{D,\vec{\lambda}^*}-f_\rho\right\|^2_\rho,
\end{equation}
which describes the relationship between the prediction error and the data size.

\subsection{Generalization error for DKRR}
Before presenting the generalization error analysis for AdaDKRR, we first study the theoretical assessments of DKRR, which have been made in \citep{zhang2015divide,lin2017distributed} to show that DKRR performs similarly to running KRR on the whole data stored in a large enough machine, provided $m$ is not so large, $|D_1|\sim\dots\sim|D_j|$, and the regularization parameters
$\lambda_1\sim\dots\sim\lambda_m$ are similar to the optimal regularization parameter of KRR with the whole data.
% $\lambda_1\sim\dots\sim\lambda_m$ are selected to be similar to the optimal regularization parameter of KRR with the whole data.

The restriction on the number of local machines is natural since it is impossible to derive satisfactory generalization error bounds for DKRR when $m=|D|$, i.e., there is only one sample in each local machine. However, the assumptions $|D_1|\sim\dots\sim|D_j|$ and $\lambda_1\sim\dots\sim\lambda_m$ are a little bit unreasonable. On the one hand, local agents attending to the distributed learning system frequently have different data sizes, making it unrealistic to assume that the data sizes of local machines are the same. On the other hand, it is difficult to develop a parameter selection strategy for local machines so that $\lambda_1\sim\dots\sim\lambda_m$ are similar to the optimal regularization parameter of KRR with the whole data, as local agents only access their own data.

Noticing these, we derive optimal generalization error bounds for DKRR without the assumptions $|D_1|\sim\dots\sim|D_j|$ and  $\lambda_1\sim\cdots\sim\lambda_m$ the same as the theoretically optimal one. For this purpose, we introduce several standard assumptions on the data $D_j$, regression function $f_\rho$, and kernel $K$. As shown in Algorithm \ref{Algorithm:AdaDKRR:1}, our first assumption is the boundedness assumption of the output.

\begin{assumption}\label{Assumption:bounded for output}
There exists a $M>0$ such that $|y|\leq M$ almost surely.
\end{assumption}

Assumption \ref{Assumption:bounded for output} is quite mild since we are always faced with finitely many data, whose outputs are naturally bounded.
% From the theoretical viewpoint, the above boundedness assumption can be easily extended to the well known sub-Gaussian assumption as shown in \citep{caponnetto2007optimal,blanchard2016convergence}. We focus on Assumption \ref{Assumption:bounded for output} for the sake of brevity.
It should be mentioned that Assumption \ref{Assumption:bounded for output} implies $\|f_\rho\|_{L^\infty}\leq M$ directly. To present the second assumption, we should introduce the integral operator $L_K$ on ${\mathcal H}_K$ (or $L_{\rho_X}^2$) given by
% deduced by the Mercer kernel $K: {\mathcal X}\times {\mathcal X}\rightarrow \mathcal R$,
$$
         L_Kf :=\int_{\mathcal X} K_x f(x)d\rho_X, \qquad f\in {\mathcal
          H}_K \quad (\mbox{or}\ f\in L_{\rho_X}^2).
$$
The following assumption shows the regularity of the regression function $f_\rho$.

\begin{assumption}\label{Assumption:regularity}
For some $r>0$, assume
\begin{equation}\label{regularitycondition}
         f_\rho=L_K^r h_\rho,~~{\rm for~some}  ~ h_\rho\in L_{\rho_X}^2,
\end{equation}
where $L_K^r$ denotes the $r$-th power of $L_K: L_{\rho_X}^2 \to
L_{\rho_X}^2$ as a compact and positive operator.
\end{assumption}

According to the no-free-lunch theory \citep[Chap.3]{gyorfi2002distribution}, it is impossible to derive a satisfactory rate for the excess generalization error if there is no restriction on the regression functions. Assumption \ref{Assumption:regularity} actually connects $f_\rho$ with the adopted kernel $K$, where the index $r$ in \eqref{regularitycondition} quantifies the relationship. Indeed, \eqref{regularitycondition} with $r=1/2$ implies $f_\rho\in\mathcal H_K$, $0<r<1/2$ implies $f_\rho\notin\mathcal H_K$, and $r>1/2$ implies that $f_\rho$ is in an RKHS generated by a smoother kernel than $K$.
% This assumption has been widely adopted in numerous papers \citep{caponnetto2007optimal,steinwart2009optimal,caponnetto2010cross,blanchard2016convergence,blanchard2019lepskii,lin2017distributed,lin2018distributed} to quantify the property of the regression functions.

Our third assumption is on the property of the kernel, measured by the effective dimension \citep{caponnetto2007optimal},
$$
        \mathcal{N}(\lambda)={\rm Tr}((\lambda I+L_K)^{-1}L_K),  \qquad \lambda>0.
$$

\begin{assumption}\label{Assumption:effective dimension}
 There exists some $s\in(0,1]$ such that
\begin{equation}\label{assumption on effect}
      \mathcal N(\lambda)\leq C_0\lambda^{-s},
\end{equation}
where $C_0\geq 1$ is  a constant independent of $\lambda$.
\end{assumption}

It is obvious that \eqref{assumption on effect} always holds with $s=0$ and $C_0=\kappa:=\sqrt{\sup_{x\in\mathcal X}K(x,x)}$. As discussed in \citep{fischer2020sobolev}, Assumption \ref{Assumption:effective dimension} is equivalent to the eigenvalue decay assumption employed in \citep{caponnetto2007optimal,steinwart2009optimal,zhang2015divide}. It quantifies the smoothness of the kernel and the structure of the marginal distribution $\rho_X$. For example, if $\rho_X$ is the uniform distribution on the unit cube in the $d$-dimensional space $\mathbb R^d$ (i.e., $\mathcal X=\mathbb I^d$), and $K$ is a Sobolev kernel of order $\tau>d/2$, then Assumption \ref{Assumption:effective dimension} holds with $s=\frac{d}{2\tau}$ \citep{steinwart2009optimal}. The above three assumptions have been widely used to analyze generalization errors for kernel-based learning algorithms \citep{blanchard2016convergence,chang2017distributed,dicker2017kernel,guo2017learning,guo2017learning1,lin2017distributed,lin2018distributed,mucke2018parallelizing,shi2019distributed,fischer2020sobolev,lin2020optimal,sun2021nystr}, and optimal rates of excess generalization error  for numerous learning algorithms have been established under these assumptions.

Under these well-developed assumptions, we provide the following theorem that DKRR can achieve the optimal rate of excess generalization error established for KRR with the whole data \citep{caponnetto2007optimal,lin2017distributed,fischer2020sobolev}, even when different local machines possess different data sizes.

\begin{theorem}\label{Theorem:DKRR-1}
 Under Assumption \ref{Assumption:bounded for output},  Assumption \ref{Assumption:regularity} with $\frac12\leq r\leq 1$, and Assumption \ref{Assumption:effective dimension} with $0<s\leq 1$, if
\begin{eqnarray}\label{choose-lambda-1}
    \lambda_j=C_1 \left\{\begin{array}{cc}
                |D|^{ -\frac{1}{2r+s}}, & \mbox{if}\quad  |D_j|\geq  |D|^{ \frac{ 1}{2r+s}}\log^4|D|,\\
                |D_j|^{-1}\log^4|D|,& \mbox{otherwise},
               \end{array}\right.
\end{eqnarray}
and
\begin{equation}\label{resriction-on-m-1}
        m\leq
        |D|^{\frac{s}{2r+s}}(\log |D|)^{-8r},
\end{equation}
then
\begin{equation}\label{Error-for-DKRR}
    E\left[\|\overline{f}_{D,\vec{\lambda}}-f_\rho\|_\rho^2\right]
    \leq C_2|D|^{-\frac{2r}{2r+s}},
\end{equation}
where $C_1$ and $C_2$ are constants independent of $|D|$ or $m$.
% , whose concrete values are given in the proof.
 \end{theorem}

Under Assumptions \ref{Assumption:bounded for output}--\ref{Assumption:effective dimension}, it can be found in \citep{caponnetto2007optimal} that the derived learning rates in \eqref{Error-for-DKRR} are optimal in the sense that there is a regression function $f^*_\rho$ satisfying the above three assumptions such that
$$
 E\left[\|\overline{f}_{D,\vec{\lambda}}-f_\rho^*\|_\rho^2\right]
    \geq C_3|D|^{-\frac{2r}{2r+s}}
$$
for a constant $C_3$ depending only on $r$ and $s$. Unlike the existing results on distributed learning  \citep{zhang2015divide,lin2017distributed,mucke2018parallelizing,lin2020optimal} that imposed strict restrictions on the data sizes of local machines, i.e., $|D_1|\sim|D_2|\sim\dots\sim|D_m|$, Theorem \ref{Theorem:DKRR-1} removes this condition since it is difficult to guarantee the same data size for all participants in the distributed learning system.
% Theorem \ref{Theorem:DKRR-1} removes this condition
% % by considering that
% since it is difficult to guarantee that all participants in the distributed learning system possess the same data size.
As a result, it requires completely different mechanisms \eqref{choose-lambda-1} to select the regularization parameter and stricter restriction on the number of local machines \eqref{resriction-on-m-1}.  The main reason for the stricter restriction on $m$ is that the distributed learning system accommodates local machines with little data, i.e., $|D_j|\leq |D|^{\frac1{2r+s}}\log^4|D|$. If we impose a qualification requirement that each participant in the distributed learning system has at least $|D|^{\frac1{2r+s}}\log^4|D|$ samples, then the restriction can be greatly relaxed, just as the following corollary shows.

\begin{corollary}\label{corollary:DKRR-1}
 Under Assumption \ref{Assumption:bounded for output},  Assumption \ref{Assumption:regularity} with $\frac12\leq r\leq 1$, and Assumption \ref{Assumption:effective dimension} with $0<s\leq 1$, if $|D_j|\geq  |D|^{ \frac{ 1}{2r+s}}\log^4|D|$,  $\lambda_j=C_1|D|^{ -\frac{1}{2r+s}}$ for all $j=1,2,\dots,m$,
and
\begin{equation}\label{resriction-on-m-2}
        m\leq
        |D|^{\frac{2r+s-1}{2r+s}}\log^{-4}|D| ,
\end{equation}
then
\begin{equation}\label{Error-for-DKRR-corollary}
    E\left[\|\overline{f}_{D,\vec{\lambda}}-f_\rho\|_\rho^2\right]
    \leq C_2|D|^{-\frac{2r}{2r+s}},
\end{equation}
where $C_1$ and $C_2$ are constants independent of $|D|$ or $m$.
 \end{corollary}

Theorem \ref{Theorem:DKRR-1} and Corollary \ref{corollary:DKRR-1} provide a baseline for the analysis of AdaDKRR in terms that the generalization error of AdaDKRR should be similar to \eqref{Error-for-DKRR}.

\subsection{Learning performance of AdaDKRR}
In this subsection, we study the theoretical behavior of AdaDKRR \eqref{AdaDKRR} by estimating its generalization error in the following theorem.

\begin{theorem}\label{Theorem:Generalization-error-AdaDKRR}
Under Assumption \ref{Assumption:bounded for output}, Assumption \ref{Assumption:regularity} with $1/2\leq r\leq 1$, and Assumption \ref{Assumption:effective dimension} with $0<s\leq 1$, if $\rho_X$ is a uniform distribution, $|D_j|\geq (8C_1^*(\log(1+\kappa)+2))^2|D|^{\frac{1}{2r+s}}\log^4|D|$, $\Lambda$ contains a $\bar{\lambda}\sim |D|^{-\frac{1}{2r+s}}$, and
\begin{equation}\label{restriction-on-m-3}
     m\leq \min\left\{|D|^{\frac{2r+s-1}{4r+2s}}\log^{-4}|D|, |D|^{\frac{s}{2r+s}}\log^{-1}L\right\},
\end{equation}
then for any
 $\mu\in \left[\left(8C_1^*\left(\log(1+\kappa)+2\right)\right)^2 \max_{j=1,\dots,m}\frac{\log^4|D_j^{tr}|}{|D_j^{tr}|},|D|^{-\frac1{2r+s}}\right]$ and
 % $\Xi_n$ satisfies \eqref{ass-points} for some $c,\beta>0$ with $n$ satisfying $\mu n^\beta\geq 2c$,
 $\Xi_n$ satisfying \eqref{ass-points} for some $c,\beta>0$ with $\mu n^\beta\geq 2c$,
 there holds
\begin{equation}\label{generalization-error-adadkrr}
     E\left[ \left\| \overline{f}^{Ada}_{D,\vec{\lambda^*}}-f_\rho\right\|_\rho^2 \right]
      \leq
  C  |D|^{-\frac{2r}{2r+s}},
\end{equation}
where $C,C_1^*$ are constants depending only on $\|h_\rho\|_\rho,M,r,C_0,c,$ and $\beta$.
\end{theorem}

Compared with Theorem \ref{Theorem:DKRR-1}, it can be found that AdaDKRR possesses the same generalization error bounds under some additional restrictions, implying that the proposed parameter selection strategy is optimal in the sense that no other strategies always perform better.
There are five additional restrictions that may prohibit the wide use of the proposed approach:
% It should be mentioned that there are five additional restrictions that may prohibit the wide use of the proposed approach:
(I) $m$ satisfies \eqref{restriction-on-m-3}; (II) $|D_j|\geq (8C_1^*(\log(1+\kappa)+2))^2|D|^{\frac{1}{2r+s}}\log^4|D|$; (III) $\Lambda$ contains a $\bar{\lambda}\sim |D|^{-\frac{1}{2r+s}}$; (IV) $\rho_X$ is a uniform distribution; (V)  $\mu\in \left(\left(8C_1^*(\log(1+\kappa)+2)\right)^2 \right.$ $\left. \max_{j=1,\dots,m}{(\log^4|D_j^{tr}|})/{|D_j^{tr}|},|D|^{-\frac1{2r+s}}\right)$ and $\Xi_n$ satisfies \eqref{ass-points} for some $c,\beta>0$ with $n$ satisfying $\mu n^\beta\geq 2c$.

Condition (I) is necessary since it is impossible to derive a satisfactory distributed learning estimator when each local machine has only one sample. Condition (II) presents a qualification requirement for the local machines participating in the distributed learning system, indicating that their data sizes should not be so small.
Condition (III) means that the candidate set $\Lambda$ should include the optimal parameter.
Noting \eqref{restriction-on-m-3}, the restriction on $m$ is logarithmic with respect to $L$, and we can set  $\Lambda=\{\lambda_k\}_{k=1}^L$ with $\lambda_k=q^{k}$ for {some $q\in(0,1)$} and $L\sim |D|$. Conditions (IV) and (V) are mainly due to setting $\{x_{i,j}^*\}_{i=1}^s$ to $\{x_{i,j}\}_{i=1}^{|D_j|}$ in the local approximation step (Step 3 in Algorithm \ref{Algorithm:AdaDKRR:1}). Therefore, we have to use the quadrature property \eqref{ass-points} of the
low-discrepancy property, which requires the samples to be drawn i.i.d. according to the uniform distribution. Furthermore, the well-conditioness of the local
approximation imposes a lower bound of $\mu$. Since $|D_j|\geq (8C_1^*(\log(1+\kappa)+2))^2|D|^{\frac{1}{2r+s}}\log^4|D|$, it is easy to check that
$(8C_1^*(\log(1+\kappa)+2))^2\frac{\log^4|D_j^{tr}|}{|D_j^{tr}|}\leq |D|^{-\frac1{2r+s}}$, and there are numerous feasible values for $\mu$. The restriction on $\mu$ is to theoretically verify the well-conditionness of the local approximation in the worst case. In practice, it can be set to $10^{-4}$ directly. It would also be interesting to set a suitable $\{x_{i,j}^*\}_{i=1}^s$ to remove or relax conditions (IV) and (V).

As shown in  Theorem \ref{Theorem:Generalization-error-AdaDKRR}, under some assumptions, we prove that AdaDKRR performs similarly to running the optimal learning algorithms on the whole data $D=\cup_{j=1}^mD_j$ without considering data privacy. Recalling in Algorithm \ref{Algorithm:AdaDKRR:1} that AdaDKRR only requires communicating non-sensitive information, it is thus a feasible strategy to address the data silos.

\section{Experimental Results}\label{sec:experiment}
In this section, we use the following parameter selection methods for distributed learning to conduct experiments on synthetic and real-world data sets:
% to verify the effectiveness of the proposed method as compared with the following two approaches of
\begin{itemize}
    \item [1)] On each local machine, the parameters are selected by cross-validation, and DKRR is executed with these selected parameters; we call this method DKRR with cross-validation (``DKRR" for short).
    \item [2)] On the $j$-th local machine, we first select parameters by cross-validation, and then transform the selected regularization parameter $\lambda_j$ by
    % $\lambda_j \leftarrow \min\left(\lambda_j^{\log{|D|}/\log{|D_j|}},1\right)$,
     $\lambda_j \leftarrow \lambda_j^{\log{|D|}/\log{|D_j|}}$,
   {and transform the selected kernel width $\sigma_j$ by $\sigma_j \leftarrow \sigma_j^{\log{|D|}/\log{|D_j|}}$ if the Gaussian kernel is used;}
    % where $N$ is the total number of training samples and $|D_j|$ is the number of training samples on the $j$-th local machine;
    DKRR is executed with these transformed parameters; we call this method DKRR with cross-validation and logarithmic transformation (``DKRRLog" for short).
    % \item [3)] DKRR is implemented with the proposed adaptive parameter selection method, in which the number of Sobol points is fixed;
    % % we call this method Adaptive DKRR with a fixed number of Sobol points (`AdaDKRR: n=\#' for short), where $\#$ represents the fixed number of Sobol points.
    % we denote this method as `AdaDKRR: $n=\#$' for short, where $\#$ represents the fixed number of Sobol points.
    % \item [4)] DKRR is implemented with the proposed adaptive parameter selection method, in which the number of Sobol points is the average number of training samples on local machines; we denote this method as `AdaDKRR: $n=|D|/m$' for short, where $m$ is the number of local machines.
    \item [3)] The proposed adaptive parameter selection method is applied to distributed learning and is denoted by ``AdaDKRR".
\end{itemize}

All the experiments are run on a desktop workstation equipped with an Intel(R) Core(TM) i9-10980XE 3.00 GHz CPU, 128 GB of RAM, and Windows 10. The results are recorded by averaging the results from multiple individual trials with the best parameters.\protect\footnotemark[1]

\footnotetext[1] {The MATLAB code, as well as the data sets, can be downloaded from https://github.com/18357710774/ AdaDKRR.}

\subsection{Synthetic Results}
In this part, the performance of the proposed method is verified by four simulations. The first one studies the influence of the number and type of center points for local approximation on generalization ability. The second one exhibits the robustness of AdaDKRR to the number of center points. The third simulation presents comparisons of the generalization ability of the three mentioned methods with changing the number of local machines, provided that all training samples are uniformly distributed to local machines. The last simulation focuses on comparisons of generalization ability for the three methods when the training samples are unevenly distributed on local machines.

% This part examines the performance of the proposed method with four simulations. The first one investigates the roles of the number and type of center points for local approximation in generalization ability. The second one exhibits the robustness of AdaDKRR to the number of center points. The third simulation presents comparisons on generalization ability with changing the number of local machines for the three mentioned methods, provided that all training samples are uniformly distributed to local machines. The last simulation focuses on comparisons of generalization ability for the three mentioned methods with non-uniform distributions of training samples on local machines.

Before carrying out experiments, we describe the generating process of the synthetic data and some important settings of the simulations. The inputs $\{x_i\}_{i=1}^N$ of training samples are independently drawn according to the uniform distribution on the (hyper-)cube $[0,1]^d$ with $d=3$ or $d=10$. The corresponding outputs $\{y_i\}_{i=1}^N$ are generated from the regression models $y_i=g_j(x_i)+\varepsilon$ with the Gaussian noise $\mathcal{N}(0,0.2)$ for $j=1,2$, where
\begin{equation}\label{g1}
g_1(x)=\left\{
\begin{array}{ll}
(1-\|x\|_2)^6(35\|x\|_2^2+18\|x\|_2+3) & \quad\mbox{if} ~ 0<\|x\|_2\leq 1, \\
0 & \quad \mbox{if} ~ \|x\|_2> 1,
\end{array}
\right.
\end{equation}
for the 3-dimensional data, and
\begin{equation}\label{g2}
g_2(x)=\left(\|x\|_2-1\right)\left(\|x\|_2-2\right)\left(\|x\|_2-3\right)
\end{equation}
for the 10-dimensional data. The generation of test sets $\{(x_i', y_i')\}_{i=1}^{N'}$ is similar to that of training sets, but it has the promise of $y_i'=g_j(x_i')$.

For the 3-dimensional data, we use the kernel function $K_1(x_1,x_2)=h(\|x_1-x_2\|_2)$ with
\begin{equation}\label{h}
h(r)=\left\{
\begin{array}{ll}
(1-r)^4(4r+1) & \quad\mbox{if} ~ 0<r \leq 1, \\
0 & \quad \mbox{if} ~ r> 1,
\end{array}
\right.
\end{equation}
and the regularization parameter $\lambda$ is chosen from the set $\{\frac{1}{2^q} | \frac{1}{2^q} \geq 10^{-10}, q=0,1,2,\cdots\}$.
For the 10-dimensional data, we use the Gaussian kernel $K_2(x_1, x_2)=\exp\left(-\frac{\|x_1 - x_2\|_2^2}{2\sigma^2}\right)$, the regularization parameter $\lambda$ is chosen from the set
$\{\frac{1}{3^q} | \frac{1}{3^q} \geq 10^{-10}, q=0,1,2,\cdots\}$, and the kernel width $\sigma$ is chosen from 10 values that are drawn in a logarithmic, equally spaced interval $[0.1, 10]$. In the simulations, we generate $10000$ samples for training and $1000$ samples for testing, and the regularization parameter $\mu$ for local approximation is fixed as $10^{-4}$.

% \begin{figure*}[t]
%     \centering
% 	\subfigure[Dim=3]{\includegraphics[width=7cm,height=5cm]{figs/Relation_testMSE_kappa_N10000D3}}\hspace{0.3in}
%     \subfigure[Dim=10]{\includegraphics[width=7cm,height=5cm]{figs/Relation_testMSE_kappa_N10000D10}}
% 	\caption{The relation between test MSE and the number of Sobol points for fixed number of local machines.}
% \label{MSEtest}
% \end{figure*}

\begin{figure*}[t]
	\centering
        \subfigcapskip=-3pt
	\subfigure{
        \rotatebox{90}{\scriptsize{~~~~~~~~~~~~~Dim=3}}
		\includegraphics[width=3.65cm, height=3.15cm]{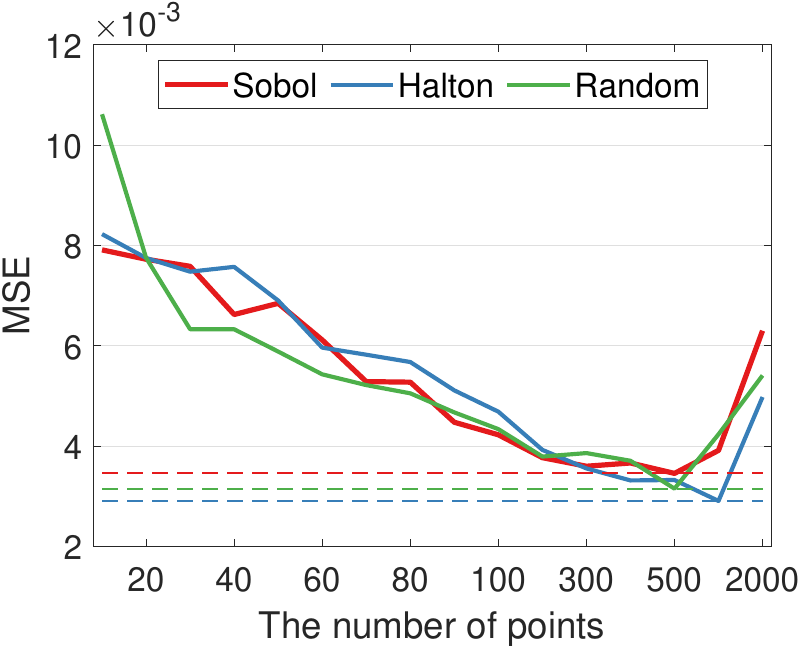}
	}\hspace{-0.145in}
		\subfigure{
		\includegraphics[width=3.65cm, height=3.15cm]{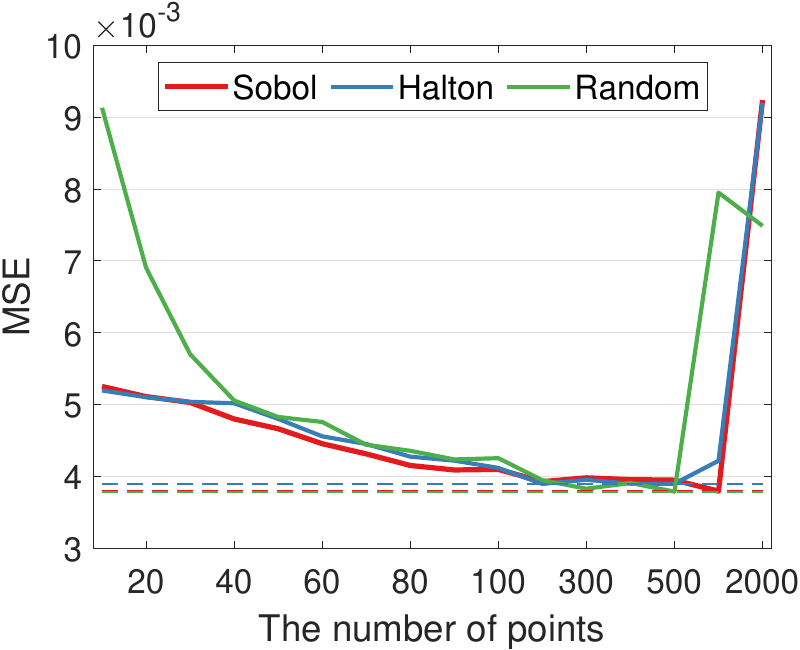}
	}\hspace{-0.145in}
		\subfigure{
		\includegraphics[width=3.65cm, height=3.15cm]{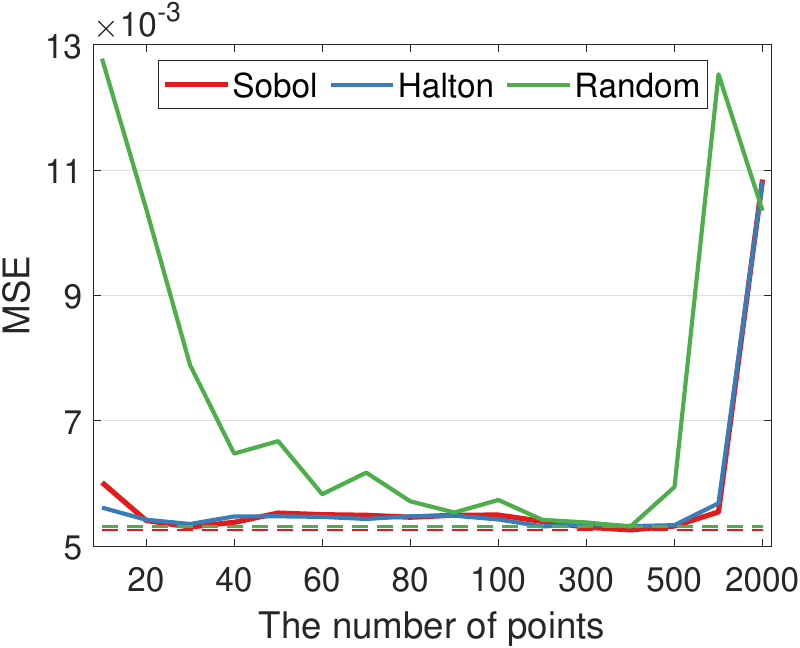}
	}\hspace{-0.145in}
		\subfigure{
		\includegraphics[width=3.65cm, height=3.15cm]{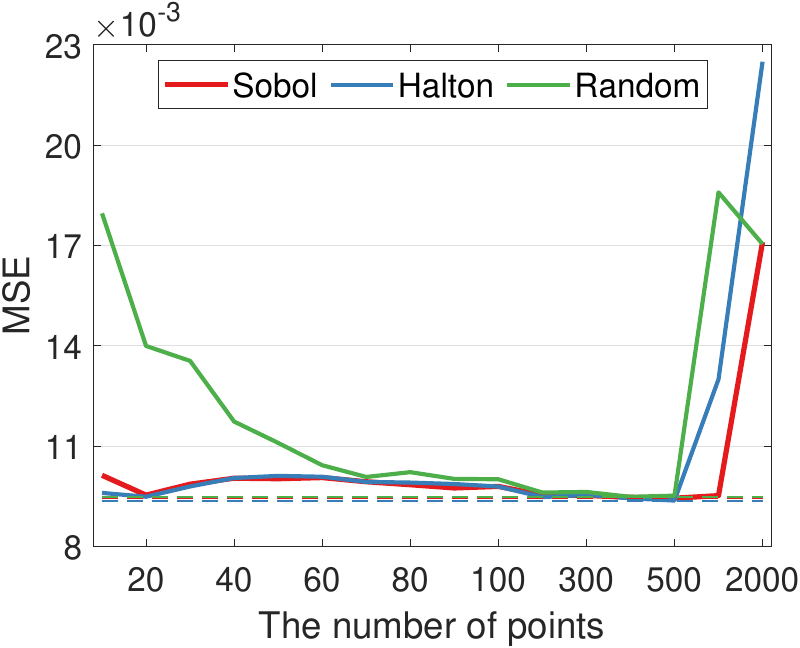}
	}\\
	\vspace{-3mm}
	\setcounter{subfigure}{0}
    \subfigure[$m=20$]{
	\rotatebox{90}{\scriptsize{~~~~~~~~~~~~~Dim=10}}
		\includegraphics[width=3.65cm, height=3.15cm]{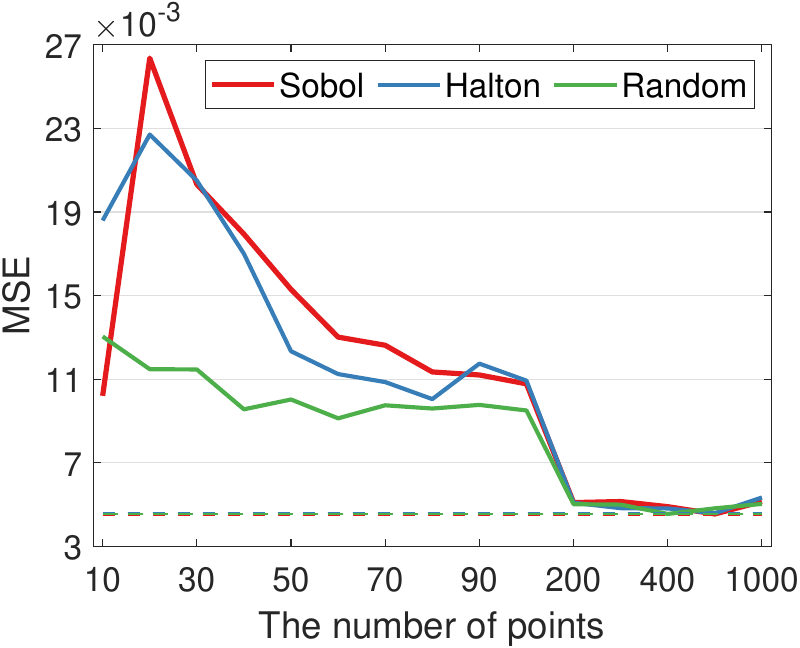}
	}\hspace{-0.145in}
	\subfigure[$m=40$]{
		\includegraphics[width=3.65cm, height=3.15cm]{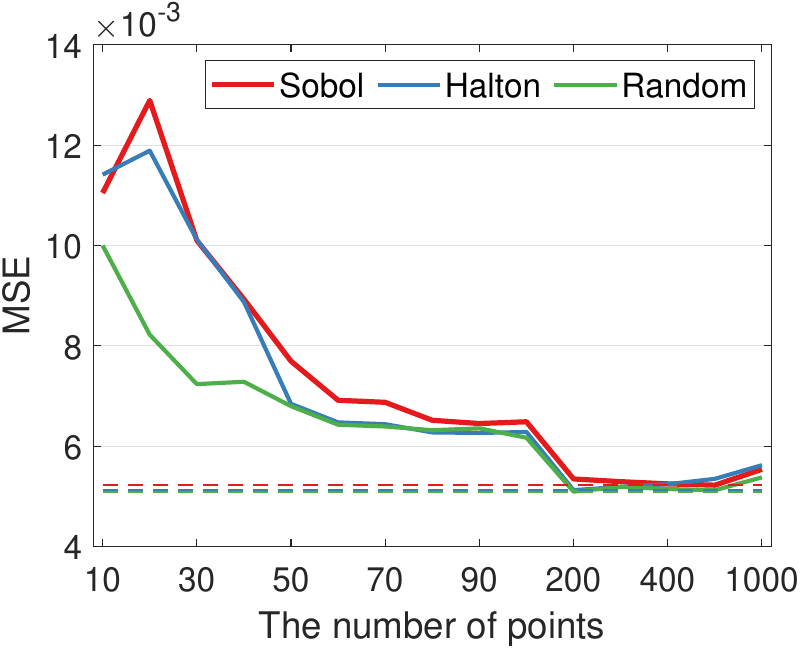}
	}\hspace{-0.145in}
		\subfigure[$m=80$]{
		\includegraphics[width=3.65cm, height=3.15cm]{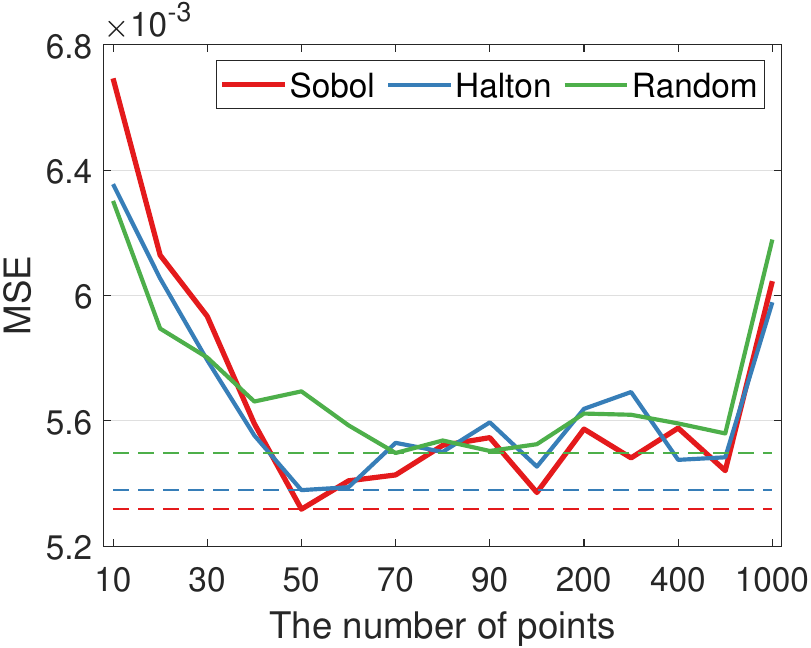}
	}\hspace{-0.145in}
		\subfigure[$m=160$]{
		\includegraphics[width=3.65cm, height=3.15cm]{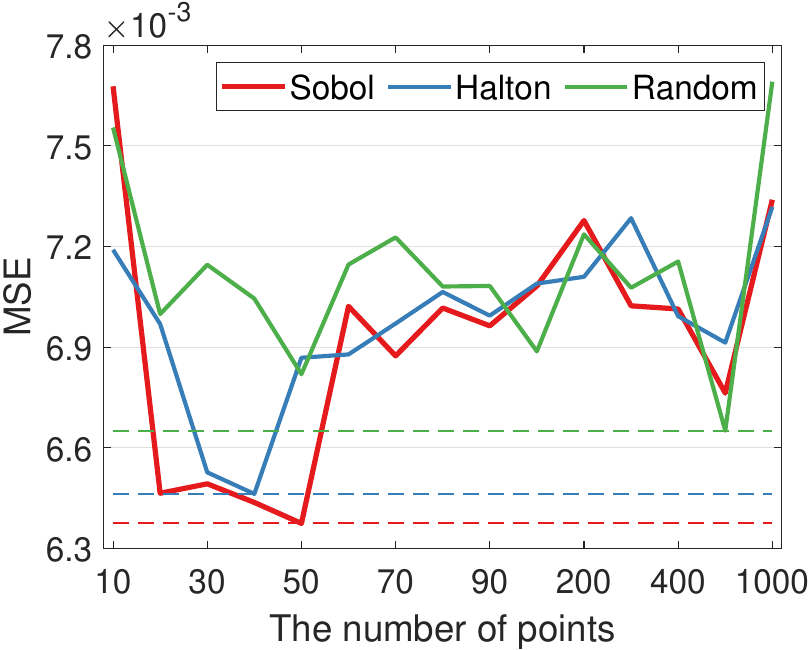}
	}
	\caption{Relationship between test MSE and the number of center points in local approximation using the three low-discrepancy sequences for AdaDKRR with different numbers of local machines}
	\label{MSEKappa}
 \vspace{-0.2in}
\end{figure*}

{\bf Simulation 1:} In this simulation, we select three types of center points for local approximation, including two QMCS (Sobol points and Halton points) and one MCS (random points). The number $m$ of local machines varies from the set $\{20,40,80,160\}$. For each fixed $m$, the relation between the test MSE and the number $n$ of center points is shown in Figure \ref{MSEKappa}, in which the dashed lines exhibit the best test MSEs with the optimal numbers of the three types of center points.
% To investigate the influence of local approximation accuracy by Sobol points on the generalization ability, the approximate MSE under the optimal parameters selected by cross-validation on each local machine is first recorded for each fixed number of local machines and Sobol points, and the average approximate MSE of the local machines against the number of Sobol points is
% % The relation of approximate MSE against the number of Sobol points is also
% plotted in Figure \ref{MSEfit}.
From the above results, we have the following observations: 1) As the number of center points increases, the curves of test MSE have a trend of descending first and then ascending. This is because very few center points cannot provide satisfactory accuracy for local approximation, resulting in the approximate function based on the center points having a large deviation from the ground truth, while a large number of center points put the estimator at risk of over-fitting. 2) The optimal number of center points generally decreases as the number of local machines increases. Because a smaller $m$ indicates that there are more training samples on each local machine, more center points are required to cover these samples to obtain a satisfactory local approximation. 3) The three types of center points perform similarly on the 3-dimensional data, but Sobol points and Halton points are obviously better than random points on the 10-dimensional data, especially for larger numbers of local machines. In addition, the optimal number of random points is usually larger than the number of Sobol points and Halton points. The reason is that the discrepancy of QMCS is smaller than that of MCS, indicating that the sample distribution of QMCS is more uniform than that of MCS. Therefore, QMCS can better describe the structural information of the data and is more effective than MCS in local approximation. Since Sobol points perform similarly to Halton points, we take Sobol points as an example to demonstrate the superiority of the proposed method in the following experiments.
% 2) Although the approximate MSE decreases with the increasing number of Sobol points, relatively small approximate MSE does not yield good generalization ability, e.g., the test MSE becomes worse when the number of Sobol points is larger than 20. The reason for this phenomenon is as follows: The approximate MSE is computed on the training data with noise, and a relatively small approximate MSE indicates that the approximate function based on Sobol points

\begin{figure*}[t]
    \centering
    \subfigcapskip=-3pt
	\subfigure[Dim=3]{\includegraphics[width=7.5cm,height=3.4cm]{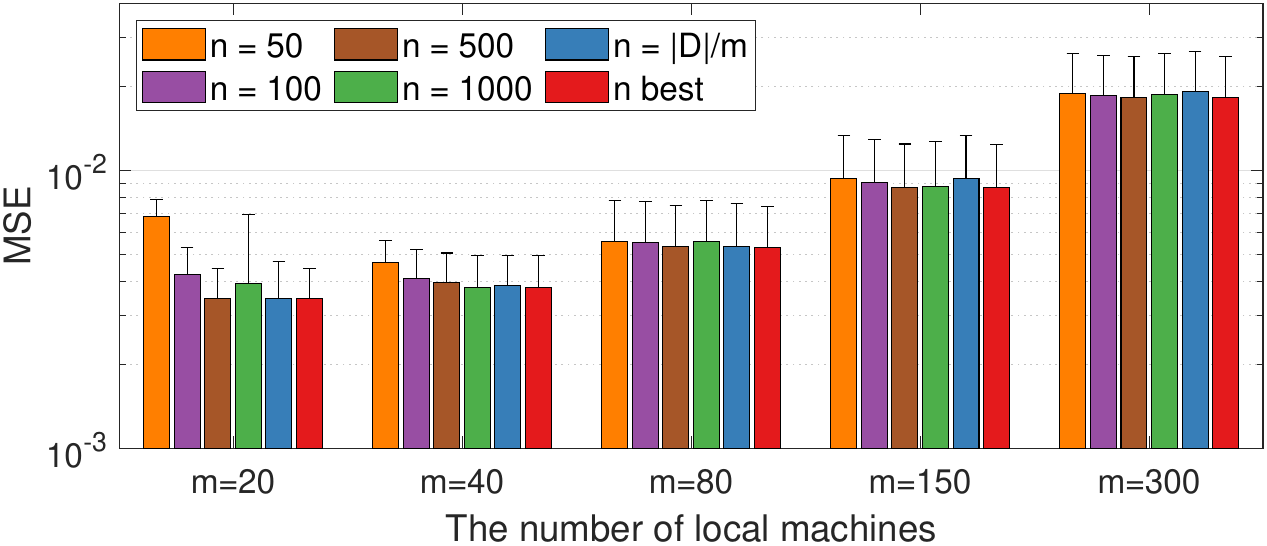}}\hspace{0.3in}\hspace{-0.26in}
    \subfigure[Dim=10]{\includegraphics[width=7.5cm,height=3.4cm]{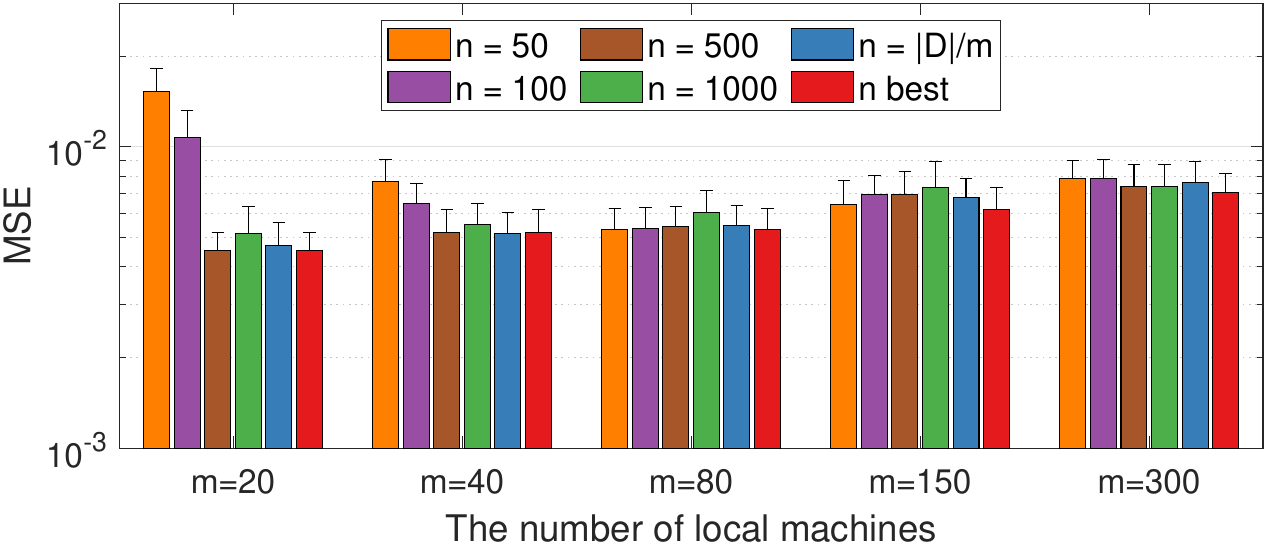}}
	\caption{Relationship between test MSE and the number of Sobol points for AdaDKRR with different numbers of local machines}
\label{MSEtest}
\vspace{-0.2in}
\end{figure*}

{\bf Simulation 2:} In this simulation, we check the robustness of the proposed method concerning the number $n$ of center points, as $n$ determines the accuracy of the local approximation. We set the number $n$ in two ways: 1) by fixing $n$ as a constant (denoted by ``$n=\#$"), and 2) by adaptively adjusting $n$ as the average number of training samples in each local machine (denoted by ``$n=|D|/m$"). We vary the number of Sobol points from the sets $\{10,20,\cdots,100,200,\cdots,500,1000,2000\}$ and $\{10,20,\cdots,100,200,\cdots,500,1000\}$ for the 3-dimensional and 10-dimensional data, respectively, and vary the number of local machines from the set $\{20,40,80,150,300\}$. The testing RMSEs with respect to different orders of magnitude $n$ under different numbers of local machines are shown in Figure \ref{MSEtest}, where ``$n$ best" represents the optimal MSE corresponding to the best $n$ chosen from the candidate set and provides a baseline for assessing the performance of the proposed method. From the results, it can be seen that the generalization performances with different orders of $n$ are all comparable with the best $n$ when $m$ is large (e.g., $m\geq 80$). Even when $m$ is small, we can also obtain a satisfactory result by simply varying a few different orders of magnitude of $n$. In addition, for different numbers of local machines, the proposed method with an adaptive number of center points shows stable performance that is comparable to the best $n$. All these results demonstrate that the proposed method is robust to the number of center points.

\begin{figure*}[t]
    \centering
    \subfigcapskip=-3pt
	\subfigure[Dim=3]{\includegraphics[width=6cm,height=5cm]{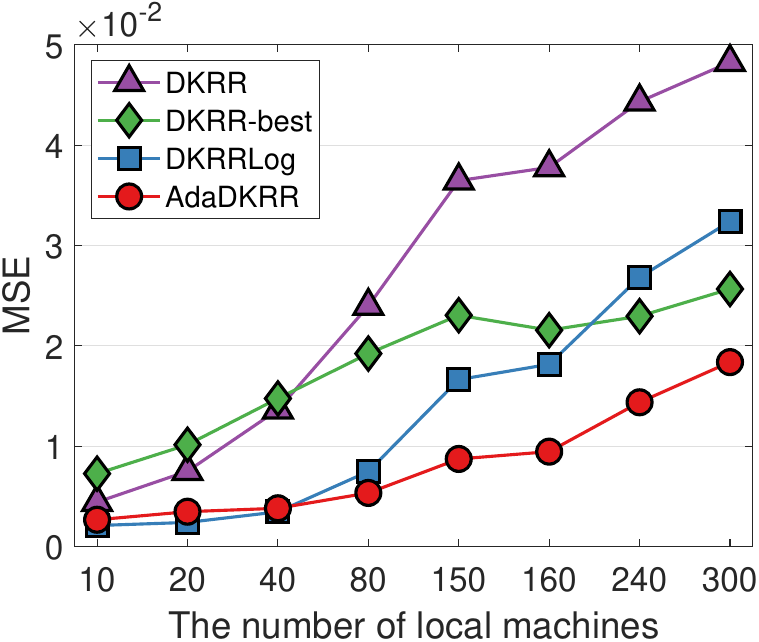}}\hspace{0.3in}
    \subfigure[Dim=10]{\includegraphics[width=6cm,height=5cm]{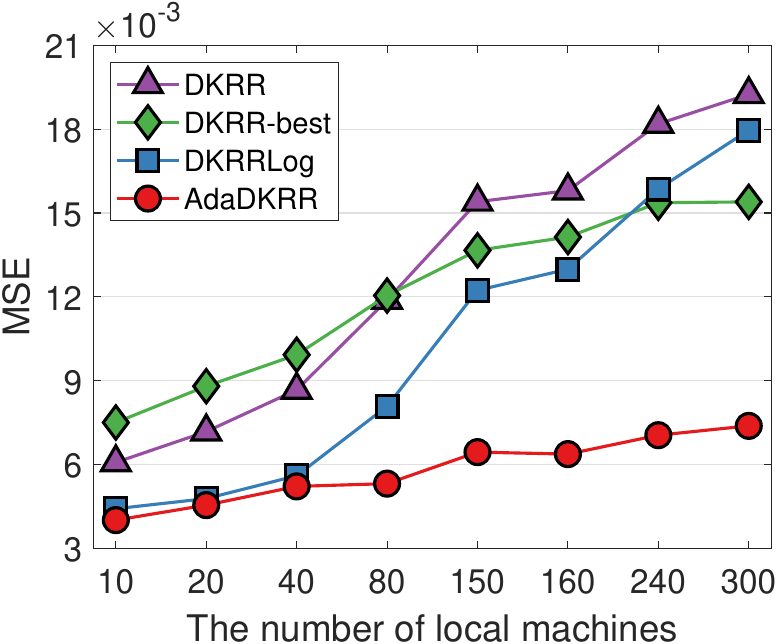}}
	\caption{Comparisons of test MSE among the three parameter selection approaches with different numbers of local machines}
\label{ComparisonMSEtest}
\vspace{-0.2in}
\end{figure*}

{\bf Simulation 3:} This simulation compares the proposed method with DKRR and DKRRLog under the condition that all training samples are uniformly distributed to local machines. The number of local machines changes from the set $\{10,20,40,80,150,160,240,300\}$. For AdaDKRR, the number $n$ of Sobol points is chosen from the set $\{50,100,500,1000\}$. The results of test MSE as a function of the number of local machines are shown in Figure \ref{ComparisonMSEtest}, where ``DKRR-best" denotes the best performance of local machines in DKRR. Based on the above results, we have the following observations: 1) The test MSE grows as the number of local machines increases for all methods, but the growth of AdaDKRR is much slower than that of other methods. 2) When the number of local machines is small (e.g., $m\leq 40$), DKRR-best has the worst performance; the MSE values of DKRR are smaller than those of DKRR-best, which verifies that distributed learning can fuse the information of local machines and achieve better generalization performance than each local machine; AdaDKRR performs similarly to DKRRLog, and both of them are significantly better than DKRR, which provides evidence that it is not a good choice to select parameters only based on the data in each local machine. 3) When the number of local machines increases, the generalization performance of DKRR and DKRRLog deteriorates dramatically, even worse than that of a single local machine (e.g., $m\geq 200$), whereas the test MSE of AdaDKRR grows slowly and has obvious superiority to other methods. These results show that the proposed method is effective and stable in parameter selection for distributed learning.

% As the number of local machines grows, the generalization performance of DKRR and DKRRLog deteriorates dramatically and even gets worse than that of a single local machine (e.g., $m\geq 200$), whereas the test MSE of AdaDKRR increases slowly and has obvious superiority to other methods, especially for large numbers of local machines.
% These results show the effectiveness and stability of the proposed method in parameter selection for distributed learning.

\begin{figure*}[t]
	\centering
        \subfigcapskip=-3pt
	\subfigure{
        % \rotatebox{90}{\small{~~~~~~~~~$d=3$}}
		\includegraphics[width=3.65cm, height=3cm]{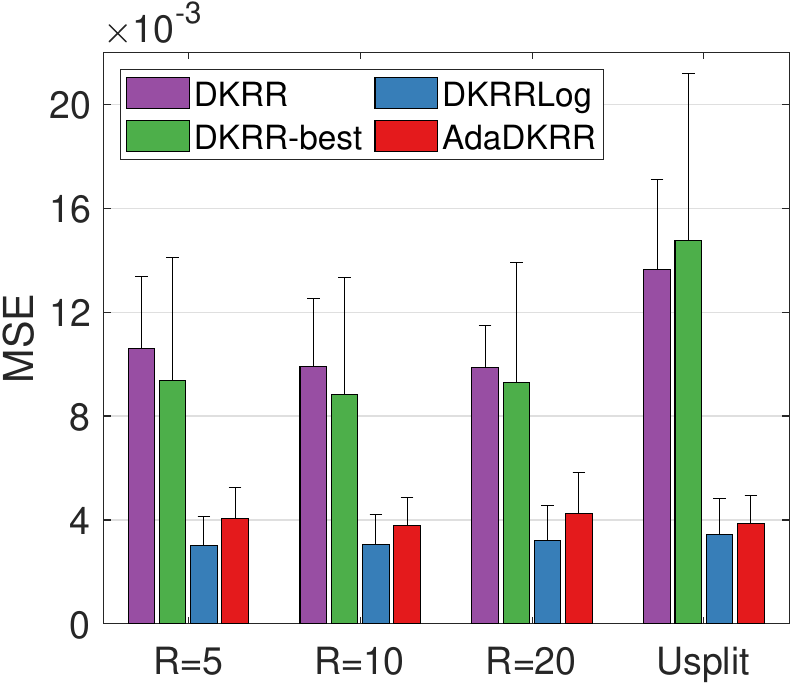}
	}\hspace{-0.12in}
		\subfigure{
		\includegraphics[width=3.65cm, height=3cm]{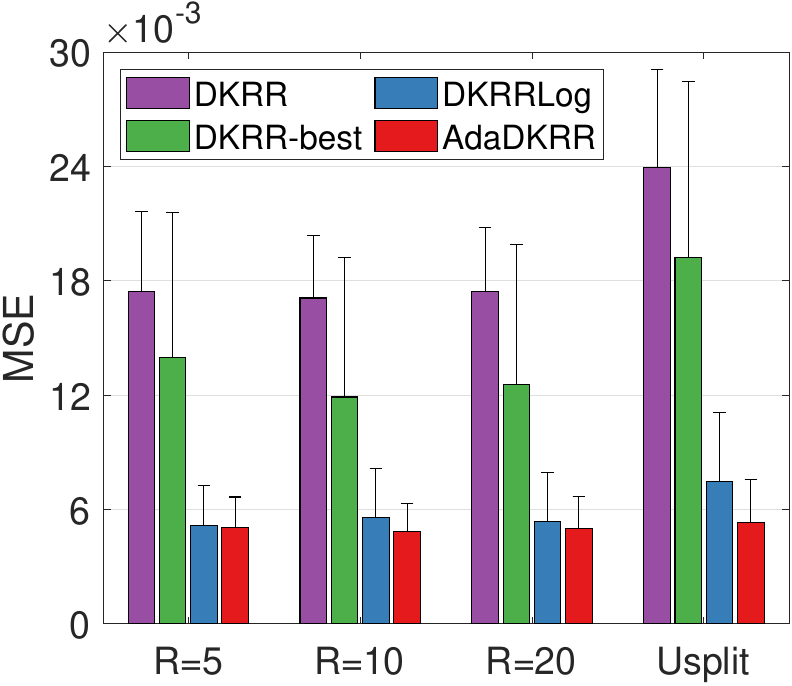}
	}\hspace{-0.12in}
		\subfigure{
		\includegraphics[width=3.65cm, height=3cm]{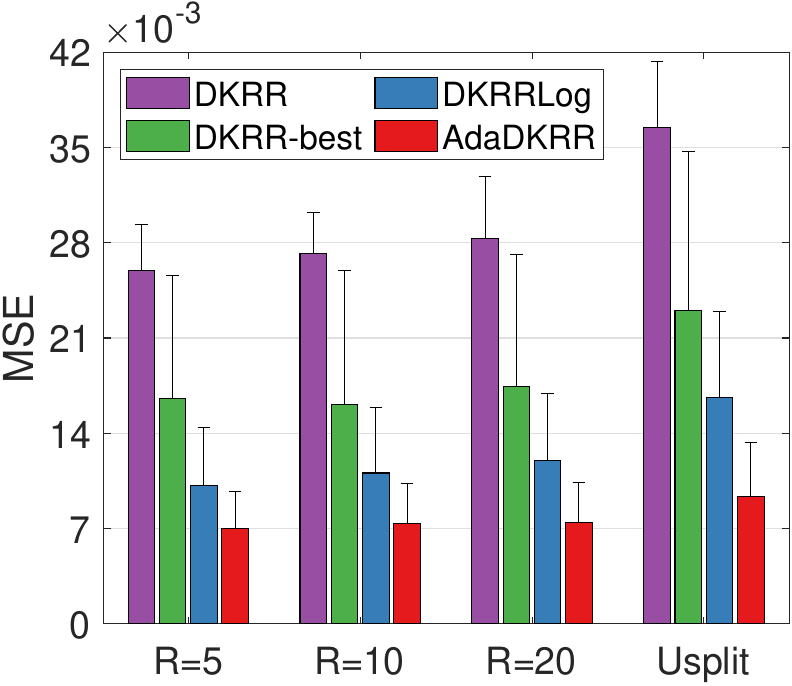}
	}\hspace{-0.12in}
		\subfigure{
		\includegraphics[width=3.65cm, height=3cm]{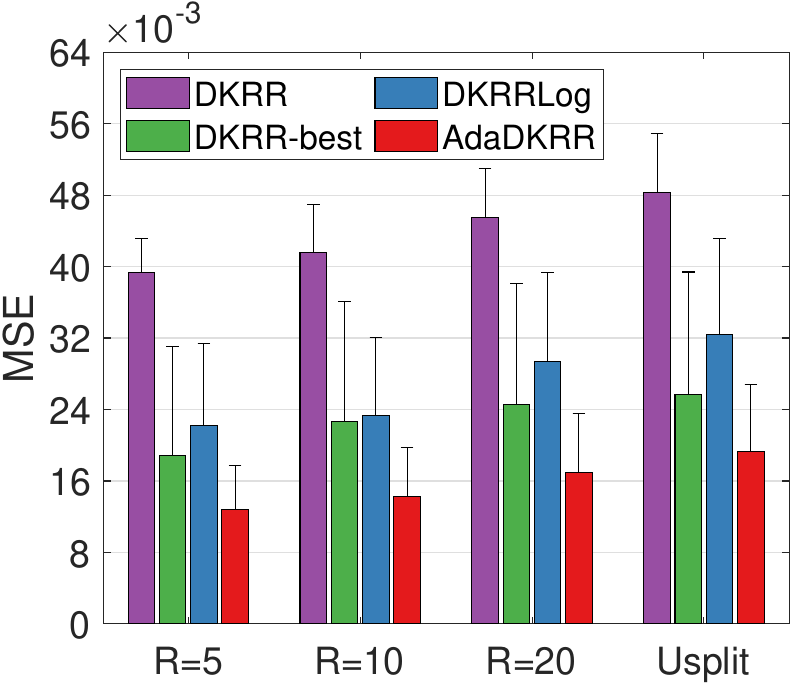}
	}\\
	\vspace{-3mm}
	\setcounter{subfigure}{0}
    \subfigure[$m=40$]{
 		% \rotatebox{90}{\small{~~~~~~~~~$\delta=0.5$}}
		\includegraphics[width=3.65cm, height=3cm]{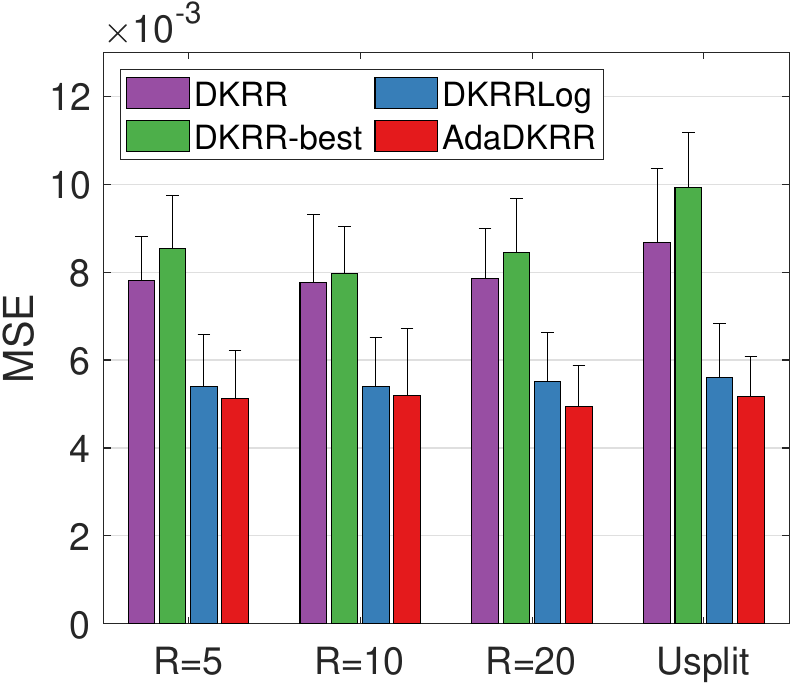}
	}\hspace{-0.12in}
	\subfigure[$m=80$]{
		\includegraphics[width=3.65cm, height=3cm]{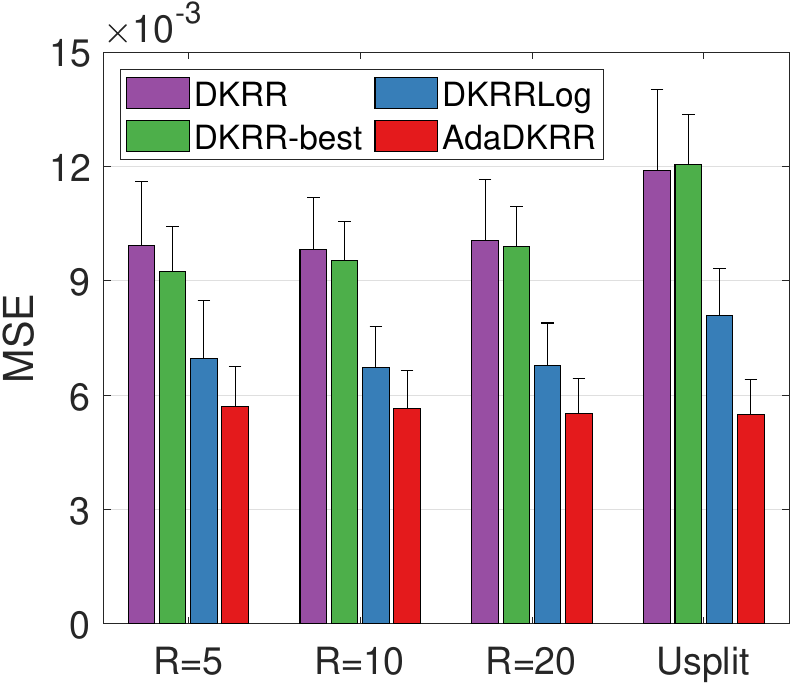}
	}\hspace{-0.12in}
		\subfigure[$m=150$]{
		\includegraphics[width=3.65cm, height=3cm]{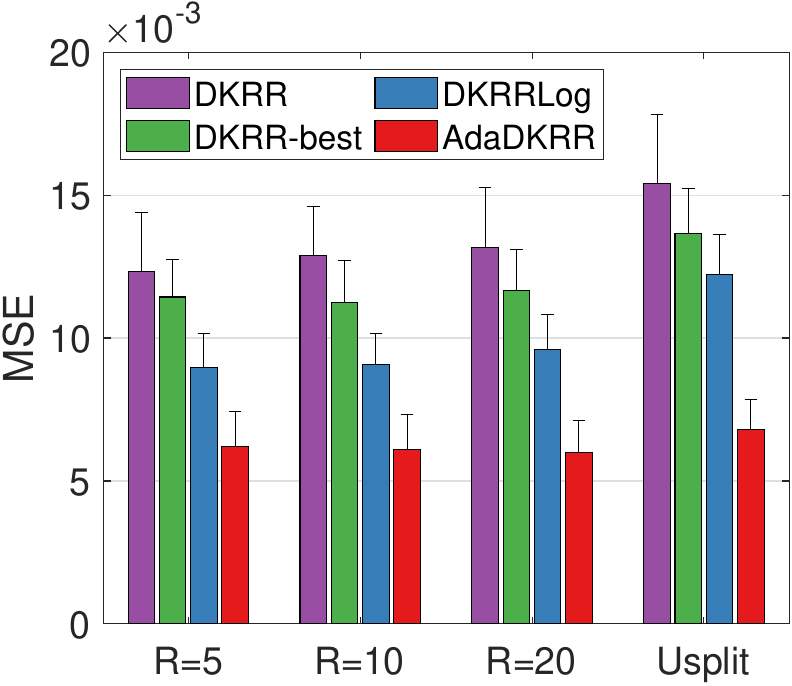}
	}\hspace{-0.12in}
		\subfigure[$m=300$]{
		\includegraphics[width=3.65cm, height=3cm]{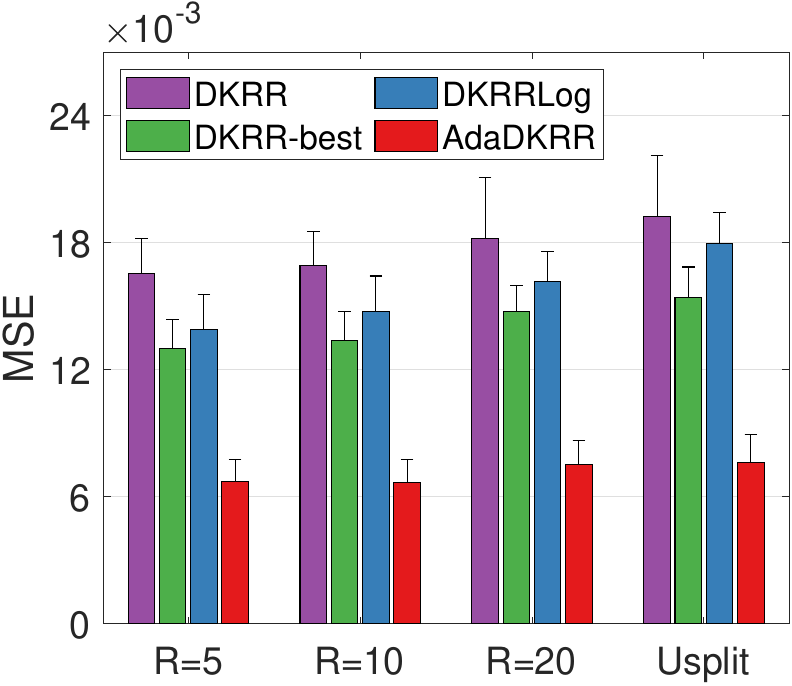}
	}
	\caption{Comparisons of test MSE among the three parameter selection approaches with different distribution patterns of data in local machines}\label{ComparisonMSEtestbar}
 \vspace{-0.2in}
\end{figure*}

{\bf Simulation 4:} In this simulation, we compare the generalization performance of the three methods under non-uniform distributions of the number of training samples in local machines. Specifically, all training samples are randomly distributed to local machines, meaning the numbers of training samples in local machines are also random. In addition, we set the minimum number of training samples on each local machine so that cross-validation could be performed. For example, ``$R=5$" means that the minimum number of training samples on each local machine is no less than 5. AdaDKRR uses the adaptive number of Sobol points (i.e., $n=|D|/m$) in local approximation for convenience. The number $m$ of local machines varies from the set $\{40,80,150,300\}$. For each fixed number $m$, we compare the generalization performance in four cases, including $R\in\{5,10,20\}$ and uniform split (denoted as ``Usplit"), and the results are shown in Figure \ref{ComparisonMSEtestbar}. Note that the case ``Usplit" can be considered as ``$R=|D|/m$".
% From the results, we can see that the three methods in each case of non-uniform split perform similarly to uniform split, as described in Simulation 3.
From the results, we can see that the performance differences among the three methods in each case of the non-uniform split are similar to those in the case of the uniform split, as described in Simulation 3.
Additionally, the test MSE usually increases from the case ``$R=5$ to the case ``Usplit". This is because a smaller value of $R$ means that the distribution of the number of samples is more uneven, and the generalization performance of a single local machine with a large number of training samples is much better than that of a combination of several local machines with the same total number of training samples, as distributed learning shows. Compared with DKRR and DKRRLog, AdaDKRR is more robust to the split of training samples distributed to local machines, especially for the 10-dimensional data.
The above results demonstrate that AdaDKRR is suitable for distributed learning with different data sizes of participants.

\subsection{Real-World Applications}

The mentioned parameter selection methods are tested on two real-world data sets: used car price forecasting and graphics processing unit (GPU) performance prediction for single-precision general matrix multiplication (SGEMM). Before discussing the experiments, it is important to clarify some implementation details: 1) For each data set, half of the data samples are randomly chosen as training samples, and the other half are used as testing samples to evaluate the performance of the mentioned methods. Following the typical evaluation procedure, 10 independent sets of training and testing samples are generated by running 10 random divisions on the data set. 2) For each data set, min-max normalization is performed for each attribute except the target attribute. Specifically, the minimum and maximum values of the $i$-th attribute of training samples are calculated and denoted by $F_{min}^{(i)}$ and $F_{max}^{(i)}$, respectively. The $i$-th attribute of samples is rescaled using the formula $\hat{\mathbf{x}}_i=\left(\mathbf{x}_i-F_{min}^{(i)}\right)/\left(F_{max}^{(i)}-F_{min}^{(i)}\right)$, where $\mathbf{x}_i$ is the $i$-th attribute vector. 3) Based on the numerical results provided in Simulation 3 of the previous subsection, we vary the number of Sobol points in the set $\{10,50,100,500\}$ and record the best result for AdaDKRR; the Gaussian kernel is used for the two data sets. 4) We consider the case that all training samples are uniformly distributed to local machines.

\subsubsection{Used Car Price Forecasting}

As the number of private cars increases and the used car industry develops, more and more buyers are making used cars their primary choice due to their cost-effectiveness and practicality. Car buyers usually purchase used cars from private sellers and auctions aside from dealerships, and there is no manufacturer-suggested retail price for used cars. Car sellers want to reasonably evaluate the residual value of used cars to ensure sufficient profit margins. Car buyers hope the car they buy is economical, or at least they won't buy overpriced cars due to their unfamiliarity with the pricing of used cars. Therefore, pricing a used car can be regarded as a very important decision problem that is related to the success of the transaction between buyers and sellers. However, the sale price of used cars is very complicated, because it depends not only on the wear and tear of the car, such as usage time, mileage, and maintenance, but also on the performance of the car, such as brand, gearbox, and power, as well as on some social factors, such as car type, fuel type, and sale region. Sellers and buyers usually spend a lot of time and effort negotiating the price of used cars, so it is desirable to develop an effective pricing model from a collection of existing transaction data to provide a reliable reference for sellers and buyers and promote the success of transactions.

% whether a transaction for both buyers and sellers is successful.
\begin{figure*}[t]
    \centering	 \subfigure{\includegraphics[width=13.5cm,height=4.2cm]{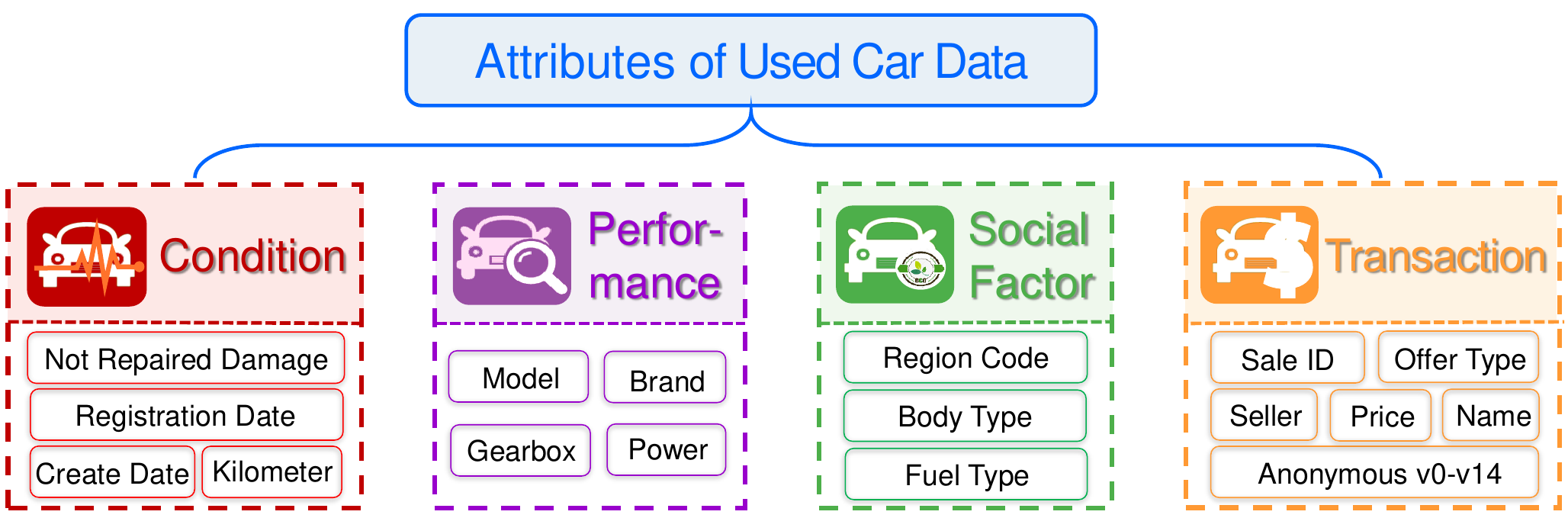}}
	\caption{Details of the attributes of the data set \texttt{CarTianchi} } \label{Fig:UsedCarFea}
 \vspace{-0.2in}
\end{figure*}

\begin{table}[t]
\centering
\caption{The details of data binning on the data set \texttt{CarTianchi}}\label{BinCar}
\renewcommand\arraystretch{1}
 % \begin{tabular}{ c c}
 \begin{tabular}{
                  *{1}{p{1.8cm}<{\centering}}
			   *{1}{|p{12.5cm}<{\centering}}
			           }
 \toprule[1.5pt]
 Attribute & Data binning\\
\hline
\hline
 \multirow{2}{*}{\shortstack{Usage time\\(day)}} & 0: [0,90], 1: (90,180], 2: (180, 365], 3: (365,730], 4: (730,1095], 5: (1095,\\
 & 1460],  6: (1460,2190], 7: (2190,3650], 8: (3650,5475], 9: (5475,$+\infty$) \\
 \hline
\multirow{3}{*}{Power} & 0: [-19.3,1931.2], 1: (1931.2,3862.4], 2: (3862.4,5793.6], 3: (5793.6,7724.8], \\
 & 4: (7724.8,9656.0], 5: (9656.0,11587.2], 6: (11587.2,13518.4], \\
 &  7: (13518.4,15449.6], 8: (15449.6,17380.8], 9: (17380.8,19312]\\
 \toprule[1.5pt]
 \end{tabular}
 \vspace{-0.2in}
\end{table}

\begin{figure*}[t]
    \centering
    \subfigcapskip=-3pt
	\subfigure[Raw price]{\includegraphics[width=6.5cm,height=4.6cm]{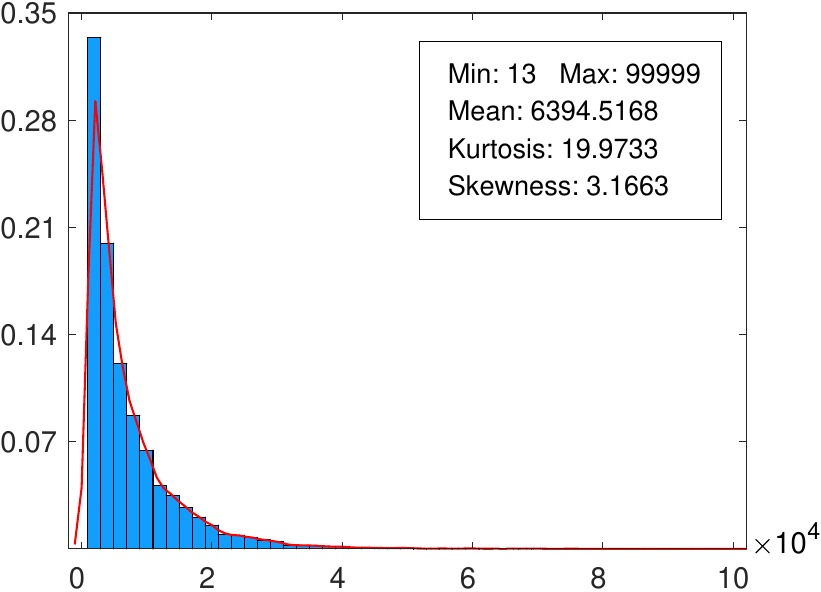}}\hspace{0.2in}
    \subfigure[Transformed price]{\includegraphics[width=6.3cm,height=4.6cm]{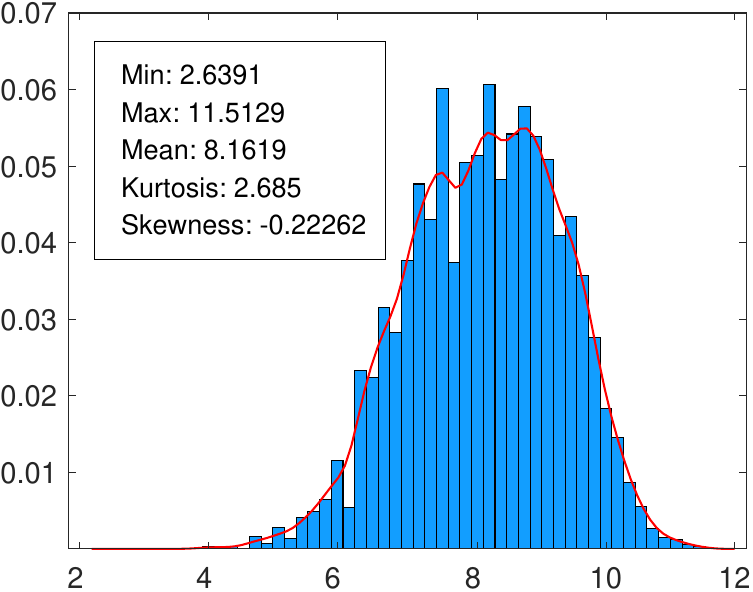}}
	\caption{The histogram of historical transaction prices}
\label{CarPriceHist}
\vspace{-0.2in}
\end{figure*}

\begin{figure*}[t]
    \centering	
    \subfigure{\includegraphics[width=6cm,height=5cm]
    {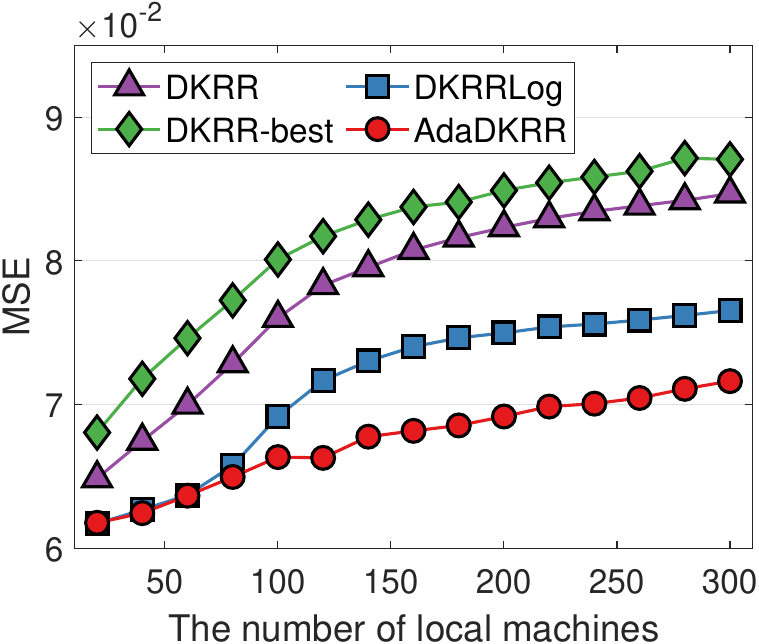}}
	\caption{Comparisons of test MSE among the three parameter selection approaches with different numbers of local machines on the data set \texttt{CarTianchi}} \label{ComparisonMSEtestehCar}
 \vspace{-0.2in}
\end{figure*}

The used car data on Tianchi (\texttt{CarTianchi} for short) \protect\footnotemark[2]
\footnotetext[2]{https://tianchi.aliyun.com/competition/entrance/231784/information}
provided by Alibaba Cloud comes from the used car transaction records of a trading platform, and its goal is to establish models to predict the price of used cars. The data set contains more than $400,000$ samples; each sample is described by 31 attributes, of which $15$ are anonymous and $4$ are masked to protect the confidentiality of the data. The attributes are described in detail in Figure \ref{Fig:UsedCarFea}, and they are grouped into four categories, including car condition, performance, social factor, and transaction. Note that the $15$ anonymous attributes are classified into the category of transaction for convenience. In the experiment, we selected a subset of $129,710$ samples with price attributes after removing samples with missing values to train and evaluate models. We use the time difference between the sales date (i.e., the create date) and the registration date as an approximation of the usage time and remove the attributes of the sale ID, create date, and registration date. Data binning is applied to the attributes of usage time and power because their values are highly dispersed, and the details are listed in Table \ref{BinCar}. The histogram of the target attribute of prices, as well as the skewness and kurtosis, are shown in Figure \ref{CarPriceHist} (a), from which it can be seen that the data distribution does not obey the normal distribution, with a sharp peak and a long tail dragging on the right. Therefore, the target attribute of price is transformed by a logarithmic operation, and the histogram of the transformed price is close to a normal distribution, as shown in Figure \ref{CarPriceHist} (b). The regularization parameter $\lambda$ is chosen from the set $\{\frac{1}{3^q} | \frac{1}{3^q} \geq 10^{-10}, q=0,1,2,\cdots\}$, and the kernel width $\sigma$ is chosen from 10 values that are drawn in a logarithmic, equally spaced interval $[1, 10]$.

% \begin{table}[t]
% \centering
% \caption{The details of data binning on the data set \texttt{CarTianchi}.}\label{BinCar}
% \renewcommand\arraystretch{0.7}
%  \begin{tabular}{ c c}
%  \toprule[1.5pt]
%  Attribute & Data binning\\
% \hline
% \hline
%  \multirow{3}{*}{\shortstack{Usage time\\(day)}} & 0: [0,90], 1: (90,180], 2: (180, 365], 3: (365,730], \\
%  & 4: (730,1095], 5: (1095,1460], 6: (1460,2190], \\
%  & 7: (2190,3650], 8: (3650,5475], 9: (5475,$+\infty$) \\
%  \hline
% \multirow{4}{*}{Power} & 0: [-19.3,1931.2], 1: (1931.2,3862.4], 2: (3862.4,5793.6], \\
%  & 3: (5793.6,7724.8], 4: (7724.8,9656.0], 5: (9656.0,11587.2] \\
%  & 6: (11587.2,13518.4], 7: (13518.4,15449.6], \\
%  & 8: (15449.6,17380.8], 9: (17380.8,19312]\\
%  \toprule[1.5pt]
%  \end{tabular}
% \end{table}

% % Comparisons of Test Error Among the 10Methods Under Different Parameter Values
% % Comparisons of Test Errors Among the 10 Methods on Low-Cross Data and High-Cross Data with Different Numbers of Noise Dimensions
% Comparisons of test MSE among the three parameter selection approaches with different numbers of local machines

The relationship between test MSE and the number $m$ of local machines for the compared methods is shown in Figure \ref{ComparisonMSEtestehCar}, where $m$ varies from the set $\{20,40,\cdots,300\}$. From the results, we can see that DKRR-best has the worst generalization performance due to the limited number of training samples in local machines; DKRR synthesizes the estimators of local machines and thus achieves better performance than each local estimator; although the logarithmic transformation on the parameters makes DKRRLog superior to DKRR, it can still be significantly improved by AdaDKRR, especially for large numbers of local machines (e.g., $m\geq 100$). The above results demonstrate the effectiveness of the proposed parameter selection approach in distributed learning.

\subsubsection{SGEMM GPU Performance Prediction}

Over the past decade, GPUs have delivered considerable performance in multi-disciplinary areas such as bioinformatics, astronomy, and machine learning. However, it is still a challenging task to achieve close-to-peak performance on GPUs, as professional programmers must carefully tune their code for various device-specific problems, each of which has its own optimal parameters such as workgroup size, vector data type, tile size, and loop unrolling factor. Therefore, it is important to design effective GPU acceleration models to automatically perform parameter tuning with the data collected from the device.

\begin{figure*}[t]
    \centering
	\subfigure{\includegraphics[width=6cm,height=5cm]{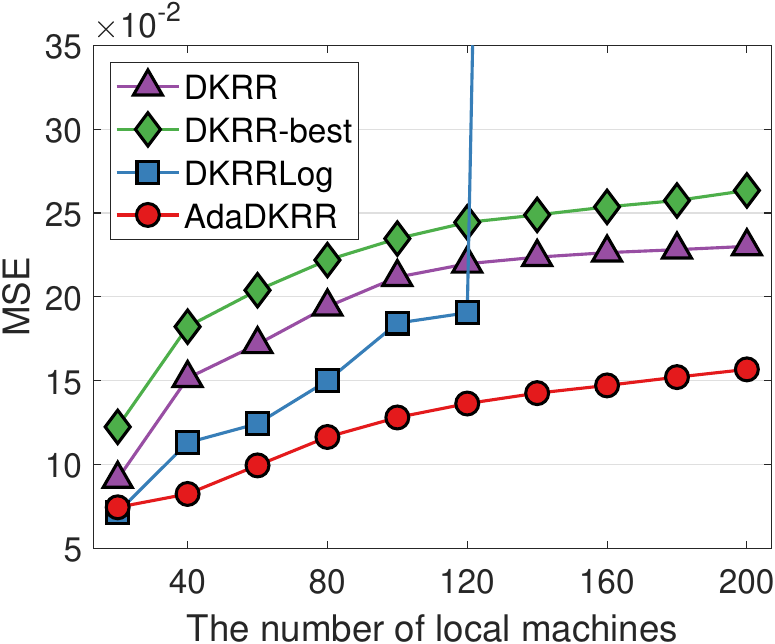}}\hspace{0.3in}
    \subfigure{\includegraphics[width=6cm,height=4.68cm]{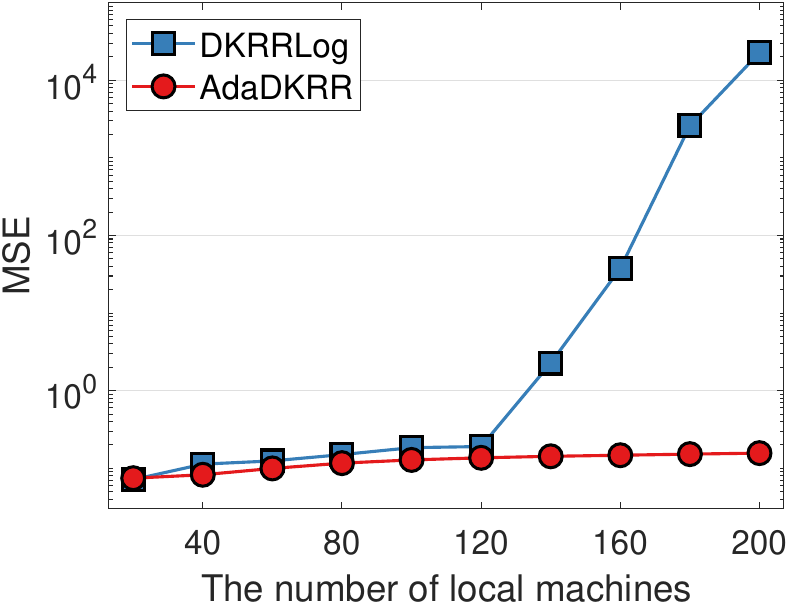}}
	\caption{Comparisons of test MSE among the three parameter selection approaches with different numbers of local machines on the data set \texttt{SGEMM GPU}}
\label{ComparisonMSEtestsgemmgpu}
\vspace{-0.2in}
\end{figure*}

% The data set \texttt{SGEMM GPU} considers the running time of a dense matrix-matrix multiplication, i.e., $C=\alpha A^\top B + \beta C$, where $\alpha$ and $\beta$ are constants and $A^\top$ is the transpose of $A$, because matrix multiplication is a fundamental building block in deep learning and other machine learning methods.
% because matrix-multiplication is one of the most important blocks in deep learning and other machine learning algorithms.

The data set \texttt{SGEMM GPU} \citep{Nugteren2015} considers the running time of dense matrix-matrix multiplication $C=\alpha A^T B + \beta C$, as matrix multiplication is a fundamental building block in deep learning and other machine learning methods, where $A\in \mathbb{R}^{K\times M}$, $B\in \mathbb{R}^{K\times N}$, $C\in \mathbb{R}^{M\times N}$, $M=N=K=2048$, $\alpha$ and $\beta$ are constants, and $A^T$ is the transpose of $A$. The data set contains $241,600$ samples; each sample includes a possible combination of $14$ parameters of the SGEMM kernel and $4$ running times for this parameter combination. The $14$ parameters and their corresponding domains are as follows:
\begin{itemize}
\setlength{\parskip}{0pt}
\setlength{\itemsep}{0pt plus 1pt}
\item Per-matrix 2D tiling at workgroup-level uses the parameters $M_{wg}, N_{wg}\in\{16,32,64,$ $128\}$, which correspond to the matrix dimensions of $M$ and $N$, respectively.
\item The inner dimension of 2D tiling at workgroup-level uses the parameter $K_{wg} \in \{16,32\}$, which corresponds to the dimension of $K$.
\item Local workgroup size uses the parameters $M_{dimC}, N_{dimC}\in\{8,16,32\}$.
\item Local memory shape uses the parameters $M_{dimA}, N_{dimB} \in \{8,16,32\}$.
\item The kernel loop unrolling factor is denoted by $K_{wi}\in\{2,8\}$.
\item Per-matrix vector widths for loading and storage use parameters $M_{vec}, N_{vec}\in\{1,2,4,$ $8\}$, where $M_{vec}$ is for matrices $A$ and $C$, and $N_{vec}$ is for matrix $B$.
\item The enabling stride for accessing off-chip memory within a single thread is denoted by $M_{stride}, N_{stride}\in\{0,1\}$, where $M_{stride}$ is for matrices $A$ and $C$, and $N_{stride}$ is for matrix $B$.
\item Per-matrix manual caching of the 2D workgroup tile can be controlled by parameters $L \$_A, L \$_B\in\{0,1\}$.
\end{itemize}
\vspace{-\topsep}

In the experiment, the first 14 columns of the data are used as data input, and the average time of the 4 runs is regarded as data output. Similar to the data set \texttt{CarTianchi}, the distribution of the average running time has a sharp peak and a long tail dragging on the right. Therefore, as suggested by \citet{Nugteren2015}, we also perform a logarithmic operation on the average running time. The regularization parameter $\lambda$ is chosen from the set $\{\frac{1}{5^q} | \frac{1}{5^q} \geq 10^{-10}, q=0,1,2,\cdots\}$, and the kernel width $\sigma$ is chosen from 10 values that are drawn in a logarithmic, equally spaced interval $[1, 100]$.

Figure \ref{ComparisonMSEtestsgemmgpu} records the relationship between test MSE and the number of local machines for the compared methods. It can be seen that the performance of these methods on the data set \texttt{SGEMM GPU} is similar to that on the data set \texttt{CarTianchi}. The only difference is that DKRRLog performs extremely poorly in generalization when the number of local machines is larger than $120$. This is because the transformed parameters are far from the optimal parameters in some local machines. These results provide another piece of evidence that the proposed AdaDKRR is stable and effective in selecting parameters.

% Furthermore, AdaDKRR is applied in assessing consumer credit and
% text categorization with four data sets and exhibits
% excellent learning performance.

% we also applied AdaDKRR to four real-world
% data sets, including ones designed to help determine
% credit approval and article subjectivity to test the
% usability of AdaDKRR in practice. The numerical results
% verify our theoretical assertions and show the power
% of AdaDKRR in applications.

\section{Conclusion}

This paper proposed an adaptive parameter selection strategy for distributed learning to settle the data silos. Specifically, by communicating the coefficients of fixed basis functions, we obtained a good approximation of the global estimator and thus determined the algorithm parameters for the global approximation without leaking any sensitive information about the data. From a theoretical perspective, we established optimal rates of excess generalization error for the proposed method in a framework of statistical learning theory by utilizing the idea of low-discrepancy sequences and the classical radial basis function approximation. According to the theoretical findings, as long as the number of local machines is not too large, the proposed method is similar to running KRR on the whole data. The theoretical results demonstrate the efficacy of the proposed method for parameter selection in distributed learning. From an application point of view, we also applied the proposed method to several simulations and two real-world data sets, used car price forecasting and GPU performance prediction. The numerical results verify our theoretical assertions and demonstrate the feasibility and effectiveness of the proposed method in applications.

% \ACKNOWLEDGMENT{
% The work of Shao-Bo Lin is supported partially by the National Natural Science Foundation of China (Grant No. 61876133).
% The work of Di Wang is supported partially by the National Natural Science Foundation of China (Grant No. 61772374).
% The work of Ding-Xuan Zhou was partially supported by InnoHK initiative, the Government of the HKSAR, and Laboratory for AI-Powered Financial Technologies.
% }

% \bibliographystyle{informs2014} % outcomment this and next line in Case 1
\bibliography{distributed} % if more than one, comma separated

 \newpage

\section*{Appendix A: Training and Testing Flows of AdaDKRR}
In this part, we give a detailed implementation of AdaDKRR (with cross-validation) as shown in Algorithm \ref{Algorithm:AdaDKRRCV}, where steps 1--19 are the training process and steps 20--24 are the testing process for a query set. In Algorithm \ref{Algorithm:AdaDKRRCV},
% $B_{n,K}:=\left\{\sum_{k=1}^na_kK_{\xi_k}:a_k\in\mathbb R\right\}$ is a set of basis functions based on the low-discrepancy sequence $\Xi_n=\{\xi_1,\cdots,\xi_n\}$;
$\mathbb{K}_{\left|D_j\right|,n}\in \mathbb{R}^{\left|D_j\right|\times n}$ is a kernel matrix with the element in the $i$-th row and the $k$-th column being $K(x_{ij},\xi_k)$, $\mathbb{K}_{n,n}\in \mathbb{R}^{n\times n}$ is a kernel matrix with the element in the $k$-th row and the $k'$-th column being $K(\xi_k, \xi_k')$, and $\mathbb{I}_{D}$ is the identity matrix with size $|D|$. $\vec{y}_{D}$ is a vector composed of the outputs of data $D$.
% $\mathbb{A}^\dag$ and $\mathbb{A}^\top$ denote the pseudo-inverse and transpose of a matrix $\mathbb{A}$, respectively. $\pi_M$ is the truncation operator defined by $\pi_M(x)=\mbox{sign}(x)\min\{|x|,M\}$.
In Step 8, $\vec{a}_{j,l,\ell}^{local} := \left(a_{j,l,1,\ell}^{local}, \cdots, a_{j,l,n,\ell}^{local}\right)^T \hspace{-0.03in}= \hspace{-0.03in}\left(\mathbb{K}_{\left|D_{j,l}^{tr}\right|,n}^T \mathbb{K}_{\left|D_{j,l}^{tr}\right|,n} + \mu \left|D_{j,l}^{tr}\right| \mathbb{K}_{n,n} \right)^\dag \mathbb{K}_{\left|D_{j,l}^{tr}\right|,n}^T  {\vec{f}}_{j,l,\ell}^{tr}$, where $\vec{f}_{j,l,\ell}^{tr}$ is a vector composed of the outputs of data $F_{j,l,\ell}^{tr}$. In Step 11, $\left|D_{l}^{tr}\right|=\sum\limits_{j=1}^m \left|D_{j,l}^{tr}\right|$. In Step 13, $\vec{a}_{l,\ell}^{global} := \left(a_{l,1,\ell}^{global}, \cdots, a_{l,n,\ell}^{global}\right)^T$. In Step 18, $\vec{\alpha}_{D_j}^\ast:=\left(\alpha_{1j}^\ast, \cdots, \alpha_{|D_j|j}^\ast\right)^T=\left(\mathbb{K}_{D_{j},D_{j}}+\lambda_j^\ast |D_{j}|\mathbb{I}_{D_{j}}\right)^{-1}\vec{y}_{D_{j}}$.
Note that we retrain the local estimator $f_{D_j,\lambda_j^\ast}$ with all training samples $D_j$ under the selected optimal regularization parameter $\lambda_j^\ast$ in the $j$-th local machine for $j=1,\cdots,m$ to improve the generalization performance of AdaDKRR.

% -------------------- version 3 -------------------------
\vspace{0.1in}
\begin{breakablealgorithm}
	\caption{\quad Training and testing flows of AdaDKRR (with cross-validation)}
	\label{Algorithm:AdaDKRRCV}
  \begin{algorithmic}[1]
 \Statex  \hspace{-0.3in} \textbf{Input:} Training data subset $D_j=\{(x_{ij},y_{ij})\}_{i=1}^{|D_j|}$ stored in the $j$-th local machine for $j=1,\cdots,m$,
 query data $D'$, and a candidate set of the regularization parameter $\Lambda=\{\lambda_\ell\}_{\ell=1}^L$.
 \Statex \hspace{-0.25in}  \texttt{$\#$ Training process}
 \For{$j=1,\cdots,m$}
    \State Generate $\iota$ groups of training and validation sets via $\iota$-fold cross validation on data
    \Statex \hspace{0.2in} $D_j$, and denote them as $\left\{\left(D_{j,l}^{tr},D_{j,l}^{val}\right)\right\}_{l=1}^\iota$. \Comment{Data Split}
    \State {Generate the same centers $\Xi_n=\{\xi_k\}_{k=1}^n$ satisfying \eqref{ass-points} and define a set of functions
    \Statex \hspace{0.2in}  $B_{n,K}:=\left\{\sum_{k=1}^na_kK_{\xi_k}:a_k\in\mathbb R\right\}$, where $K_{\xi}(x)=K(\xi,x)$. \Comment{Basis Generation}
}
\EndFor
\For{$l=1,\cdots,\iota$}\Comment{Operate on the $l$-th Split}
    \For{$j=1,\cdots,m$}
        \State Given $\lambda_\ell\in\Lambda$, run KRR with data $D_{j,l}^{tr}$ to obtain \Comment{Local Processing}
            \begin{equation*}
            \setlength{\abovedisplayskip}{0pt}
             \setlength{\belowdisplayskip}{3pt}
                f_{D_{j,l}^{tr},\lambda_\ell} = \arg\min\limits_{f\in\mathcal{H}_K}\left\{ \frac{1}{\left|D_{j,l}^{tr}\right|}\sum\limits_{(x,y)\in D_{j,l}^{tr}} (f(x)-y)^2 + \lambda_\ell\|f\|_K^2\right\}.
            \end{equation*}
        \State Generate a set of points $F_{j,l,\ell}^{tr}=\left\{ \left(x_{i,j},f_{D_{j,l}^{tr},\lambda_\ell}(x_{i,j}) \right)\right\}_{i=1}^{\left|D_{j,l}^{tr}\right|}$, and run KRR with
        \Statex \hspace{0.42in} $F_{j,l,\ell}^{tr}$ to obtain an approximation of $f_{D_{j,l}^{tr},\lambda_\ell}$ by  \Comment{Local Approximation}
            \begin{equation*}
            \setlength{\abovedisplayskip}{0pt}
             \setlength{\belowdisplayskip}{3pt}
          \hspace{0.4in}  f_{D_{j,l}^{tr},\lambda_\ell,n,\mu}^{local} = \arg\min\limits_{f\in B_{n,K}} \frac{1}{\left|D_{j,l}^{tr}\right|}\sum\limits_{(x,y)\in F_{j,l,\ell}^{tr}} (f(x)-y)^2 + \mu\|f\|_K^2 = \sum\limits_{k=1}^{n}a_{j,l,k,\ell}^{local} K_{\xi_{k}}.
            \end{equation*}
        \State Transmit $\left(a_{j,l,k,\ell}^{local}\right)_{k=1,\ell=1}^{n,L}$  to the global machine. \Comment{Communication(I)}
    \EndFor
    \State Synthesize the coefficients by $
        a_{l,k,\ell}^{global} = \sum\limits_{j=1}^m \frac{\left|D_{j,l}^{tr}\right|}{\left|D_{l}^{tr}\right|}a_{j,l,k,\ell}^{local}$, and communicate the matrix
    \Statex\hspace{0.2in} $\left(a_{l,k,\ell}^{global}\right)_{k=1,\ell=1}^{n,L}$ to each local machine. \Comment{Synthesization and Communication(II)}
    \For{$j=1,\cdots,m$}
        \State Compute $L$ MSEs on the validation set $D_{j,l}^{val}$ by \Comment{Local Validation}
        $$
        \setlength{\abovedisplayskip}{0pt}
             \setlength{\belowdisplayskip}{3pt}
        e_{j,l,\ell}=\frac{1}{|D_{j,l}^{val}|} \left\|\vec{y}_{D_{j,l}^{val}} - \pi_M \left(\mathbb{K}_{\left|D_{j,l}^{val}\right|,n}\vec{a}_{l,\ell}^{global}\right) \right\|_2^2 \quad \mbox{for} \quad \ell=1,\cdots,L,
        $$
        \Statex \hspace{0.42in} and denote the MSE vector with size $L$ as $\vec{e}_{j,l}:=\left(e_{j,l,1}, \cdots, e_{j,l,L}\right)^T$.
    \EndFor
\EndFor
\For{$j=1,\cdots,m$}
\State Average $\vec{e}_{j,l}$ over the $\iota$-fold, i.e., $\vec{\bar{e}}_j = \frac{1}{\iota} \sum\limits_{l=1}^\iota \vec{e}_{j,l}$, and select the optimal parameter
\Statex \hspace{0.2in}  $\lambda_j^\ast \in \Lambda$ with the minimum element in the vector $\vec{\bar{e}}_j$. \Comment{Parameter Selection}
\State Run KRR with data $D_{j}$ under the optimal parameter $\lambda_j^\ast$ to obtain\Comment{Retraining}
\begin{equation*}
\setlength{\abovedisplayskip}{0pt}
\setlength{\belowdisplayskip}{3pt}
f_{D_j,\lambda_j^\ast} = \arg\min\limits_{f\in\mathcal{H}_K}\left\{ \frac{1}{|D_{j}|}\sum\limits_{(x,y)\in D_{j}} (f(x)-y)^2 + \lambda_j^\ast\|f\|_K^2\right\}=\sum\limits_{i=1}^{|D_j|} \alpha_{ij}^\ast K_{x_{ij}}.
\end{equation*}
% \Statex \hspace{0.2in} where $\vec{\alpha}_{D_j}^\ast:=\left[\alpha_{1j}^\ast, \cdots, \alpha_{|D_j|j}^\ast\right]^\top=\left(\mathbb{K}_{D_{j},D_{j}}+\lambda_j^\ast |D_{j}|\mathbb{I}_{D_{j}}\right)^{-1}\vec{y}_{D_{j}}$.
\EndFor
 \Statex \hspace{-0.25in}  \texttt{$\#$ Testing process}
 \State Distribute the query data $D'$ to $m$ local machines.
 \For{$j=1,\cdots,m$}
    \State Compute a vector of size $|D'|$ by
    $ \pi_M f_{D_j,\lambda_j^\ast}(D') = \pi_M \left(\mathbb{K}_{\left|D'\right|,\left|D_j\right|}\vec{\alpha}_{D_j}^\ast\right)$,
    and commu-
    \Statex \hspace{0.2in} nicate it to the global machine.  \Comment{Communication(III)}
\EndFor
\State Synthesize the final estimate by
% $$
%        f^{\mbox{\scriptsize{Ada}}}_{D,\vec{\lambda}^\ast}(D'):=\sum_{j=1}^m\frac{|D_j|}{|D|}\pi_M f_{D_j,\lambda_j^\ast}(D') =\sum_{j=1}^m\frac{|D_j|}{|D|}\pi_M \left(\mathbb{K}_{D',D_j}\vec{\alpha}_{D_j}^\ast\right).
% $$
$$
\setlength{\abovedisplayskip}{0pt}
\setlength{\belowdisplayskip}{3pt}
\overline{f}^{Ada}_{D,\vec{\lambda}^\ast}(D'):=\sum_{j=1}^m\frac{|D_j|}{|D|}\pi_M f_{D_j,\lambda_j^\ast}(D').
$$ \Comment{Global Estimates}
\Statex \hspace{-0.3in} \textbf{Output:} The global estimate $\overline{f}^{Ada}_{D,\vec{\lambda}^\ast}(D')$ for the query data $D'$.
%    \EndProcedure
  \end{algorithmic}
\end{breakablealgorithm}

\section*{Appendix B: Proofs}\label{Sec.Proofs}
In this part, we use the well-developed integral operator approach \citep{smale2007learning,lin2017distributed,rudi2015less} to prove our main results. Our main novelty in the proof is the detailed analysis of the role of the weight $\frac{|D_j|}{|D|}$ in \eqref{DKRR} and a tight bound on local approximation. Our proofs are divided into four steps: error decomposition, local approximation and global approximation, generalization error of DKRR, and generalization error of AdaDKRR.

\subsection{Error decomposition based on integral operators}
Let $D$ be a set of data drawn i.i.d. according to a distribution $\rho$.
Denote by  $S_{D}:\mathcal H_K\rightarrow\mathbb R^{|D|}$ the sampling operator on $\mathcal H_K$
$$
         S_{D}f:=\{f(x_i)\}_{(x_i,y_i)\in D}.
$$
Its scaled adjoint $S_{D}^T:\mathbb R^{|D|}\rightarrow \mathcal H_K$ is
$$
       S_{D}^T{\bf c}:=\frac1{|D|}\sum_{(x_i,y_i)\in D}c_iK_{x_i},\qquad {\bf c}\in\mathbb R^{|D|}.
$$
 Let
$L_{K,D}$ be the empirical version of $L_K$ and it is defined by
$$
         L_{K,D}f:=S_{D}^TS_{D }f=\frac1{|D|}\sum_{(x,y)\in D}f(x)K_x.
$$
Then we have \citep{smale2005shannon}
\begin{equation}\label{KRR:operator}
         f_{D,\lambda}=(L_{K,D}+\lambda I)^{-1}S^T_Dy_D,
\end{equation}
where $y_D:=(y_1,\dots,y_{|D|})^T$.

To present the error decomposition of DKRR, we need the following lemma that can be   easily deduced  from \cite[Prop.4]{guo2017learning} or  \cite[Prop.5]{chang2017distributed}.
\begin{lemma}\label{Lemma: dis error decomposition}
 Let $\overline{f}_{D,\vec{\lambda}}$ be defined by (\ref{DKRR}).
 We have
 \begin{eqnarray}\label{dis error decomp rho}
         E\left[\|\overline{f}_{D,\vec{\lambda}}-f_\rho\|_\rho^2\right]
         \leq
         2\sum_{j=1}^m\frac{|D_j|^2}{|D|^2}E\left[ \left\| f_{D_j,\lambda_j}
         -f^{\diamond}_{D_j,\lambda_j} \right\|_\rho^2\right]
         + 2\sum_{j=1}^m\frac{|D_j|}{|D|}
        E\left[\left\|f^{\diamond}_{D_j,\lambda_j}-f_{\rho}\right\|_\rho^2\right],
\end{eqnarray}
where
\begin{eqnarray}\label{noise free version}
  f^{\diamond}_{D,\lambda} := (L_{K,D}+\lambda
  I)^{-1}L_{K,D}f_\rho
\end{eqnarray}
is the noise-free version of $f_{D,\lambda}$.
\end{lemma}

With the help of the above lemma, we can derive the following error decomposition for DKRR based on integral operators.

\begin{proposition}\label{Prop:error-decomposition}
 Let $\overline{f}_{D,\vec{\lambda}}$ be defined by (\ref{DKRR}). If Assumption \ref{Assumption:regularity} holds with $\frac12\leq r\leq 1$, then
 \begin{eqnarray}\label{error-dec-dis}
         E\left[\|\overline{f}_{D,\vec{\lambda}}-f_\rho\|_\rho^2\right]
         &\leq&
         2\sum_{j=1}^m\frac{|D_j|}{|D|}\lambda^{2r}_jE\left[\mathcal Q_{D_j,\lambda}^{4r}\right]\|h_\rho\|_\rho^2
           +
         2\sum_{j=1}^m\frac{|D_j|^2}{|D|^2}E\left[\mathcal Q_{D_j,\lambda_j}^4\mathcal P^2_{D_j,\lambda_j}\right],
\end{eqnarray}
where
\begin{eqnarray}
          \mathcal Q_{D,\lambda}
          &:=&\left\|(L_{K,D}+\lambda I)^{-1/2}(L_{K}+\lambda I)^{1/2}\right\|, \label{Define Q}\\
           \mathcal P_{D,\lambda}&:=&
         \left\|(L_K+\lambda
          I)^{-1/2}(L_{K,D}f_\rho-S^T_{D}y_D)\right\|_K. \label{Define P}
          %\\
   %\mathcal U_{D,\lambda,g}&:=&
%   \left\|\left(L_{K} +\lambda I\right)^{-1/2} \left(\frac{1}{|D|} \sum_{z \in D}  g (z)K_x - E\left[ g(z)K_x\right]\right)\right\|_K,\label{Define U}
\end{eqnarray}
\end{proposition}

\proof{Proof.}
 Due to (\ref{KRR:operator}) and (\ref{noise free version}), we obtain
% \begin{eqnarray}\label{data-free-dif}
%   &&f_{D,\lambda}-f_\lambda
%   =
%   (L_{K,D}+\lambda I)^{-1}S_D^Ty_D
%   -
%   (L_K+\lambda I)^{-1}L_Kf_\rho \nonumber \\
%   &=&
%     (L_{K,D}+\lambda I)^{-1} (S^T_Dy_D-L_Kf_\rho)
%     -
%   (L_{K,D}+\lambda I)^{-1} ( L_{K,D}-L_K)(f_\lambda-f_\rho) \nonumber \\
%   &-&
%   (L_{K,D}+\lambda I)^{-1} ( L_{K,D}-L_K)f_\rho.
% \end{eqnarray}
% and
\begin{eqnarray}\label{noise-free-diff}
    f^\diamond_{D,\lambda}-f_\rho
  =
   ((L_{K,D}+\lambda I)^{-1}L_{K,D}-I)f_\rho=
  \lambda(L_{K,D}+\lambda I)^{-1}f_\rho
\end{eqnarray}
and
\begin{equation}\label{data-free-dif}
    f^\diamond_{D,\lambda}-f_{D,\lambda}=(L_{K,D}+\lambda I)^{-1}(L_{K,D}f_\rho-S_D^Ty_D).
\end{equation}
Then, we get from (\ref{Define Q}), (\ref{Define P}), (\ref{data-free-dif}), (\ref{noise-free-diff}) with $D=D_j$, $\lambda=\lambda_j$, and Assumption \ref{Assumption:regularity} with $\frac12\leq r\leq 1$ that for any $1\leq j\leq m$, there holds
\begin{eqnarray*}
     &&\|f^\diamond_{D_j,\lambda_j}-f_\rho\|_\rho
      \leq\lambda_j\|(L_K+\lambda_jI)^{1/2}(L_{K,D_j}+\lambda_j I)^{-1}f_\rho\|_K\\
      &\leq&
       \lambda_j \mathcal Q_{D_j,\lambda_j}\|(L_{K,D_j}+\lambda_j I)^{-1/2}L_K^{r-1/2}\|\|h_\rho\|_\rho
      \leq \lambda^{r}_j\mathcal Q_{D_j,\lambda}^{2r}\|h_\rho\|_\rho
\end{eqnarray*}
and
\begin{eqnarray*}
      \|f_{D_j,\lambda_j}-f^{\diamond}_{D_j,\lambda_j}\|_\rho
     \leq
     \|(L_K+\lambda_j I)^{1/2}(L_{K,D_j}+\lambda_j I)^{-1} (S^T_{D_j}y_{D_j}-L_{K,D}f_\rho)\|_K
    \leq
   \mathcal Q_{D_j,\lambda_j}^2\mathcal P_{D_j,\lambda_j}.
\end{eqnarray*}
Plugging the above two estimates into (\ref{dis error decomp rho}), we then obtain
 \begin{eqnarray*}
         \frac12E\left[\|\overline{f}_{D,\vec{\lambda}}-f_\rho\|_\rho^2\right]
         \leq
         \sum_{j=1}^m\frac{|D_j|}{|D|}\lambda^{2r}_jE\left[\mathcal Q_{D_j,\lambda}^{4r}\right]\|h_\rho\|_\rho^2
           +
         \sum_{j=1}^m\frac{|D_j|^2}{|D|^2}E\left[\mathcal Q_{D_j,\lambda_j}^4\mathcal P^2_{D_j,\lambda_j}\right].
\end{eqnarray*}
This completes the proof of Proposition \ref{Prop:error-decomposition}.
\endproof

We next derive an error decomposition for AdaDKRR in the following proposition.

\begin{proposition}\label{Prop.error-dec-adadkrr}
For any $\delta>0$ and any $\lambda\in\Lambda$, if Assumption \ref{Assumption:bounded for output} holds and $|D_j^{tr}|\sim|D_j^{val}|$, then
\begin{eqnarray}
    && E\left[ \left\| \overline{f}^{Ada}_{D,\vec{\lambda^*}}-f_\rho\right\|_\rho^2 \right]
      \leq
  \frac{C'_1(1+\log|\Lambda|)m}{|D|}  \nonumber\\
  &+&\hspace{-0.09in} 4\min_{\lambda\in\Lambda}\hspace{-0.04in}
   \left\{\hspace{-0.03in}\sum_{j=1}^m\frac{|D_j^{tr}|}{|D^{tr}|}   E\hspace{-0.05in}\left[\left\|{f}^{global}_{D^{tr}, {\lambda},n,\mu}\hspace{-0.05in}-\hspace{-0.05in}\sum_{j=1}^m\frac{|D_j^{tr}|}{|D^{tr|}}f_{D_j,{\lambda}}\right\|_\rho^2\right]
   \hspace{-0.05in} + \hspace{-0.05in} E\hspace{-0.05in}\left[\left\|\sum_{j=1}^m\frac{|D_j^{tr}|}{|D^{tr|}}f_{D_j,{\lambda}}\hspace{-0.05in}-\hspace{-0.05in} f_\rho\right\|_\rho^2\right]\right\}, \label{error-dec-adadkrr}
\end{eqnarray}
where $C_1'$ is an absolute constant.
\end{proposition}

To prove the above proposition, we need the following lemma which can be found in \citep[Theorem 7.1]{gyorfi2002distribution} (see also \citep{caponnetto2010cross} for a probabilistic argument).
\begin{lemma}\label{Lemma:CV-lemma}
 For a given data set $D$, let $D^{tr}$ and $D^{val}$ be the training set and validation set respectively. Under Assumption \ref{Assumption:bounded for output}, for any $\delta>0$ and any $\lambda\in\Lambda$, there holds
$$
 E\left[ \left\|\pi_Mf_{D^{tr},\textcolor{red}{\hat{\lambda}}}-f_\rho\right\|_\rho^2\right]
  \leq (1+\delta)\min_{\lambda\in\Lambda}E\left[ \left\|\pi_Mf_{D^{tr},\lambda}-f_\rho \right\|_\rho^2\right]
  +C'\frac{1+\log|\Lambda|}{|D^{val}|},
$$
where $C'=M^2(16/\delta+35+19\delta)$.
\end{lemma}

With the help of the above lemma, we can prove Proposition \ref{Prop.error-dec-adadkrr} as follows.

\proof{Proof of Proposition \ref{Prop.error-dec-adadkrr}.}
Due to \eqref{AdaDKRR}, we have from Lemma \ref{Lemma:CV-lemma}, $|D_j^{val}|\sim|D_j^{tr}|$, \eqref{def.optimal-parameter}, and Jensen's inequality that
\begin{eqnarray*}
 && E\left[ \left\| \overline{f}^{Ada}_{D,\vec{\lambda^*}}-f_\rho\right\|_\rho^2 \right]
     \leq
    \sum_{j=1}^m\frac{|D_j^{tr}|}{|D^{tr}|}E\left[\left\|\pi_M{f}^{global}_{D^{tr}, \lambda^*_j,n,\mu}-f_\rho\right\|_\rho^2 \right]\\
    &\leq&
    \sum_{j=1}^m\frac{|D_j^{tr}|}{|D^{tr}|}\left((1+\delta)\min_{\lambda\in\Lambda}E\left[\left\|\pi_M{f}^{global}_{D^{tr}, \lambda,n,\mu}-f_\rho\right\|_\rho^2\right]+\frac{C'(1+\log|\Lambda|)}{|D_j^{val}|}\right)\\
    &\leq&
    (1+\delta)\sum_{j=1}^m\frac{|D_j^{tr}|}{|D^{tr}|} \min_{\lambda\in\Lambda}E\left[\left\|{f}^{global}_{D^{tr}, \lambda,n,\mu}-f_\rho\right\|_\rho^2\right]+\frac{C'_1(1+\log|\Lambda|)m}{{|D|}}\\
    &\leq&
   \frac{C'_1m(1+\log|\Lambda|)}{|D|} + 2(1+\delta)\min_{\lambda\in\Lambda}
   \left\{\sum_{j=1}^m\frac{|D_j^{tr}|}{|D^{tr}|}   E\left[\left\|{f}^{global}_{D^{tr}, {\lambda},n,\mu}-\sum_{j=1}^m\frac{|D_j^{tr}|}{|D^{tr|}}f_{D_j,{\lambda}}\right\|_\rho^2\right] \right.\\
   &&
    \left.+ E\left[\left\|\sum_{j=1}^m\frac{|D_j^{tr}|}{|D^{tr|}}f_{D_j,{\lambda}}- f_\rho\right\|_\rho^2\right]\right\},
\end{eqnarray*}
 where  we use $(a+b)^2\leq 2a^2+2b^2$ in the last inequality and $C'$, $C_1$ are absolute constants. This completes the proof of Proposition \ref{Prop.error-dec-adadkrr} by setting $\delta=1$.
\endproof

\subsection{Local approximation and global approximation}
Due to Proposition \ref{Prop.error-dec-adadkrr}, it is crucial to derive the error of the global approximation \eqref{global approximation} as well as the local approximation \eqref{local-approximation}. In this appendix only, we denote by ${f}^{loc}_{D,\lambda,\mu}$ the local approximation \eqref{local-approximation} with $D_j^{tr}=D$, $s=|D|$, and $\lambda_\ell=\lambda$ for the sake of brevity. The derived error of local approximation is shown in the following proposition.

\begin{proposition}\label{Proposition:Error-local-app}
If (\ref{regularitycondition}) holds with $1/2\leq r\leq 1$, then we
have
\begin{eqnarray}\label{Error-decomposition-detailed-local}
     &&\left\| {f}^{loc}_{D,\lambda,\mu}-f_{D,\lambda}\right\|_\rho
     \leq
     \mu^r\left\|h_\rho\right\|_\rho\left(\left(\mathcal Q_{D,\mu}\mathcal Q_{D,\mu}^*+1\right)\mathcal
     Q_{\Xi_n,\mu}^{2r}+\mathcal Q_{D,\mu}^{2r}\right) \nonumber\\
     &+&
     \mu^{1/2}\left(\lambda^{-1/2}\mathcal Q^2_{D,\lambda}\mathcal P_{D,\lambda}+\lambda^{r-1/2}\mathcal Q_{D,\lambda}^{2r-1}\|h\|_\rho\right)\left(\mathcal Q_{D,\mu} +\left(\mathcal Q_{D,\mu}\mathcal Q_{D,\mu}^*+1\right)\mathcal Q_{\Xi_n,\mu}\right),
\end{eqnarray}
where $\|\cdot\|$ denotes the operator norm and
\begin{equation}\label{def.Qstar}
    \mathcal Q_{D,\mu}^*:=\left\|(L_K+\mu I)^{-1/2}(L_{K,D}+\mu I)^{1/2}\right\|.
\end{equation}
\end{proposition}

Before providing the proof of the above proposition, we introduce several interesting tools.
Let $P_{\Xi_n}$ be the projection from $\mathcal H_K$ to $
B_{n,K}$.
Then for an arbitrary $\nu>0$, there holds
\begin{equation}\label{projection-p-1}
      (I-P_{\Xi_n})^\nu=I-P_{\Xi_n}.
\end{equation}
Let $ T_{\Xi_n}:  B_{n,K}\rightarrow \mathbb R^{n}$ be a sampling operator defined by $T_{\Xi_n}f:=\{f(x_i)\}_{x_i \in \Xi_n}$ such that the range of its adjoint operator $ T^T_{\Xi_n}$ is exactly in $B_{n,K}$, and let
$
       U\Sigma V^T
$
be the SVD of $  T_{\Xi_n}$. Then we have
\begin{equation}\label{VD-property}
   V^TV=I,\qquad   V V^T=P_{\Xi_n}.
\end{equation}
Write
\begin{equation}\label{spetral definetion}
     g_{\Xi_n,\mu}(L_{K,D}):=V(V^TL_{K,D}V+\mu
     I)^{-1}V^T.
\end{equation}
Then it can be found in \eqref{Analytic-solution1} and \citep{rudi2015less}
that
\begin{equation}\label{Nystrom operator}
        {f}^{loc}_{D,\lambda,\mu}=g_{\Xi_n,\mu}(L_{K,D})L_{K,D}f_{D,\lambda}.
\end{equation}
  For an arbitrary bounded linear operator $B$, it follows from  \eqref{VD-property} and \eqref{spetral definetion} that
\begin{eqnarray}\label{Important1}
     &&g_{\Xi_n,\mu}(L_{K,D})(L_{K,D}+\mu
     I)VBV^T
      =
     VBV^T.
\end{eqnarray}
Inserting $B=(V^TL_{K,D}V+\mu I)^{-1}$ into (\ref{Important1}), it is easy to derive  \citep{sun2021nystr}
\begin{equation}\label{Important2}
    \|(L_{K,D}+\mu
      I)^{1/2}g_{\Xi_n,\mu}(L_{K,D})(L_{K,D}+\mu
      I)^{1/2}\|\leq  1.
\end{equation}

Besides the above tools, we also  need
the following two lemmas
that can be found in \cite[Proposition 3]{rudi2015less} and \cite[Proposition 6]{rudi2015less}, respectively.

\begin{lemma}\label{Lemma:Projection general}
Let $\mathcal H$, $\mathcal K$, and $\mathcal F$ be three separable
Hilbert spaces. Let $Z:\mathcal H\rightarrow\mathcal K$ be a bounded
linear operator and $P$ be a projection operator on $\mathcal H$
such that $\mbox{range}P=\overline{\mbox{range}Z^T}$. Then for any
bounded linear operator $F:\mathcal F\rightarrow\mathcal H$ and any
$\lambda>0$, we have
$$
     \|(I-P)F\|\leq\lambda^{1/2}\left\|(Z^TZ+\lambda I)^{-1/2}F\right\|.
$$
\end{lemma}

\begin{lemma}\label{Lemma:operator inequality general}
Let $\mathcal H$ and $\mathcal K$ be two separable Hilbert spaces,
$A:\mathcal H\rightarrow\mathcal H$ be a positive linear operator,
$V_{\mathcal H,\mathcal K}:\mathcal H\rightarrow\mathcal K$ be a partial isometry, and
$B:\mathcal K\rightarrow\mathcal K$ be a bounded operator. Then for
all $0\leq r^*,s^*\leq1/2$, there holds
$$
      \left\|A^{r^*}V_{\mathcal H,\mathcal K}BV_{\mathcal H,\mathcal K}^TA^{s^*}\right\|\leq\left\|(V_{\mathcal H,\mathcal K}^TAV_{H,K})^{r^*}B(V_{\mathcal H,\mathcal K}^TAV_{\mathcal H,\mathcal K})^{s^*}\right\|.
$$
\end{lemma}

With the help of the above lemmas and the important properties of $g_{D_m,\mu}$ in \eqref{Important1} and \eqref{Important2}, we prove Proposition \ref{Proposition:Error-local-app} as follows.

\proof{Proof of Proposition \ref{Proposition:Error-local-app}.} The triangle inequality yields
\begin{eqnarray}\label{Error-decomposition-local-app}
   \|f^{loc}_{D,\lambda,\mu}-f_{D,\lambda}\|_\rho
   \leq
  \mathcal A_n(D,\lambda,\mu)+\mathcal C_n(D,\lambda,\mu),
\end{eqnarray}
where
\begin{eqnarray*}
   \mathcal A_n(D,\lambda,\mu)&=&\left\|(g_{\Xi_n,\mu}(L_{K,D})L_{K,D}-I)P_{\Xi_n}f_{D,\lambda}\right\|_\rho,\\
  \mathcal C_n(D,\lambda,\mu)&=&\left\|(g_{\Xi_n,\mu}(L_{K,D})L_{K,D}-I)(I-P_{\Xi_n})f_{D,\lambda}\right\|_\rho.
\end{eqnarray*}
% We at first use (\ref{Important1}) and (\ref{Important2}) to bound $\mathcal S(D,\lambda,m)$.
% % Due to the well known Codes inequality \citep{bhatia2013matrix}
% % \begin{equation}\label{Codes inequality}
% %          \|A^u B^u\|\le\|AB\|^u, \qquad 0<u\leq 1
% % \end{equation}
% for arbitrary positive operators $A,B$, we have
% \begin{equation}\label{Codes-for-Q}
%      \|(L_K+\lambda I)^{1/2}(L_{K,D}+\lambda I)^{-1/2}\|
%      \leq\mathcal Q_{D,\lambda}^{1/2}.
% \end{equation}
% Then, it follows from   (\ref{Important2}) and $\|f\|_\rho=\|L_K^{1/2}f\|_K$ for any $f\in L_{\rho_X}^2$ that
% \begin{eqnarray}\label{sample error estimate}
%     &&\mathcal S(D,\lambda,m)
%     =\|g_{D_m,\lambda}(L_{K,D})(S_D^Ty_D-L_{K,D}f_\rho)\|_\rho\nonumber\\
%     &=&\|L_K^{1/2}g_{D_m,\lambda}(L_{K,D})(S_D^Ty_D-L_{K,D}f_\rho)\|_K\nonumber \\
%     &\leq&
%     \|(L_K+\lambda I)^{1/2}( {L_{K,D}}+\lambda I)^{-1/2}\|^2
%     \|(L_{K,D}+\lambda I)^{1/2}g_{D_m,\lambda}(L_{K,D})(L_{K,D}+\lambda I)^{1/2}\|\nonumber\\
%     &\times&
%     \|(L_K+\lambda I)^{-1/2}
%     (S_D^Ty_D-L_{K,D}f_\rho)\|_K\nonumber\\
%     &\leq&
%   {\mathcal Q_{D,\lambda}}\mathcal P_{D,\lambda}.
% \end{eqnarray}
We first bound $\mathcal A_n(D,\lambda,\mu)$. It follows from   (\ref{Important1}) with $B=I$ that
$$
    P_{\Xi_n}=g_{\Xi_n,\mu}(L_{K,D})(L_{K,D}+\mu   I)P_{\Xi_n},
$$
which together with the definition of $\mathcal A_n(D,\lambda,\mu)$ yields
$$
 \mathcal A_n(D,\lambda,\mu)=\mu\| g_{\Xi_n,\mu}(L_{K,D}) P_{\Xi_n}f_{D,\lambda}\|_\rho.
$$
Then we have from the triangle inequality that
\begin{eqnarray}\label{A1-dec}
    \mathcal A_n(D,\lambda,\mu)
    &\leq&
    \mu\left\| g_{\Xi_n,\mu}(L_{K,D}) P_{\Xi_n}f_\rho\right\|_\rho+\mu\left\| g_{\Xi_n,\mu}(L_{K,D}) P_{\Xi_n}(f_{D,\lambda}-f_\rho)\right\|_\rho\nonumber\\
    &=:&
    \mathcal A_{n,1}(D,\mu)+\mathcal A_{n,2}(D,\lambda,\mu).
\end{eqnarray}
To bound $\mathcal A_{n,1}(D,\mu)$,
noting that $L_{K,D}$ is a positive operator, we have $\|(V^T(L_{K,D}+\mu
     I)V)^{r-1}\|\leq\mu^{r-1}$ for $r\leq 1$ \citep{rudi2015less}.
Recalling further (\ref{VD-property}) and the Cordes inequality \citep{bhatia2013matrix}
\begin{equation}\label{Codes inequality}
         \|A^u B^u\|\le\|AB\|^u, \qquad 0<u\leq 1
\end{equation}
for arbitrary positive operators $A$ and $B$, we get from Lemma
\ref{Lemma:operator inequality general} with $A=(L_{K,D}+\mu
I)$, $V_{\mathcal H,\mathcal K}=V$, $B=(V^TL_{K,D}V+\mu I)^{-1}$,
$r^*=1/2$, and $s^*=r-1/2$ that
\begin{eqnarray}\label{boundAn-1}
     &&\mathcal A_{n,1}(D,\mu)
      \leq
     \mu\left\|L_K^{1/2}g_{ {\Xi_n},\mu}(L_{K,D})VV^TL_K^{r-1/2}\right\|\left\|h_\rho\right\|_\rho \nonumber\\
     &\leq&
     \mu\mathcal Q_{D,\mu}^{2r}\left\|h_\rho\right\|_\rho
     \left\|(L_{K,D}+\mu I)^{1/2}g_{ {\Xi_n},\mu}(L_{K,D})VV^T(L_{K,D}+\mu I)^{r-1/2}\right\| \nonumber\\
     &\leq&
     \mu\mathcal Q_{D,\mu}^{2r}\left\|(V^T(L_{K,D}+\mu
     I)V)^{1/2}(V^T(L_{K,D}+\mu
     I)V)^{-1}
   (V^T(L_{K,D}+\mu
     I)V)^{r-1/2}\right\|\left\|h_\rho\right\|_\rho\nonumber\\
     &\leq&
      \mu\mathcal Q_{D,\mu}^{2r}\left\|(V^T(L_{K,D}+\mu
     I))^{r-1}V\right\|\left\|h_\rho\right\|_\rho\nonumber\\
     &\leq&
     \mu^r\mathcal Q_{D,\mu}^{2r}\left\|h_\rho\right\|_\rho.
\end{eqnarray}
To bound $\mathcal A_{n,2}(D,\lambda,\mu)$, we need the following standard estimate of $\|f_{D,\lambda}-f_\rho\|_K$
% , which is standard in the existing literature
\citep{caponnetto2007optimal,steinwart2009optimal,lin2017distributed,chang2017distributed}, which can be derived in a way similar to that used to prove Proposition \ref{Prop:error-decomposition}.
Under \eqref{regularitycondition} with $r\leq 1$, there holds
\begin{equation}\label{HK-bound-KRR}
    \|f_{D,\lambda}-f_\rho\|_K
    \leq \lambda^{-1/2}\mathcal Q^2_{D,\lambda}\mathcal P_{D,\lambda}+\lambda^{r-1/2}\mathcal Q_{D,\lambda}^{2r-1}\|h\|_\rho.
\end{equation}
Then, it follows from (\ref{VD-property}), \eqref{Codes inequality}, and Lemma
\ref{Lemma:operator inequality general} with $A=(L_{K,D}+\mu
I)$, $V_{H,K}=V$, $B=(V^TL_{K,D}V+\mu I)^{-1}$,
$r^*=1/2$, and $s^*=0$ that
\begin{eqnarray}\label{boundAn-2}
     &&\mathcal A_{n,2}(D,\lambda,\mu)
      \leq
     \mu \left\|L_K^{1/2}g_{ {\Xi_n},\mu}(L_{K,D})VV^T \right\| \left\|f_{D,\lambda}-f_\rho \right\|_K \nonumber\\
     &\leq&
     \mu\mathcal Q_{D,\mu}  \left(\lambda^{-1/2}\mathcal Q^2_{D,\lambda}\mathcal P_{D,\lambda}+\lambda^{r-1/2}\mathcal Q_{D,\lambda}^{2r-1}\|h\|_\rho\right)
     \left\|(L_{K,D}+\mu I)^{1/2}g_{ \Xi_n,\mu}(L_{K,D})VV^T\right\|  \nonumber\\
     &\leq&
    \mu\mathcal Q_{D,\mu}  \left(\lambda^{-1/2}\mathcal Q^2_{D,\lambda}\mathcal P_{D,\lambda}+\lambda^{r-1/2}\mathcal Q_{D,\lambda}^{2r-1}\|h\|_\rho\right)\left\|(V^T(L_{K,D}+\mu I)V)^{-1/2} \right\| \nonumber\\
     &\leq&
     \mu^{1/2}\mathcal Q_{D,\mu}  \left(\lambda^{-1/2}\mathcal Q^2_{D,\lambda}\mathcal P_{D,\lambda}+\lambda^{r-1/2}\mathcal Q_{D,\lambda}^{2r-1}\|h\|_\rho\right).
\end{eqnarray}
Plugging \eqref{boundAn-1} and \eqref{boundAn-2} into \eqref{A1-dec}, we obtain
\begin{equation}\label{bound-A1}
    \mathcal A_n(D,\lambda,\mu)
    \leq
    \mu^r\mathcal Q_{D,\mu}^{2r}\|h_\rho\|_\rho
    +\mu^{1/2}\mathcal Q_{D,\mu}  \left(\lambda^{-1/2}\mathcal Q^2_{D,\lambda}\mathcal P_{D,\lambda}+\lambda^{r-1/2}\mathcal Q_{D,\lambda}^{2r-1}\|h\|_\rho\right).
\end{equation}
We next bound $\mathcal C_n(D,\lambda,\mu)$. Similar to the above, we have
\begin{eqnarray}\label{cn-error-dec}
  \mathcal C_n(D,\lambda,\mu)&\leq &  \left\|(g_{\Xi_n,\mu}(L_{K,D})L_{K,D}-I)(I-P_{\Xi_n})f_\rho\right\|_\rho   \nonumber\\
  &+&\left\|(g_{\Xi_n,\mu}(L_{K,D})L_{K,D}-I)(I-P_{\Xi_n})(f_{D,\lambda}-f_\rho)\right\|_\rho\nonumber\\
  &=:&
  \mathcal C_{n,1}(D,\mu)+\mathcal C_{n,2}(D,\lambda,\mu).
\end{eqnarray}
Due to Lemma \ref{Lemma:Projection general}, we have
\begin{eqnarray}\label{projection-bound}
        \left\|(I-P_{\Xi_n})(L_K+\mu
        I)^{1/2}\right\|
       \leq
        \mu^{1/2}\left\|(L_{K,\Xi_n}+\mu I)^{-1/2}(L_{K}+\mu
        I)^{1/2}\right\|\leq \mu^{1/2}\mathcal Q_{\Xi_n,\mu}.
\end{eqnarray}
Then, it follows from (\ref{Important2}), (\ref{Codes inequality}), and \eqref{projection-p-1} with $\tau=2r$
that
\begin{eqnarray}\label{bound-CND-1}
     &&\mathcal C_{n,1}(D,\mu)
      \leq
     \left\|L_K^{1/2}g_{\Xi_n,\mu}(L_{K,D})L_{K,D}(I-P_{\Xi_n})L_K^{r-1/2}\right\|\left\|h_\rho\right\|_\rho\nonumber\\
     &+&\left\|L_K^{1/2}(I-P_{\Xi_n})L_K^{r-1/2}\right\| \left\|h_\rho\right\|_\rho\nonumber\\
     &\leq&
     \mathcal Q_{D,\mu}\mathcal Q_{D,\mu}^*\left\|(L_{K,D}+\mu I)^{1/2}g_{\Xi_n,\mu}(L_{K,D})(L_{K,D}+\mu
     I)^{1/2}\right\|\nonumber\\
     &\times&\left\|(L_{K}+\mu
     I)^{1/2}(I-P_{\Xi_n})^{2r}L_K^{r-1/2}\right\|
      \left\|h_\rho\right\|_\rho\nonumber\\
      &+&
      \left\|(L_{K}+\mu
     I)^{1/2}(I-P_{\Xi_n})^{2r}L_K^{r-1/2}\right\|
      \left\|h_\rho\right\|_\rho \nonumber\\
     &\leq&
       \left(\mathcal Q_{D,\mu}\mathcal Q_{D,\mu}^*+1\right)\left\|h_\rho\right\|_\rho
       \left\|(L_K+\mu I)^{1/2}(I-P_{\Xi_n})\right\|\left\|(I-P_{\Xi_n})^{2r-1}
     L_K^{r-1/2}\right\|\nonumber\\
     &\leq&
     \left(\mathcal Q_{D,\mu}\mathcal Q_{D,\mu}^*+1\right)\left\|h_\rho\right\|_\rho\mu^{r}\mathcal
     Q_{\Xi_n,\mu}^{2r}.
\end{eqnarray}
Noting \eqref{HK-bound-KRR}, we get from \eqref{projection-p-1}, (\ref{Important2}),  (\ref{Codes inequality}), and \eqref{projection-bound} again that
\begin{eqnarray}\label{bound-CND-2}
     &&\mathcal C_{n,2}(D,\lambda,\mu)\\
     & \leq&
     \left\|L_K^{1/2}g_{\Xi_n,\mu}(L_{K,D})L_{K,D}(I-P_{\Xi_n})\right\|\left(\lambda^{-1/2}\mathcal Q^2_{D,\lambda}\mathcal P_{D,\lambda}+\lambda^{r-1/2}\mathcal Q_{D,\lambda}^{2r-1}\|h\|_\rho\right)\nonumber\\
     &+&\left\|L_K^{1/2}(I-P_{\Xi_n})\right\| \left(\lambda^{-1/2}\mathcal Q^2_{D,\lambda}\mathcal P_{D,\lambda}+\lambda^{r-1/2}\mathcal Q_{D,\lambda}^{2r-1}\|h\|_\rho \right)\nonumber\\
     &\leq&
     \mathcal Q_{D,\mu}\mathcal Q_{D,\mu}^*\left\|(L_{K,D}+\mu I)^{1/2}g_{\Xi_n,\mu}(L_{K,D})(L_{K,D}+\mu I)^{1/2}\right\|\nonumber\\
     &\times&\left\|(L_{K}+\mu
     I)^{1/2}(I-P_{\Xi_n})\right\|
     \left(\lambda^{-1/2}\mathcal Q^2_{D,\lambda}\mathcal P_{D,\lambda}+\lambda^{r-1/2}\mathcal Q_{D,\lambda}^{2r-1}\|h\|_\rho \right)\nonumber\\
      &+&
      \left\|(L_{K}+\mu  I)^{1/2}(I-P_{\Xi_n})\right\|
     \left(\lambda^{-1/2}\mathcal Q^2_{D,\lambda}\mathcal P_{D,\lambda}+\lambda^{r-1/2}\mathcal Q_{D,\lambda}^{2r-1}\|h\|_\rho \right) \nonumber\\
     &\leq&
       \left(\mathcal Q_{D,\mu}\mathcal Q_{D,\mu}^*+1\right)\left(\lambda^{-1/2}\mathcal Q^2_{D,\lambda}\mathcal P_{D,\lambda}+\lambda^{r-1/2}\mathcal Q_{D,\lambda}^{2r-1}\|h\|_\rho\right)
       \left\|(L_K+\mu I)^{1/2}(I-P_{\Xi_n})\right\| \nonumber\\
     &\leq&
     \left(\mathcal Q_{D,\mu}\mathcal Q_{D,\mu}^*+1\right)\left(\lambda^{-1/2}\mathcal Q^2_{D,\lambda}\mathcal P_{D,\lambda}+\lambda^{r-1/2}\mathcal Q_{D,\lambda}^{2r-1}\|h\|_\rho\right)  \mu^{1/2}\mathcal
     Q_{\Xi_n,\mu}.
\end{eqnarray}
Inserting \eqref{bound-CND-1} and \eqref{bound-CND-2} into \eqref{cn-error-dec}, we have
\begin{eqnarray}\label{Bound-CDN}
     \mathcal C_n(D,\lambda,\mu)
     &\leq&
     \left(\mathcal Q_{D,\mu}\mathcal Q_{D,\mu}^*+1\right)\left(\lambda^{-1/2}\mathcal Q^2_{D,\lambda}\mathcal P_{D,\lambda}+\lambda^{r-1/2}\mathcal Q_{D,\lambda}^{2r-1}\|h\|_\rho\right)  \mu^{1/2}\mathcal
     Q_{\Xi_n,\mu}   \nonumber \\
     &+& \left(\mathcal Q_{D,\mu}\mathcal Q_{D,\mu}^*+1\right)\|h_\rho\|_\rho\mu^{r}\mathcal
     Q_{\Xi_n,\mu}^{2r}.
\end{eqnarray}
Plugging \eqref{bound-A1} and \eqref{Bound-CDN} into \eqref{Error-decomposition-local-app}, we get
\begin{eqnarray*}
    &&\left\| {f}^{loc}_{D,\lambda,\mu}-f_{D,\lambda}\right\|_\rho
     \leq
     \mu^r\left\|h_\rho\right\|_\rho\left(\left(\mathcal Q_{D,\mu}\mathcal Q_{D,\mu}^*+1\right)\mathcal
     Q_{\Xi_n,\mu}^{2r}+\mathcal Q_{D,\mu}^{2r}\right) \\
     &+&
     \mu^{1/2}\left(\lambda^{-1/2}\mathcal Q^2_{D,\lambda}\mathcal P_{D,\lambda}+\lambda^{r-1/2}\mathcal Q_{D,\lambda}^{2r-1}\|h\|_\rho\right)\left(\mathcal Q_{D,\mu} +\left(\mathcal Q_{D,\mu}\mathcal Q_{D,\mu}^*+1\right)\mathcal Q_{\Xi_n,\mu}\right).
\end{eqnarray*}
This completes the proof of Proposition \ref{Proposition:Error-local-app}.
\endproof

Based on Proposition \ref{Proposition:Error-local-app}, we can derive an error estimate for the global approximation directly.

\begin{proposition}\label{Prop:global-app-err}
If (\ref{regularitycondition}) holds with $1/2\leq r\leq 1$, then
\begin{eqnarray*}
&& \left\|{f}^{global}_{D^{tr}, \lambda,\mu}-\sum_{j=1}^m\frac{|D_j^{tr}|}{|D^{tr}|}f_{D_j,\lambda}
 \right\|_\rho
 \leq
 \sum_{j=1}^m\frac{|D_j^{tr}|}{|D^{tr}|}\left[\mu^r\left\|h_\rho\right\|_\rho\left(\left(\mathcal Q_{D_j^{tr},\mu}\mathcal Q_{D_j^{tr},\mu}^*+1\right)\mathcal
     Q_{\Xi_n,\mu}^{2r}+\mathcal Q_{D_j^{tr},\mu}^{2r}\right)  \right.\\
     &&+
     \left.\mu^{1/2}\left(\lambda^{-1/2}\mathcal Q^2_{D_j^{tr},\lambda}\mathcal P_{D_j^{tr},\lambda}+\lambda^{r-1/2}\mathcal Q_{D_j^{tr},\lambda}^{2r-1}\left\|h\right\|_\rho\right)\left(\mathcal Q_{D_j^{tr},\mu} +\left(\mathcal Q_{D_j^{tr},\mu}\mathcal Q_{D_j^{tr},\mu}^*+1\right)\mathcal Q_{\Xi_n,\mu}\right)\right].
\end{eqnarray*}
\end{proposition}

\proof{Proof.}
It  follows from
\eqref{global approximation} and Jensen's inequality  that
\begin{eqnarray*}
   \left\|{f}^{global}_{D^{tr}, \lambda,\mu}-\sum_{j=1}^m\frac{|D_j^{tr}|}{|D^{tr}|}f_{D_j,\lambda}
 \right\|_\rho
   =\left\|\sum_{j=1}^m\frac{|D_j^{tr}|}{|D^{tr}|}\left({f}^{loc}_{D_j^{tr},\lambda,\mu}-f_{D_j,\lambda}\right)\right\|_\rho
   \leq
  \sum_{j=1}^m\frac{|D_j^{tr}|}{|D^{tr}|}\left\|{f}^{loc}_{D_j^{tr},\lambda,\mu}-f_{D_j,\lambda}\right\|_\rho.
\end{eqnarray*}
But
Proposition \ref{Proposition:Error-local-app} with $D=D_j^{tr}$ for $j=1,\dots,m$ yields
\begin{eqnarray*}
  &&\left\|{f}^{loc}_{D_j^{tr},\lambda,\mu}-f_{D_j^{tr},\lambda}\right\|_\rho
  \leq
  \mu^r\left\|h_\rho\right\|_\rho\left(\left(\mathcal Q_{D_j^{tr},\mu}\mathcal Q_{D_j^{tr},\mu}^*+1\right)\mathcal
     Q_{\Xi_n,\mu}^{2r}+\mathcal Q_{D_j^{tr},\mu}^{2r}\right)  \\
     &+&
     \mu^{1/2}\left(\lambda^{-1/2}\mathcal Q^2_{D_j^{tr},\lambda}\mathcal P_{D_j^{tr},\lambda}+\lambda^{r-1/2}\mathcal Q_{D_j^{tr},\lambda}^{2r-1}\|h\|_\rho\right)
     \left(\mathcal Q_{D_j^{tr},\mu} +\left(\mathcal Q_{D_j^{tr},\mu}\mathcal Q_{D_j^{tr},\mu}^*+1\right)\mathcal Q_{\Xi_n,\mu}\right).
\end{eqnarray*}
Combining the above two estimates proves Proposition \ref{Prop:global-app-err}.
\endproof

\subsection{Proof of Theorem \ref{Theorem:DKRR-1} and Corollary \ref{corollary:DKRR-1}}
To prove Theorem \ref{Theorem:DKRR-1}, it suffices to bound $E\left[\mathcal Q_{D_j,\lambda}^{4r}\right]$ and $E\left[\mathcal Q_{D_j,\lambda_j}^4\mathcal P^2_{D_j,\lambda_j}\right]$, which requires several auxiliary lemmas.
The first one focuses on bounding    $\mathcal P_{D,\lambda}$ derived in   \citep{caponnetto2007optimal,lin2017distributed}.

\begin{lemma}\label{Lemma:Bound-P}
Let $D$ be a set of samples drawn i.i.d. according to $\rho$ and $0<
\delta <1$. Under Assumption \ref{Assumption:bounded for output},  with confidence at least $1-\delta$, there holds
\begin{eqnarray}\label{Bound-P}
   \mathcal P_{D,\lambda}  &\leq&  2M(\kappa   +1) \left(\frac{1}{|D|\sqrt{ \lambda}}
           +\sqrt{\frac{\mathcal{N}(\lambda)}{|D|}}\right) \log\frac2\delta.
\end{eqnarray}
\end{lemma}

The second one aims to bounding
\begin{equation}\label{Define R}
    \mathcal R_{D,\lambda}
          :=\left\|(L_K+\lambda I)^{-1/2}(L_K-L_{K,D})(L_{K}+\lambda I)^{-1/2}\right\|
\end{equation}
that can be deduced from \citep[Lemma 6]{lin2021distributed}.

\begin{lemma}\label{Lemma:Bound-R}
Let $D$ be a set of samples drawn i.i.d. according to $\rho$. If $\frac1{|D|}<\lambda\leq1$ and $\mathcal
N(\lambda)\geq 1$, then
\begin{equation}\label{operator difference concentration 1}
         P\left[\mathcal R_{D,\lambda}\geq\frac14\right]
         \leq  4\exp\left\{-\frac{\sqrt{\lambda|D|}}{C_1^*(1+\log\mathcal N(\lambda))}\right\},
\end{equation}
where $C_1^*:=4\max\{(\kappa^2+1)/3,2\sqrt{\kappa^2+1}\}$.
\end{lemma}

Due to the definition of $\mathcal R_{D,\lambda}$, the bounds of $\mathcal Q_{D,\lambda}$ and $\mathcal Q^*_{D,\lambda}$ can be given in the following lemma, whose proof is rather standard.
\begin{lemma}\label{Lemma:Bound-Q}
If $\mathcal R_{D,\lambda}\leq 1/4$, there holds
\begin{equation}\label{Bound:Q}
    \mathcal Q_{D,\lambda}\leq\frac{2\sqrt{3}}3
\end{equation}
and
\begin{equation}\label{Bound:Q*}
     \mathcal Q^*_{D,\lambda}\leq\frac{\sqrt{6}}2.
\end{equation}
\end{lemma}

\proof{Proof.}
A direct computation yields
\begin{eqnarray*}
      &&(L_K+\lambda I)^{1/2}(L_{K,D}+\lambda I)^{-1}(L_K+\lambda I)^{1/2}\\
         &=& (L_K+\lambda I)^{1/2}[(L_{K,D}+\lambda I)^{-1}-(L_{K}+\lambda I)^{-1}]   (L_K+\lambda I)^{1/2}+I
          = I\\
          &+& (L_K+\lambda I)^{-1/2}(L_K-L_{K,D})(L_K+\lambda I)^{-1/2}(L_K+\lambda I)^{1/2}
          (L_{K,D}+\lambda I)^{-1}(L_K+\lambda I)^{1/2}.
\end{eqnarray*}
Thus,
\begin{eqnarray*}
   && \left\|(L_K+\lambda I)^{1/2}(L_{K,D}+\lambda I)^{-1}(L_K+\lambda I)^{1/2}\right\|\\
   &\leq& 1+ \frac14\left\|(L_K+\lambda I)^{1/2}(L_{K,D}+\lambda I)^{-1}(L_K+\lambda I)^{1/2}\right\|.
\end{eqnarray*}
This implies
$$
   \|(L_K+\lambda I)^{1/2}(L_{K,D}+\lambda I)^{-1}(L_K+\lambda I)^{1/2}\|\leq \frac43
$$
and \eqref{Bound:Q}. Then,
\begin{eqnarray*}
  &&\left\|(L_{K,D}+\lambda I)^{-1/2}
          (L_{K}-L_{K,D})(L_{K,D}+\lambda I)^{-1/2}\right\|
          \leq
          \mathcal Q_{D,\lambda}^2\mathcal R_{D,\lambda}
          \leq \frac13.
\end{eqnarray*}
But
\begin{eqnarray*}
      &&\left\|(L_{K,D}+\lambda I)^{1/2}(L_{K}+\lambda I)^{-1}(L_{K,D}+\lambda I)^{1/2}\right\|\\
          &\leq  &1
          + \frac13 \left\|(L_{K,D}+\lambda I)^{1/2}
          (L_{K}+\lambda I)^{-1}(L_{K,D}+\lambda I)^{1/2} \right\|.
\end{eqnarray*}
Therefore, we have
$$
  \left\| (L_{K,D}+\lambda I)^{1/2}(L_{K}+\lambda I)^{-1}(L_{K,D}+\lambda I)^{1/2} \right\| \leq \frac32.
$$
This completes the proof of Lemma \ref{Lemma:Bound-Q}.
\endproof

For further use, we also need the following probability to expectation formula \citep{lin2023sketching}.

\begin{lemma}\label{Lemma:prob-to-exp}
Let $0<\delta<1$, and let $\xi\in\mathbb R_+$ be a random variable. If $\xi\leq \mathcal A\log^b\frac{c}{\delta}$ holds with confidence $1-\delta$  for some $\mathcal A,b,c>0$, then
$$
      E[\xi]\leq c\Gamma(b+1) \mathcal A,
$$
where $\Gamma(\cdot)$ is the Gamma function.
\end{lemma}

% \proof{Proof of Lemma \ref{Lemma:prob-to-exp}.}
% Since $\xi\leq \mathcal A\log^b\frac{c}{\delta}$ holds with confidence $1-\delta$, we have
% $$
%     P[\xi>t]\leq c\exp\{\mathcal A^{-1/b}t^{1/b}\}.
% $$
% Using the probability to expectation formula
% \begin{equation}\label{expectation formula}
%               E[\xi] =\int_0^\infty P\left[\xi > t\right] d t
% \end{equation}
%   to the positive random variable $\xi$, we have
% \begin{eqnarray*}
%      E[\xi]\leq  c\int_{0}^\infty\exp\{\mathcal A^{-1/b}t^{1/b}\}
%      \leq c\mathcal A\Gamma(b+1).
% \end{eqnarray*}
% This completes the proof of Lemma \ref{Lemma:prob-to-exp}.
% \endproof

With the help of the above lemmas, we can derive the bounds of $E\left[\mathcal Q_{D,\lambda}^4\mathcal P_{D,\lambda}^2\right]$ and $ E\left[\mathcal Q_{D,\lambda}^{4r}\right]$ as follows.

\begin{lemma}\label{Lemma:bound-est-prod}
 Under Assumption \ref{Assumption:bounded for output}, if $\lambda |D|\geq (2vC_1^*(\log(1+\kappa)+2))^2\log^4|D| $ for some $v\geq 1$, then
\begin{eqnarray}
    E\left[\mathcal Q_{D,\lambda}^{2v}\mathcal P_{D,\lambda}^v\right]
    &\leq&
   C_0^*\left(\left(\frac{ 1 }{|D|\sqrt{\lambda}}    + \sqrt{ \frac{ \mathcal N(\lambda)}{|D|}}\right)^v+ |D|^{-v/2}\right),
   \label{Bound-dif-exp-2}\\
   E\left[\mathcal Q_{D,\lambda}^u\right]&\leq& 5(2+\kappa)^{u/2},\qquad\forall\ 0\leq u\leq 4v,
   \label{Bound-dif-exp-3}\\
   E[( \mathcal Q_{D,\lambda}^*)^u]
   &\leq&
   5(4+\kappa)^{u/2},\qquad\forall\ 0\leq u\leq 4v,\label{bound-dif-exp-4}
\end{eqnarray}
where $C_0^*:=2(2(\kappa+1)M)^v\max\{\gamma(v+1)2^v,2\kappa^v\} $.
\end{lemma}

\proof{Proof.} If $R_{D,\lambda}\leq 1/4$, it follows from Lemma \ref{Lemma:Bound-Q} and Lemma \ref{Lemma:Bound-P} that, with confidence $1-\delta$, there holds
$$
    \mathcal Q_{D,\lambda}^{2v} \mathcal P^v_{D,\lambda}
        \leq
        \left(4M(\kappa   +1) \left(\frac{ 1 }{|D|\sqrt{\lambda}}    + \sqrt{ \frac{ \mathcal N(\lambda)}{|D|}}\right)      \log\frac{2}{\delta}\right)^v.
$$
Then Lemma \ref{Lemma:prob-to-exp} with $\mathcal A=\left(4M(\kappa+1) \left(\frac{ 1 }{|D|\sqrt{\lambda}}    + \sqrt{ \frac{ \mathcal N(\lambda)}{|D|}}\right)\right)^v$, $b=v$, and $c=2$ implies
\begin{equation}\label{Bound-dif-exp-1}
   E\left[\mathcal Q_{D,\lambda}^{2v}\mathcal P_{D,\lambda}^v|R_{D,\lambda}\leq 1/4\right]
    \leq
 2\Gamma(v+1) \left(4M(\kappa   +1) \left(\frac{ 1 }{|D|\sqrt{\lambda}}    + \sqrt{ \frac{ \mathcal N(\lambda)}{|D|}}\right)\right)^v.
\end{equation}
But
$
     \lambda |D|\geq (2vC_1^*(\log(1+\kappa)+2))^2\log^4|D|
$
together with Lemma \ref{Lemma:Bound-R}, $\mathcal N(\lambda)\leq \kappa\lambda^{-1}$, and  $\lambda\geq |D|^{-1}$ yields
\begin{eqnarray}\label{probab.big.r}
   &&P\left[R_{D,\lambda}> 1/2\right]
   \leq 4\exp\left\{-\frac{2vC_1^*(\log(1+\kappa)+2)\log^2{|D|}}{C_1^*(2+\log((\kappa+1)|D|))}\right\}\nonumber\\
   &\leq&
   4\exp\left\{-2v\log|D|\right\}
   =4|D|^{-2v}.
\end{eqnarray}
Therefore, we get from
$
   \mathcal Q_{D,\lambda}^{2v} \mathcal P^v_{D,\lambda}\leq (2(\kappa+1)\kappa M\lambda^{-3/2})^v
$
and (\ref{probab.big.r}) that
\begin{eqnarray*}
    &&E\left[\mathcal Q_{D,\lambda}^{2v} \mathcal P^v_{D,\lambda}\right]
     =
    E\left[\mathcal Q_{D,\lambda_j}^4 \mathcal P^2_{D,\lambda_j}|R_{D,\lambda}\leq 1/4\right]P[R_{D,\lambda}\leq 1/4]\\
   & +&
      E\left[\mathcal Q_{D,\lambda_j}^4 \mathcal P^2_{D,\lambda_j}|R_{D,\lambda}>1/4\right]P[R_{D,\lambda}>1/4]\\
    &\leq&
    2\Gamma(v+1) \left(4M(\kappa   +1) \left(\frac{ 1 }{|D|\sqrt{\lambda}}    + \sqrt{ \frac{ \mathcal N(\lambda)}{|D|}}\right)\right)^v
    +4(2(\kappa+1)\kappa M\lambda^{-3/2})^v|D|^{-2v}
    \\
    &\leq&
       2(2(\kappa+1)M)^v\max\{\Gamma(v+1)2^v,2\kappa^v\} \left(\left(\frac{ 1 }{|D|\sqrt{\lambda}}    + \sqrt{ \frac{ \mathcal N(\lambda)}{|D|}}\right)^v+ |D|^{-v/2}\right).
\end{eqnarray*}
 This proves (\ref{Bound-dif-exp-2}).
Noting further
$$
       \mathcal Q_{D,\lambda}\leq \sqrt{1+\kappa}\lambda^{-1/2},
$$
we have from $0\leq u\leq 4v$, Lemma \ref{Lemma:Bound-Q}, and (\ref{probab.big.r}) that
\begin{eqnarray*}
      E\left[\mathcal Q_{D,\lambda}^u\right]
     &\leq&
       E\left[\mathcal Q_{D,\lambda}^u|R_{D,\lambda}>1/4\right]P[R_{D,\lambda}>1/4]
      +
      E\left[\mathcal Q_{D,\lambda}^u|R_{D,\lambda}\leq1/4\right]P[R_{D,\lambda}\leq1/4]\\
      &\leq&
      4(1+\kappa)^{u/2}|D|^{u/2-2v}+2^{u/2}
      \leq 5(2+\kappa)^{u/2}.
\end{eqnarray*}
The bound of \eqref{bound-dif-exp-4} can be derived in the same way. This completes the proof of Lemma \ref{Lemma:bound-est-prod}.
\endproof

We then turn to prove Theorem \ref{Theorem:DKRR-1}.

\proof{Proof of Theorem \ref{Theorem:DKRR-1}.}
%It follows from (\ref{choose-lambda-1} that
%\begin{equation}\label{lambda-and-1}
%     \lambda_j|D_j|\geq 1.
%\end{equation}
%Due to  Assumption \ref{Assumption:regularity} with $\frac12\leq r\leq 1$, we have
%\begin{equation}\label{app.123}
%    \|(L_{K,D_j}+\lambda_j I)^{-1/2}f_\rho\|_K \leq
%    \|(L_{K,D_j}+\lambda_j I)^{-1/2}L_K^{r-1/2}\|h_\rho\|_K
%    \leq \lambda^{-1+r}_j\mathcal Q_{D_j,\lambda}^{r}\|h_\rho\|_K.
%\end{equation}
Noting (\ref{choose-lambda-1}) with $C_1:=(4C_1^*(\log(1+\kappa)+2))^2$, we   have $\lambda |D_j|\geq (4C_1^*(\log(1+\kappa)+2))^2\log^4|D_j| $ for all $j=1,\dots,m$.
Then, it follows from Lemma \ref{Lemma:bound-est-prod} with $v=2$, $u=4r$, (\ref{choose-lambda-1}), $r\geq1/2$,  and (\ref{resriction-on-m-1})
that
\begin{eqnarray}\label{app.err-1}
    && \sum_{j=1}^m\frac{|D_j|}{|D|}\lambda_j^{2r}\|h_\rho\|_\rho^2 E\left[\mathcal Q_{D_j,\lambda_j}^{2r}\right]\nonumber\\
    &\leq& C_2^*\left(\sum_{j:|D_j|\geq   |D|^{ \frac{ 1}{2r+s}}\log^4|D|} \frac{|D_j|}{|D|}|D|^{-\frac{2r}{2r+s}}
     +
     \sum_{j:|D_j|\leq |D|^{ \frac{ 1}{2r+s}}\log^4|D|}\frac{|D_j|}{|D|}|D_j|^{-2r}\log^{8r}|D|\right)\nonumber\\
     &\leq&
      2C_2^*|D|^{-\frac{2r}{2r+s}},
\end{eqnarray}
where $C_2^*:=5(2+\kappa)^{2r} (4C_1^*(\log(1+\kappa)+2))^{4r}\|h_\rho\|_\rho^2$.
Furthermore, it follows from Lemma \ref{Lemma:bound-est-prod} with $v=2$, Assumption \ref{Assumption:effective dimension}, and (\ref{choose-lambda-1}) that
\begin{eqnarray*}
     &&\sum_{j=1}^m\frac{|D_j|^2}{|D|^2}E\left[\mathcal Q_{D_j,\lambda_j}^4 \mathcal P^2_{D_j,\lambda_j}\right]\\
     &\leq&
     80M^2(\kappa   +1)^4\sum_{j=1}^m\frac{|D_j|^2}{|D|^2}  \left(\left(\frac{ 1 }{|D_j|\sqrt{\lambda_j}}    + \sqrt{ \frac{ \mathcal N(\lambda_j)}{|D_j|}}\right)^2+ |D_j|^{-1}\right)\\
      &\leq&
      160 M^2(\kappa   +1)^4|D|^{-1} \sum_{j=1}^m \left(2|D|^{-1}\lambda_j^{-1} +\frac{|D_j|}{|D|}\mathcal N(\lambda_j)\right)\\
      &\leq&
      160(C_0+2)  M^2(\kappa   +1)^4|D|^{-1}\sum_{j=1}^m \left(|D|^{-1}
       \lambda_j^{-1} +\frac{|D_j|}{|D|}\lambda_j^{-s}\right).
\end{eqnarray*}
But (\ref{choose-lambda-1}) and (\ref{resriction-on-m-1}) yield
\begin{eqnarray*}
   &&\sum_{j=1}^m\left(
       \lambda_j^{-1} +\frac{|D_j|}{|D|}\lambda_j^{-s}\right)\\
      &\leq&
      \sum_{j:|D_j|\geq   |D|^{ \frac{ 1}{2r+s}}\log^4|D|}\left(|D|^{-1}
       \lambda_j^{-1} +\frac{|D_j|}{|D|}\lambda_j^{-s}\right)\\
      &+&\sum_{j:|D_j|<|D|^{ \frac{ 1}{2r+s}}\log^4|D|}\left(|D|^{-1}
       \lambda_j^{-1} +\frac{|D_j|}{|D|}\lambda_j^{-s}\right)\\
       &\leq&
       (4C_1^*(\log(1+\kappa)+2))^{-2s}\left(m|D|^{\frac{-2r-s+1}{2r+s}}+|D|^{\frac{s}{2r+s}}+\log^{-4}|D|
       +|D|^{\frac{s}{2r+s}}\right)\\
       &\leq&
       4(4C_1^*(\log(1+\kappa)+2))^{-2s}|D|^{\frac{s}{2r+s}}.
\end{eqnarray*}
Thus, we have
\begin{equation}\label{Sample-error-1}
   \sum_{j=1}^m\frac{|D_j|^2}{|D|^2}E\left[\mathcal Q_{D_j,\lambda_j}^4 \mathcal P^2_{D_j,\lambda_j}\right]\leq C_3^*|D|^{-\frac{2r}{2r+s}},
\end{equation}
where $C_3^*:=640(C_0+2)  M^2(\kappa   +1)^4 (4C_1^*(\log(1+\kappa)+2))^{-2s}.$
Plugging (\ref{app.err-1}) and (\ref{Sample-error-1}) into
  Proposition \ref{Prop:error-decomposition}, we complete the proof of Theorem \ref{Theorem:DKRR-1} with $C_2=4C_2^*+2C_3^*$.
\endproof

The proof of Corollary \ref{corollary:DKRR-1} is almost the same as above. We present it for the sake of completeness.

\proof{Proof of Corollary \ref{corollary:DKRR-1}.}
Setting $C_1:=(4C_1^*(\log(1+\kappa)+2))^2$, we   have $\lambda |D_j|\geq (4C_1^*(\log(1+\kappa)+2))^2\log^4|D_j| $ for all $j=1,\dots,m$.
Then, it follows from Lemma \ref{Lemma:bound-est-prod} with $u=4r$ and $v=2$
that
$$
    \sum_{j=1}^m\frac{|D_j|}{|D|}\lambda_j^{2r}\|h_\rho\|_\rho^2 E\left[\mathcal Q_{D_j,\lambda}^{2r}\right]
     \leq  C_2^* \sum_{j=1}^m \frac{|D_j|}{|D|}|D|^{-\frac{2r}{2r+s}}
     =
       C_2^*|D|^{-\frac{2r}{2r+s}}.
$$
Furthermore, it follows from Lemma \ref{Lemma:bound-est-prod} with $v=2$,  Assumption \ref{Assumption:effective dimension}, and \eqref{resriction-on-m-2} that
\begin{eqnarray*}
     &&\sum_{j=1}^m\frac{|D_j|^2}{|D|^2}E\left[\mathcal Q_{D_j,\lambda_j}^4 \mathcal P^2_{D_j,\lambda_j}\right]\\
     &\leq&
      160(C_0+2)  M^2(\kappa   +1)^4|D|^{-1}\sum_{j=1}^m \left(|D|^{-1}
       \lambda_j^{-1} +\frac{|D_j|}{|D|}\lambda_j^{-s}\right)\\
       &\leq&
       160(C_0+2)  M^2(\kappa   +1)^4(4C_1^*(\log(1+\kappa)+2))^{-2s}|D|^{-1}\left(m|D|^{\frac{-2r-s+1}{2r+s}}+|D|^{\frac{s}{2r+s}}\right)\\
       &\leq&
       C_3^*|D|^{-\frac{2r}{2r+s}}.
\end{eqnarray*}
Plugging the above two estimates into Proposition \ref{Prop:error-decomposition} completes the proof of Corollary \ref{corollary:DKRR-1}.
\endproof

\subsection{Proof of Theorem \ref{Theorem:Generalization-error-AdaDKRR}}
In this part, we use Theorem \ref{Theorem:DKRR-1} and Proposition \ref{Prop.error-dec-adadkrr} to prove Theorem \ref{Theorem:Generalization-error-AdaDKRR}.  Firstly, we present a detailed bound of the global approximation as follows.

\begin{proposition}\label{Proposition:Global-app-1}
Under Assumption \ref{Assumption:bounded for output} and Assumption \ref{Assumption:regularity}    with $1/2\leq r\leq 1$,
if $\rho_X$ is a uniform distribution, $\mu\geq (8C_1^*(\log(1+\kappa)+2))^2 \max_{j=1,\dots,m}\frac{\log^4|D_j^{tr}|}{|D_j^{tr}|}$,  $\Xi_n$ satisfies \eqref{ass-points} for some $c,\beta>0$, and $n$ satisfies $\mu n^\beta\geq 2c$, then for any $\lambda\in (\mu,1)$, there holds
\begin{eqnarray*}
&& E\left[\left\|{f}^{global}_{D^{tr}, \lambda,\mu}-\sum_{j=1}^m\frac{|D_j^{tr}|}{|D^{tr}|}f_{D_j,\lambda}
 \right\|_\rho^2\right]\\
    &\leq& C_4\mu^{2r}+
    C_4\mu\left(\lambda^{2r-1}+  \sum_{j=1}^m\frac{|D_j^{tr}|}{|D^{tr}|}
    \left(\left(\frac{ 1 }{|D_j^{tr}|{\lambda}}    + \sqrt{ \frac{ \mathcal N(\lambda)}{\lambda|D_j^{tr}|}}\right)^2+ |D_j^{tr}|^{-1}\right)  \right)  ,
\end{eqnarray*}
where $C_4$ is a constant depending only on $\|h_\rho\|_\rho,M,r,c$, and $\beta$.
\end{proposition}

To prove the above proposition, we need the following preliminary lemma.

\begin{lemma}\label{Lemma:bound-Q-monte}
If $\rho_X$ is a uniform distribution, $\Xi_n$ satisfies \eqref{ass-points} for some $c,\beta>0$,  and $n$ satisfies $\mu n^\beta\geq 2c$, then
\begin{equation}\label{Bound-Q-monte}
    \mathcal Q_{\Xi_n,\mu}\leq \sqrt{2}.
\end{equation}
\end{lemma}

\proof{Proof.} From the definition of the operator norm $\|\cdot\|$ and the fact $\|f\|_K=\sup\limits_{\|g\|_K\leq 1}\langle f,g\rangle_K$, we obtain
% Let $\{\sigma_\ell,\varphi_\ell\}$ be the eigen-pair of $L_K$   with $\sigma_1\geq \sigma_2\geq\dots>0$ and $\{\varphi_\ell\}_{\ell}$ the normalized eigen-functions in $\mathcal H_K$.
%   If we regard $L_K$ as an operator on
% $L^2_{\rho_X}$, the normalized eigen-functions in $L^2_{\rho_X}$ are
% $\{\frac{1}{\sqrt{\sigma_\ell}} \varphi_\ell\}_{\ell}$, Due to the definition  of operator norm and $(\mathcal L_\phi+\lambda I )^{-v} g\in \mathcal N_\phi$, we have
\begin{eqnarray*}
  &&\|L_K-L_{K,\Xi_n}\|=\sup_{\|f\|_K\leq 1}\|( (L_{K,\Xi_n}-  L_K)f\|_K
   =
  \sup_{\|f\|_K\leq 1}\sup_{\|g\|_K\leq 1}
  \langle  (L_{K,\Xi_n}- L_K)f,g\rangle_K\\
  &=&
   \sup_{\|g\|_K\leq 1,\|f\|_K\leq 1}
   \left|\left\langle\int_{\mathcal X}f(x')K_{x'}d\rho_X-\frac1n\sum_{k=1}^nf(\xi_k)K_{\xi_k},  g\right\rangle_K\right|\\
   &=&
   \sup_{\|g\|_K\leq 1,\|f\|_K\leq 1} \left|\int_{\mathcal X}f(x')\langle K_{x'},  g\rangle_K d\rho_X-\frac1n\sum_{k=1}^{n} f(\xi_k)\langle  K_{\xi_k},  g\rangle_K\right|\\
   &=&
   \sup_{\|g\|_K\leq 1,\|f\|_K\leq 1}\left|\int_{\mathcal X}f(x') g(x')d\rho_X-\frac1n\sum_{k=1}^{n} f(\xi_k) g(x_i)\right|.
\end{eqnarray*}
Then, it  follows from \eqref{ass-points} that
$$
   \|L_{K,\Xi_n}-L_K\|\leq cn^{-\beta}.
$$
Therefore,
$$
   \|(L_K+\mu I)^{-1/2}(L_K-L_{K,\Xi_n})(L_K+\mu I)^{-1/2}\|\leq c\mu^{-1}n^{-\beta}.
$$
Then it follows from  $\mu n^\beta\geq 2c$ that
$$
   \|(L_K+\mu I)^{-1/2}(L_K-L_{K,\Xi_n})(L_K+\mu I)^{-1/2}\|\leq 1/2.
$$
Noting further
\begin{eqnarray*}
   &&\mathcal Q_{D,\Xi_n}^2=
   \left\|(L_K+\mu I)^{1/2}(L_{K,\Xi_n}+\mu I)^{-1}(L_K+\mu I)^{1/2}\right\|\\
   &\leq &\left\| (L_K+\mu I)^{1/2}\left[(L_{K,\Xi_n}+\mu I)^{-1}-(L_{K}+\mu I)^{-1}\right](L_K+\mu I)^{1/2} \right\| +1\\
   &\leq &
   \frac12
   \mathcal Q_{D,\Xi_n}^2+1,
\end{eqnarray*}
we then obtain $\mathcal Q_{D,\Xi_n}\leq \sqrt{2}$. This completes the proof of Lemma \ref{Lemma:bound-Q-monte}.
\endproof

With the help of the above two lemmas and Proposition \ref{Prop:global-app-err}, we are in a position to prove Proposition \ref{Proposition:Global-app-1} as follows.

\proof{Proof of Proposition \ref{Proposition:Global-app-1}.}
Due to Proposition \ref{Prop:global-app-err}, H\"{o}lder's inequality, Jensen's inequality, and the basic inequalities $(a+b)^2\leq 2a^2+2b^2$ and $(a+b)^4\leq 8a^4+8b^4$ for $a,b\geq 0$, we have
% \begin{eqnarray*}
% && E\left[\left\|{f}^{global}_{D^{tr}, \lambda,\mu}-\sum_{j=1}^m\frac{|D_j^{tr}|}{|D^{tr}|}f_{D_j,\lambda}
%  \right\|_\rho^2\right]
%  \leq
%  4\mu^{2r}\|h_\rho\|_\rho^2 \sum_{j=1}^m\frac{|D_j^{tr}|}{|D^{tr}|}
%    E\left[\left(\mathcal Q_{D_j^{tr},\mu}\mathcal Q_{D_j^{tr},\mu}^*+1\right)^2\mathcal
%      Q_{\Xi_n,\mu}^{4r}\right]\\
%      &+& 4\mu^{2r}\|h_\rho\|_\rho^2 \sum_{j=1}^m\frac{|D_j^{tr}|}{|D^{tr}|}E\left[\mathcal Q_{D_j^{tr},\mu}^{4r}\right]
%      +
%      16\mu\sum_{j=1}^m\frac{|D_j^{tr}|}{|D^{tr}|}\left(E\left[ \lambda^{-2}\mathcal Q^8_{D_j^{tr},\lambda}\mathcal P^4_{D_j^{tr},\lambda}+\lambda^{4r-2}\mathcal Q_{D_j^{tr},\lambda}^{8r-4}\|h\|_\rho^4\right]\right)^{1/2}\\
%      &\times&
%     \left(E\left[\mathcal Q_{D_j^{tr},\mu}^4 +\left(\mathcal Q_{D_j^{tr},\mu}\mathcal Q_{D_j^{tr},\mu}^*+1\right)^4\mathcal Q^4_{\Xi_n,\mu}\right]\right)^{1/2}.
% \end{eqnarray*}
\begin{eqnarray*}
&& E\left[\left\|{f}^{global}_{D^{tr}, \lambda,\mu}-\sum_{j=1}^m\frac{|D_j^{tr}|}{|D^{tr}|}f_{D_j,\lambda}
 \right\|_\rho^2\right]\\
 &\leq&
 4\mu^{2r}\|h_\rho\|_\rho^2 \sum_{j=1}^m\frac{|D_j^{tr}|}{|D^{tr}|}
   E\left[\left(\mathcal Q_{D_j^{tr},\mu}\mathcal Q_{D_j^{tr},\mu}^*+1\right)^2\mathcal
     Q_{\Xi_n,\mu}^{4r}\right]
     + 4\mu^{2r}\|h_\rho\|_\rho^2 \sum_{j=1}^m\frac{|D_j^{tr}|}{|D^{tr}|}E\left[\mathcal Q_{D_j^{tr},\mu}^{4r}\right]\\
     &+&
     16\mu\sum_{j=1}^m\frac{|D_j^{tr}|}{|D^{tr}|}\left(E\left[ \lambda^{-2}\mathcal Q^8_{D_j^{tr},\lambda}\mathcal P^4_{D_j^{tr},\lambda}+\lambda^{4r-2}\mathcal Q_{D_j^{tr},\lambda}^{8r-4}\|h\|_\rho^4\right]\right)^{1/2}\\
     &\times&
    \left(E\left[\mathcal Q_{D_j^{tr},\mu}^4 +\left(\mathcal Q_{D_j^{tr},\mu}\mathcal Q_{D_j^{tr},\mu}^*+1\right)^4\mathcal Q^4_{\Xi_n,\mu}\right]\right)^{1/2}.
\end{eqnarray*}
Using \eqref{Bound-dif-exp-3}, \eqref{bound-dif-exp-4}, and Lemma \ref{Lemma:bound-Q-monte} with $u=4$ and $v=4$, we get from H\"{o}lder's inequality and   $\mu\geq  (8C_1^*(\log(1+\kappa)+2))^2 \max_{j=1,\dots,m}\frac{\log^4|D_j^{tr}|}{|D_j^{tr}|}$  that
\begin{eqnarray*}
    E\left[\left(\mathcal Q_{D_j^{tr},\mu}\mathcal Q_{D_j^{tr},\mu}^*+1\right)^2\mathcal
     Q_{\Xi_n,\mu}^{4r}\right]
     \leq
     2^{2r+1}(1+5(2+\kappa)(4+\kappa)).
\end{eqnarray*}
Moreover, it follows from \eqref{Bound-dif-exp-3} with $u=4r$ and $v=4$ that
$$
      E\left[\mathcal Q_{D_j^{tr},\mu}^{4r}\right]\leq 5(2+\kappa)^{2r}.
$$
Using \eqref{Bound-dif-exp-2} with $v=4$ and \eqref{Bound-dif-exp-3} with $u=8r-4$ and $v=4$, we obtain from $\lambda\geq(8C_1^*(\log(1+\kappa)+2))^2 \max_{j=1,\dots,m}\frac{\log^4|D_j^{tr}|}{|D_j^{tr}|}$ that
\begin{eqnarray*}
    &&\left(E\left[ \lambda^{-2}\mathcal Q^8_{D_j^{tr},\lambda}\mathcal P^4_{D_j^{tr},\lambda}+\lambda^{4r-2}\mathcal Q_{D_j^{tr},\lambda}^{8r-4}\|h\|_\rho^4\right]\right)^{1/2}\\
    &\leq&
    \lambda^{-1}( C_0^*)^{1/2}
    \left(\left(\frac{ 1 }{|D_j^{tr}|\sqrt{\lambda}}    + \sqrt{ \frac{ \mathcal N(\lambda)}{|D_j^{tr}|}}\right)^2+ |D_j^{tr}|^{-1}\right)
    +
    \lambda^{2r-1}\|h\|_\rho^2 \sqrt{5}(2+\kappa)^{2r-1}.
\end{eqnarray*}
 % Furthermore, Lemma \ref{Lemma:bound-Q-monte} with $u=4$ and Lemma \ref{Lemma:bound-est-prod} yields
{Furthermore, Lemma \ref{Lemma:bound-est-prod} with $u=4$ and Lemma \ref{Lemma:bound-Q-monte} yield}
\begin{eqnarray*}
 &&\left(E\left[\mathcal Q_{D_j^{tr},\mu}^4 +\left(\mathcal
    Q_{D_j^{tr},\mu}\mathcal Q_{D_j^{tr},\mu}^*+1\right)^4\mathcal Q^4_{\Xi_n,\mu}\right]\right)^{1/2}\\
    &\leq&
    \left(E\left[\mathcal Q_{D_j^{tr},\mu}^4\right]\right)^{1/2}
    +
   4\sqrt{2} \left(1+ \left(E\left[\mathcal
    Q_{D_j^{tr},\mu}^8\right]\right)^{1/2} \left(E\left[\left(\mathcal
    Q^*_{D_j^{tr},\mu}\right)^8\right]\right)^{1/2} \right)^{1/2}\\
    &\leq&
    \sqrt{5}(2+\kappa)+4\sqrt{2}+4\sqrt{2}   4\sqrt{10}  (2+\kappa)(4+\kappa).
\end{eqnarray*}
Combining all the above estimates, we have from $\mu\geq  (8C_1^*(\log(1+\kappa)+2))^2 \max_{j=1,\dots,m}\frac{\log^4|D_j^{tr}|}{|D_j^{tr}|}$  that
\begin{eqnarray*}
&& E\left[\left\|{f}^{global}_{D^{tr}, \lambda,\mu}-\sum_{j=1}^m\frac{|D_j^{tr}|}{|D^{tr}|}f_{D_j,\lambda}
 \right\|_\rho^2\right]
 \leq
 \mu^{2r}\|h_\rho\|_\rho^2 \left(  2^{2r+3}(1+5(2+\kappa)(4+\kappa))
     +  20(2+\kappa)^{2r} \right) \\
     &+&
     16\mu\left(\sqrt{5}(2+\kappa)+4\sqrt{2}+4\sqrt{2}   4\sqrt{10}  (2+\kappa)(4+\kappa)\right)\\
     &\times&\hspace{-0.03in}
   \left(\hspace{-0.03in}\lambda^{-1}\left(  C_0^*\right)^{1/2}\sum_{j=1}^m\frac{|D_j^{tr}|}{|D^{tr}|}
    \hspace{-0.03in}\left(\hspace{-0.03in}\left(\frac{ 1 }{|D_j^{tr}|\sqrt{\lambda}}  \hspace{-0.03in}  + \hspace{-0.03in}\sqrt{ \frac{ \mathcal N(\lambda)}{|D_j^{tr}|}}\right)^2 \hspace{-0.03in}+ \hspace{-0.03in}|D_j^{tr}|^{-1}\hspace{-0.03in}\right)
   \hspace{-0.03in} +\hspace{-0.03in}
    \lambda^{2r-1}\|h\|_\rho^2 \sqrt{5}(2+\kappa)^{2r-1}\hspace{-0.03in}\right)\\
    &\leq&
    C_4\mu^{2r}+
    C_4\mu\left(\lambda^{2r-1}+  \sum_{j=1}^m\frac{|D_j^{tr}|}{|D^{tr}|}
    \left(\left(\frac{ 1 }{|D_j^{tr}|{\lambda}}    + \sqrt{ \frac{ \mathcal N(\lambda)}{\lambda|D_j^{tr}|}}\right)^2+ |D_j^{tr}|^{-1}\right)  \right),
\end{eqnarray*}
where $C_4$ is a constant depending only on $\kappa$, $M$, $r$, $C_0^*$, and $\|h_\rho\|_\rho$. This completes the proof of Proposition \ref{Proposition:Global-app-1}.
\endproof

Finally, we prove Theorem \ref{Theorem:Generalization-error-AdaDKRR}.

\proof{Proof of Theorem \ref{Theorem:Generalization-error-AdaDKRR}.} Let $\bar{\lambda}:=C_1|D|^{-\frac{1}{2r+s}}$. We have from $|D_j|\geq |D|^{\frac{1}{2r+s}}\log ^4|D|$ and Assumption \ref{Assumption:effective dimension} that
\begin{eqnarray*}
    \sum_{j=1}^m\frac{|D_j^{tr}|}{|D^{tr}|}
    \left(\left(\frac{ 1 }{|D_j^{tr}|{\bar{\lambda}}}    + \sqrt{ \frac{ \mathcal N(\bar{\lambda})}{\bar{\lambda}|D_j^{tr}|}}\right)^2+ |D_j^{tr}|^{-1}\right)\leq\tilde{C}_2 m|D|^{-\frac{2r-1}{2r+s}},
\end{eqnarray*}
where $\tilde{C}_2$ is a constant depending only on $C_0$ and $C_1$. Then it follows from   Proposition \ref{Proposition:Global-app-1} that, under \eqref{restriction-on-m-3}, there holds
$$
   E\left[\left\|{f}^{global}_{D^{tr}, \bar{\lambda},\mu}-\sum_{j=1}^m\frac{|D_j^{tr}|}{|D^{tr}|}f_{D_j,\lambda}
   \right\|_\rho^2\right]\leq \tilde{C}_3\left(\mu^{2r}+\mu|D|^{-\frac{2r-1}{2r+s}}\right),
$$
where $\tilde{C}_3$ is a constant depending only on $C_4,C_1$, and $C_0$.
Furthermore, it follows from Corollary \ref{corollary:DKRR-1} that, under \eqref{restriction-on-m-3}, there holds
$$
    E\left[\left\|\sum_{j=1}^m\frac{|D_j^{tr}|}{|D^{tr|}}f_{D_j,\bar{\lambda}}- f_\rho\right\|_\rho^2\right]
    \leq
    C_2|D|^{-\frac{2r}{2r+s}}.
$$
Plugging the above two estimates into Proposition \ref{Prop.error-dec-adadkrr} and noting $\mu\leq |D|^{-\frac1{2r+s}}$, we have
$$
     E\left[ \left\| \overline{f}^{Ada}_{D,\vec{\lambda^*}}-f_\rho\right\|_\rho^2 \right]
      \leq
  C_5\left(m\frac{\log|\Lambda|}{|D|}+  |D|^{-\frac{2r}{2r+s}}\right),
$$
where $C_5$ is a constant depending only on $\tilde{C}_2$, $C_2$, and $C_1'$.  This completes the proof of Theorem \ref{Theorem:Generalization-error-AdaDKRR}.
\endproof

\end{document}